\documentclass{article}

\usepackage{microtype}
\usepackage{graphicx}
\usepackage{subcaption}
\usepackage{booktabs} 

\usepackage{hyperref}

 \usepackage[accepted]{icml2026}

\usepackage{amsmath}
\usepackage{amssymb}
\usepackage{mathtools}
\usepackage{amsthm}
\usepackage{subcaption}
\usepackage{algorithm}
\usepackage{algorithmicx}
\usepackage{algpseudocode}
\usepackage{amsmath}
\usepackage{xfrac}
\usepackage{wrapfig}
\usepackage{booktabs}
\usepackage{enumitem}
\usepackage{multirow}
\usepackage{fix-cm}
\usepackage{cuted} 
\usepackage[capitalize,noabbrev]{cleveref}
\Crefname{figure}{Fig.}{Figs.}
\Crefname{table}{Tab.}{Tabs.}
\Crefname{equation}{Eq.}{Eqs.}
\Crefname{section}{Sect.}{Sect.}

\theoremstyle{plain}

\theoremstyle{definition}

\theoremstyle{remark}

\usepackage[textsize=tiny]{todonotes}

\icmltitlerunning{Softplus Attention with Re-weighting Boosts Length Extrapolation in Large
Language Models}

\begin{document}

\twocolumn[
    \icmltitle{Softplus Attention with Re-weighting Boosts Length Extrapolation in Large
        Language Models}



    \icmlsetsymbol{corresponding}{*}
    \begin{icmlauthorlist}
        \icmlauthor{Bo Gao }{corresponding,a,b,c,d}
        \icmlauthor{Michael~W.~Spratling}{e}
        \icmlauthor{Letizia Gionfrida}{d}
    \end{icmlauthorlist}

    \icmlaffiliation{a}{Department of Intelligent Manufacturing and Electrical Engineering, Nanyang Normal University, Nanyang 473061, China.}
    \icmlaffiliation{b}{Collaborative Innovation Center of Intelligent Explosion-proof Equipment, Henan Province.}
    \icmlaffiliation{c}{Baopu Lab.}
    \icmlaffiliation{d}{The Department of Informatics, King's College London, Strand, London, WC2R 2LS, UK.}
    \icmlaffiliation{e}{Department of Behavioural and Cognitive Sciences, University of Luxembourg, L-4366 Esch-Belval, Luxembourg}

    \icmlcorrespondingauthor{Bo Gao}{bo.gao@nynu.edu.cn}

    \icmlkeywords{Machine Learning, ICML}

    \vskip 0.3in
]

\ifdefined\ispreprint
    \begin{strip}
        \centering
        \includegraphics[width=\textwidth]{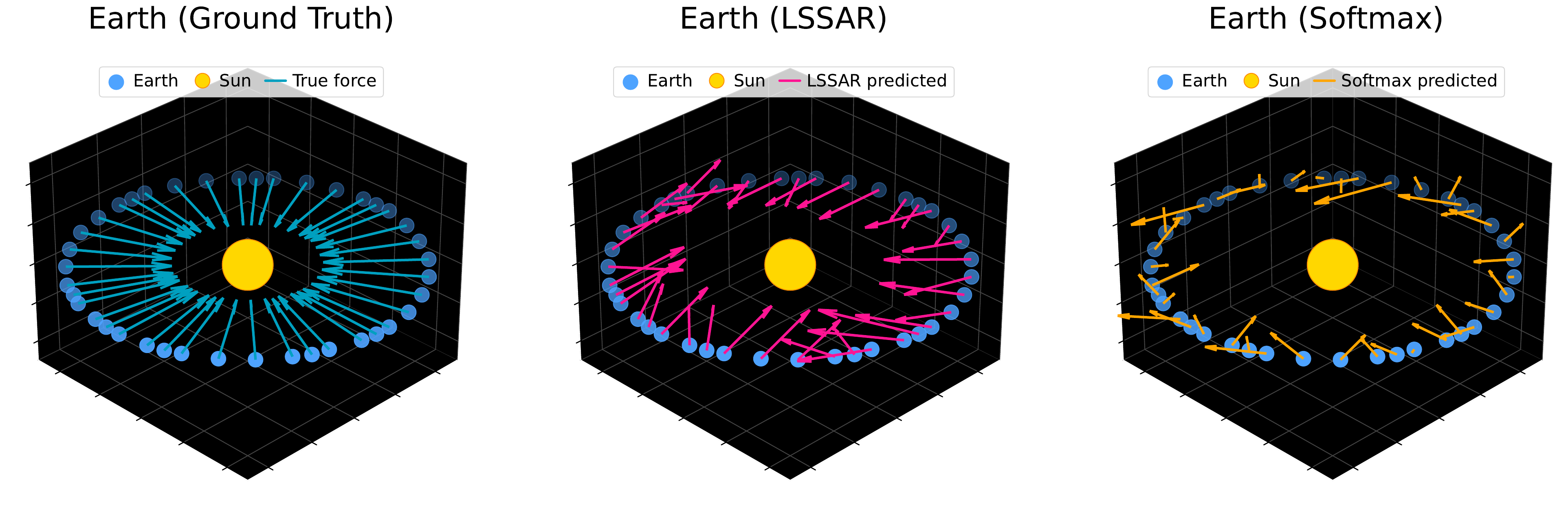}
        \captionsetup{type=figure, width=\textwidth}
        \captionof{figure}{Force predictions for Earth's orbit. Left: true forces. Middle: GPT (LSSAR) correctly captures radial gravitational structure. Right: GPT (Softmax) is incoherent.}
        \label{fig:force_earth}
    \end{strip}
\fi



\printAffiliationsAndNotice{}  

\begin{abstract}
    Large language models have achieved remarkable success in recent years,
    primarily due to self-attention. However,
    traditional Softmax attention suffers from numerical instability and
    reduced performance as the number of inference tokens increases. This work
    addresses these issues by proposing a new design principle for attention,
    viewing it as a two-stage process.
    The first stage (normalisation) refines standard attention by replacing Softmax with the more numerically stable Softplus followed by
    $l_{1}$-normalisation. Furthermore, we introduce a dynamic scale factor based on invariance entropy. We show that this novel attention mechanism outperforms conventional Softmax attention, and state-of-the-art Softmax-free alternatives.
    Our second proposal is to introduce a second processing stage (sharpening) which consists of a
    re-weighting mechanism that
    amplifies significant attentional weights while diminishing weaker ones. This enables
    the model to concentrate more effectively on relevant tokens, mitigating
    the attention sink phenomenon, and fundamentally improving length extrapolation.
    This novel, two-stage, replacement for self-attention is shown to ensure numerical stability and dramatically
    improve length extrapolation, maintaining a nearly constant validation
    loss at 16$\times$ the training length while achieving superior results on
    challenging long-context retrieval tasks and downstream benchmarks.
    Furthermore, symbolic regression experiments demonstrate that our method enables
    models to recover Newton's gravitational law from orbital trajectory sequences,
    providing evidence that appropriate attention mechanisms are crucial for
    foundation models to develop genuine physical world models. Our code is
    available at \url{https://github.com/iminfine/freeattn}.
\end{abstract}

\section{Introduction}

\ifdefined\ispreprint\else
    \begin{figure}[t!]
        \centering
        \includegraphics[width=0.9\linewidth]{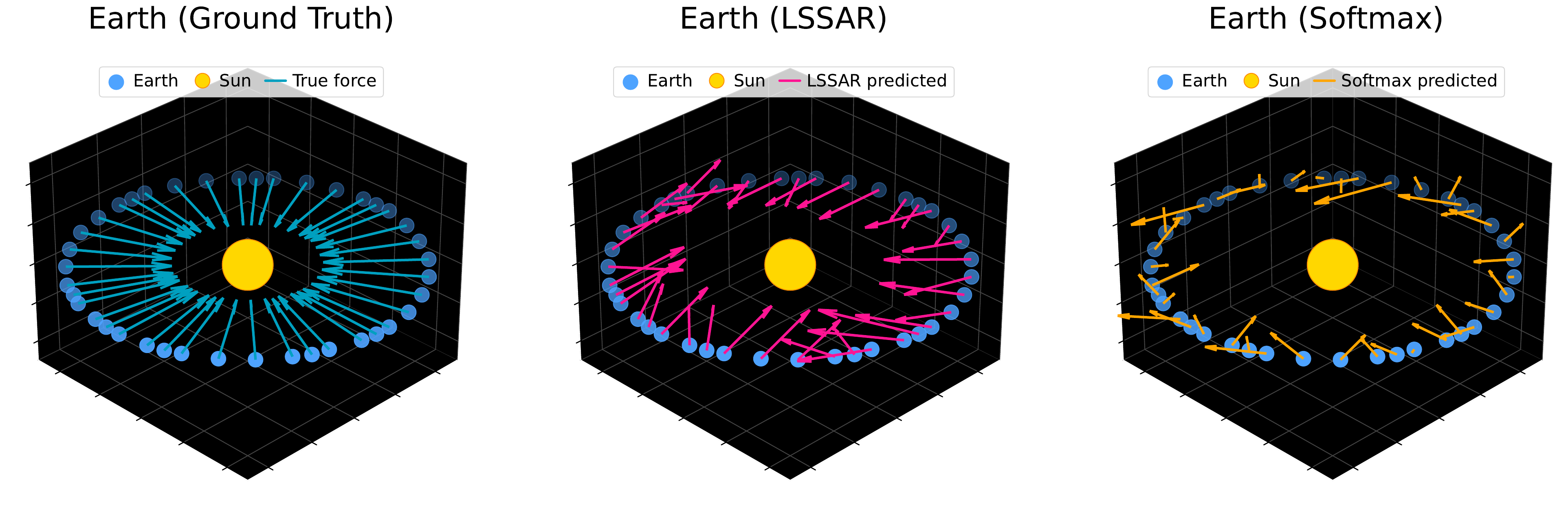}
        \caption{Force predictions for Earth's orbit. Left: true forces. Middle: GPT (LSSAR) which mostly captures the radial gravitational structure. Right: GPT (Softmax) predictions are incoherent.}
        \label{fig:force_earth}
    \end{figure}
\fi



Self-attention has been primarily responsible for the recent success of Large
Language Models (LLMs). Self-attention enables models to assess the
importance of different words within a sentence, capturing complex relationships
and dependencies in the data. Its effectiveness is mainly attributed to the
Softmax operation, as when the Softmax operation is omitted or replaced with
alternatives, model performance tends to decline \citep{wortsman2023replacing,ramapuram2024theory,shen2023study}.
Although widely used, Softmax self-attention has two main limitations.
Firstly, it suffers from numerical instability due to the exponential function
($e^{x}$), especially when scaling model sizes to trillions of parameters
\citep{stable_transformer}. Secondly, as the token length increases during inference,
the attention scores calculated by self-attention become smoother and lack
distinct peaks. This ``attention smoothing'' hinders the model's ability to establish
connections between relevant tokens, thereby crippling the length
extrapolation capabilities of transformers \citep{chiang2022overcoming,velivckovic2024softmax}.
This issue is compounded by the ``attention sink'' phenomenon, where a few initial
tokens, often regardless of their semantic importance, attract a disproportionate
amount of attention, leading to suboptimal performance \citep{xiaoefficient}.
This paper addresses these problems simultaneously by proposing a new design
principle for attention mechanisms that offers improved numerical stability
and dramatically better performance at large token lengths. Our core proposal
is a new attention architecture consisting of two stages: a \textbf{Normalisation
    Stage} followed by a \textbf{Sharpening Stage}.

To develop a better normalisation stage, we first analysed the standard Softmax
function. We deconstructed it into its two functional components: a non-linear
positivity transformation ($e^{x}$) and a subsequent scaling by the $l_{1}$-norm. Our experiments revealed that the $l_{1}$-norm scaling is the
critical component for maintaining model performance. This insight frees us to
redesign the attention process for better stability and performance. We
replace the exponential function with the more numerically stable Softplus
activation and incorporate a dynamic length scale factor based on invariance
entropy. This creates our new normalisation stage, a novel mechanism we call
\textit{Length Scaled Softplus Attention} (LSSA), which outperforms the
standard attention mechanism not only at the training sequence length but
also for longer sequences.

However, a normalisation stage alone does not solve the attention smoothing problem.
Therefore, we introduce a re-weighting mechanism that sharpens the attention
distribution. Applied after LSSA, this mechanism amplifies
critical token relationships and suppresses noise using a power transformation.
This inherently sharpens attention peaks without needing post-hoc fixes like
positional interpolation \citep{chen2023extending,li2023functional}, which
retroactively stretches embeddings but fails to address the root cause of
attention smoothing.

The combination of these two proposed mechanisms, LSSA (for stable normalisation)
and re-weighting (for sharpening), results in our final model, LSSAR. The
effectiveness of this two-stage design is starkly demonstrated in
challenging ``needle-in-a-haystack'' passkey retrieval tasks, where LSSAR succeeds
far beyond its training length while standard attention fails completely.
Furthermore, LSSAR effectively mitigates the attention sink phenomenon by promoting
a more balanced attention distribution. LSSAR maintains a nearly constant
validation loss even at 16$\times$ the training token length and translates its
superior internal metrics into better performance on real-world NLP tasks.

Beyond standard benchmarks, we evaluate whether LSSAR helps models capture
physically meaningful structure through symbolic regression experiments on
planetary orbits. As illustrated in \cref{fig:force_earth}, a GPT-109M model
equipped with LSSAR predicts force vectors that are more consistent with the
Newtonian radial direction than the standard Softmax baseline. Symbolic
regression identifies an inverse-square form ($F \propto m/r^2$) from LSSAR's
outputs, whereas standard attention fails to match this physical structure.
Notably, even state-of-the-art LLMs that are assumed to be substantially larger
than our GPT-109M model (o3, Claude 4 Sonnet, and Gemini 2.5 Pro) do not recover
the inverse-square relation in the same protocol
\citep{vafa2025inductive}, suggesting that appropriate attention mechanisms are
crucial for foundation models to develop physically grounded world-model
representations.

In summary, this paper makes several contributions:
\begin{enumerate}[leftmargin=0.5cm]
    \item We introduce a novel Softmax-free attention mechanism, LSSA, which
          incorporates a dynamic length scale factor based on invariance entropy,
          demonstrating superior performance compared to standard attention
          mechanisms across a variety of sequence lengths.

    \item We propose a novel architectural re-weighting mechanism that fundamentally
          redesigns how attention scores are computed, inherently sharpening
          token relevance by amplifying critical weights and suppressing noise.

    \item When LSSA is combined with the proposed re-weighting method, the
          resulting model (LSSAR) not only excels in length extrapolation
          while ensuring numerical stability, as demonstrated by its success in
          passkey retrieval tasks where standard attention fails, but also
          translates into superior performance in real-world applications.

    \item We demonstrate through symbolic regression that LSSAR induces
          inductive biases aligned with physically meaningful sequence
          structure. Unlike standard Softmax attention, which fails to recover a
          compact physical relation in this setup, LSSAR recovers an
          inverse-square functional dependence from orbital trajectories.
\end{enumerate}

\section{Method}
\label{sec:method} The proposed attention mechanism consists of two stages. The
first involves normalising the raw attention scores to create a stable distribution,
and is described in \cref{sec:norm_stage}. The second re-weights
this distribution to sharpen focus and enhance length extrapolation, as
described in \cref{sec:sharpening_stage}. 

\subsection{The Normalisation Stage}\label{sec:norm_stage}

\subsubsection{Softmax Decomposition}
\label{sec:softmaxdecom}

\begin{equation}
    \mathbf{A}= \operatorname{Softmax}\left(\frac{\mathbf{Q}\mathbf{K}^{T}}{\sqrt{d}}
    +\mathbf{M}\right) \label{equ:rowatt}
\end{equation}

Scaled dot-product attention transforms queries ($\mathbf{Q}$), keys ($\mathbf{K}$),
and values ($\mathbf{V}$) into an output. First, attention scores $\mathbf{A}$
are produced via \cref{equ:rowatt}. Here, $\mathbf{Q}, \mathbf{K}, \mathbf{V}
    \in \mathbb{R}^{L \times d}$, where $L$ is the sequence length and $d$ is
the dimensionality. The optional term $\mathbf{M}$ is a mask matrix of shape
$L \times L$, which is essential in causal self-attention. The mask $\mathbf{M}$
is constructed with zeros on and below the diagonal and $-\infty$ above it,
ensuring the attention mechanism only considers past and present tokens. The
attention scores are then used to compute the output as $\mathbf{A}
    \mathbf{V}$.

The Softmax function is a cornerstone of the attention mechanism. Its
non-negative outputs are often viewed as an important factor for stable
attention. However, prior results with other non-negative activation functions
suggest that non-negativity alone is not sufficient for good performance
\citep{wortsman2023replacing,ramapuram2024theory,shen2023study}.

To investigate this decomposition, we conducted experiments that modify the sign
and normalisation of attention scores. Our results (see \cref{tab:modifiedatt} in
Appendix~\ref{app:modified_attention}) suggest that Softmax's effectiveness
comes from the combination of a nonlinear positivity transformation and
subsequent $l_{1}$ normalisation. Within this decomposition, the normalisation step
plays a central role in preserving performance, rather than positivity alone.

Motivated by this observation, we decompose the Softmax operation into a non-linear
positivity transformation, $\phi(\mathbf{x}) = e^{\mathbf{x}}$, followed by
an $l_{1}$-norm scaling:
\begin{equation}
    \begin{aligned}
        \operatorname{Softmax}(\mathbf{x}) & = \frac{e^{\mathbf{x}}}{\sum_{j}e^{x_j}} = \frac{e^{\mathbf{x}}}{\|e^{\mathbf{x}}\|_{1}} = \frac{\phi(\mathbf{x})}{\|\phi(\mathbf{x})\|_{1}} \label{equ:softmax}
    \end{aligned}
\end{equation}
This allows us to generalise attention as
shown in \cref{equ:newatt}.
\begin{equation}
    \begin{aligned}
        \mathbf{A}     & = \phi\left(\frac{\mathbf{Q} \mathbf{K}^T}{\sqrt{d}}\right) \otimes \mathbf{M'}\label{equ:newatt} \\
        \mathbf{A}_{i} & \leftarrow \frac{\mathbf{A}_i}{\|\mathbf{A}_i\|_1}
    \end{aligned}
\end{equation}
The subscripts $i$ and $j$ represent the row and
column indices respectively. $\otimes$ denotes element-wise multiplication.
The mask $\mathbf{M'}$ is constructed with ones on and below the diagonal
and zeros above it. It is evident that the original attention function (\cref{equ:rowatt})
is a special case of this general form where $\phi(\mathbf{x}) = e^{\mathbf{x}}$.

\subsubsection{Length Scaled Softplus Attention}
\label{sec:lengthscaled}

Based on the general form of \cref{equ:newatt}, we now introduce \textit{Length
    Scaled Softplus Attention} (LSSA), a novel attention variant designed to
enhance the scalability and performance of LLMs. LSSA replaces the exponential
function with $\operatorname{Softplus}(x)=\log(1+e^x)$ to introduce non-linearity
while maintaining smooth gradients essential for stable training dynamics. Empirical
testing of several widely-used activation functions in place of $\phi$ in \cref{equ:newatt},
revealed that Softplus delivers the best performance (see \cref{tab:activationcomp}),
justifying its adoption. As described below, LSSA also introduces a novel scaling
factor that accounts for both the sequence length and the model's dimensionality,
thereby addressing limitations associated with traditional attention methods
in handling long sequences during inference.

In contrast to scaled dot-product attention, cosine similarity attention employs
the $l_2$-norm for each row of $\mathbf{Q}$ and $\mathbf{K}$. This approach has
been shown to produce more moderate attention weights, which can enhance
performance across various tasks \citep{henry2020query, houlsby2023scaling, liu2022swin}. Furthermore,
the $l_{2}$-norm restricts the dot product values to the interval $[-1, 1]$,
effectively placing them in the region where the derivative of the Softplus function,
the Sigmoid $\sigma(x)=1/(1+e^{-x})$, exhibits a steep slope. This characteristic ensures that the
gradients remain distinct across different inputs, rendering the attention mechanism
acutely sensitive to small variations in the latent representations.

However, in high-dimensional spaces, as the number of dimensions increases,
two randomly selected vectors are likely to become orthogonal. This phenomenon
causes the elements of the product of $\sfrac{\mathbf{Q}_i}{\|\mathbf{Q}_i\|_2}$
and $\sfrac{\mathbf{K}_i^T}{\|\mathbf{K}_i\|_2^T}$ to approach zero, thereby
compressing the dynamic range of the attention scores. Consequently, a scale
factor becomes essential for cosine similarity attention to restore
discriminability, and the factor should be associated with the dimensionality
$d$.

Previous work demonstrated that replacing the traditional scaling factor $\sfrac
    {1}{\sqrt{d}}$ with $\sfrac{\log{L}}{\sqrt{d}}$ in scaled dot-product attention
enhances the length extrapolation capabilities of transformers \citep{chiang2022overcoming,nakanishi2025scalable}.
Furthermore, \citet{su2021viewing} highlighted that the inclusion of the $\log
    {L}$ factor aids in maintaining entropy invariance with different token length,
thereby facilitating better extrapolation to unknown sequence lengths. We extend
this concept by introducing a dynamic length scale factor that adapts to the
varying number of attended tokens in each row of the attention matrix.
Specifically, we set the scaling factor to $\log{d}\log{\mathbf{N}}$, where
$\mathbf{N}$ is an $L \times L$ matrix where each element in row $i$ is
equal to $i$ (the number of tokens attended to in that row). This ensures the
attention mechanism remains robust across varying sequence lengths. The
formulation of LSSA is mathematically defined in \cref{equ:lengthscaleatt}.
\begin{equation}
    \begin{aligned}
        \mathbf{Q}_{i} & \leftarrow\frac{\mathbf{Q}_i}{\|\mathbf{Q}_i\|_2}, \space \mathbf{K}_{i}\leftarrow\frac{\mathbf{K}_i}{\|\mathbf{K}_i\|_2}                     \\
        \mathbf{A}     & = \operatorname{Softplus}\left((\log{d}\log{\mathbf{N}})\otimes \mathbf{Q}\mathbf{K}^{T}\right) \otimes \mathbf{M'}\label{equ:lengthscaleatt} \\
        \mathbf{A}_{i} & \leftarrow \frac{\mathbf{A}_i}{\|\mathbf{A}_i\|_1}
    \end{aligned}
\end{equation}
It is important to note that due to hardware limitations, the proposed LSSA was
evaluated with a sequence length of $L=1024$ and dimension $d=64$. For LLMs
trained with longer sequence lengths and higher dimensionalities, the scaling
factor may require further adjustment. However, the component $\log{\mathbf{N}}$
should remain unchanged, as it is crucial to maintain the length
extrapolation capability of the model.

\subsection{The Sharpening Stage}\label{sec:sharpening_stage}
\subsubsection{Rationale for Two-Stage Attention}
\label{sec:design}

While the normalisation stage provides a stable attention distribution, it
fundamentally retains the dense nature of Softmax, activating all input tokens
regardless of their relevance. This counter-intuitive behaviour exacerbates
the attention smoothing problem \citep{chiang2022overcoming,velivckovic2024softmax},
where attention scores lack distinct peaks as the sequence length increases
during inference, hindering the model's ability to distinguish signal from noise.

A naive solution is to replace $\phi$ in \cref{equ:newatt} with a sparsity-inducing
activation function, such as ReLU. However, as shown in
\cref{tab:activationcomp}, ReLU-based attention results in performance degradation.
In ReLU-based attention, positive values are predominantly
concentrated around the diagonal, while off-diagonal positions often remain negative
and are suppressed to zero. This creates a `dead neuron' phenomenon during
training: once a token's score falls below zero, it loses all gradient
feedback, effectively isolating it from the optimisation process and restricting
the model's capacity to capture long-range dependencies.

Fundamentally, the efficacy of the attention mechanism stems from the
synergy between two distinct functional components: a consistent non-linear transformation
and a subsequent normalisation. Functions such as Softplus and $e^{x}$ transform
raw scores while keeping every token's contribution non-zero. Unlike ReLU-based
mappings, they do not collapse any admissible input to exactly zero before
normalisation. This property ensures that every token remains eligible to
participate in the subsequent $l_{1}$-norm competition. The $l_{1}$-norm then serves two critical roles that are essential
for stable training. In the forward pass, it functions as a \textit{weighted
    sum} operator (ensuring $\sum |A_{ij}| = 1$), which maintains the scale stability
of the output representations regardless of the input magnitude or sequence length.
In the backward pass, the $l_{1}$-norm introduces a mechanism of \textit{lateral
    inhibition} via gradient coupling (see biological justification in Appendix~\ref{app:neuroscience}). By linking all tokens through the denominator
term, the optimisation of `winner' tokens propagates gradients back to
suppress `losers'. This global coupling guarantees that even currently suppressed
tokens receive gradient feedback, thereby preventing the dead neuron issue
associated with hard thresholding.

\subsubsection{Attention Re-weighting}
\label{sec:lengthgate}
The theoretical insight reframes both standard Softmax and our LSSA as a
unified \textbf{Normalisation Stage}, which establishes a numerically stable
and gradient-connected landscape. It is upon this robust foundation that we introduce
our re-weighting mechanism, defined in \cref{equ:reweighting}, as a distinct
\textbf{Sharpening Stage}. By applying the shift and re-weighting operations
\textit{after} the gradient pathways have been secured by the normalisation
stage, the proposed method achieves the desired sparsity and noise filtering
without sacrificing trainability.
\begin{equation}
    \begin{aligned}
        \mathbf{A}     & \leftarrow \operatorname{ReLU}^{p}\left(\mathbf{A}\otimes \mathbf{N}-\mathbf{O}\right)\label{equ:reweighting} \\
        \mathbf{A}_{i} & \leftarrow \frac{\mathbf{A}_i}{\|\mathbf{A}_i\|_1}
    \end{aligned}
\end{equation}

Here, the normalised scores $\mathbf{A}$ are first scaled by the token count
$\mathbf{N}$ and then shifted by a matrix $\mathbf{O}$ (an offset matrix of
ones, with zeros in the first three rows to prevent instability). This centres
the distribution around zero. The $\operatorname{ReLU}^{p}$ function then
masks scores below zero and sharpens the remaining positive scores by
raising them to the power of a hyper-parameter $p$. A final scaling by the $l_{1}$-norm ensures
the output is a valid probability distribution. The re-weighting mechanism
is an additional component of our proposed attention layer, we refer to the combination
of LSSA (stage 1) and this re-weighting mechanism (stage 2) as LSSAR in the rest
of this paper. The complete LSSAR algorithm is provided in Algorithm~\ref{alg:lssar_detailed} in Appendix~\ref{app:algorithm}.

This power operation is key to solving the attention smoothing problem. Let
$x_{1}, x_{2}, \ldots, x_{n}$ be the positive elements in one row of the ReLU
output from \cref{equ:reweighting}, and let $M = \max_{1 \le k \le n} x_{k}$. The
re-weighted value of the $j$-th entry is
\begin{equation}
    \overline{x}_{j} = \frac{x_j^p}{\sum_{k=1}^{n} x_k^p}.
\end{equation}
If $x_j < M$, then
\begin{equation}
    \overline{x}_{j}
    = \frac{(x_j/M)^p}{\sum_{k=1}^{n} (x_k/M)^p}
    \to 0
    \quad \text{as } p \to \infty.
\end{equation}
If $t$ entries attain the maximum value $M$, then each of those entries
converges to $1/t$ as $p \to \infty$. In the generic case where the maximum is
unique, this reduces to a one-hot limit on the largest entry. The tied-maximum
case is a boundary case, in which the limiting mass is shared uniformly among
the tied maxima. This limit shows that increasing $p$ suppresses smaller values
while concentrating mass on the largest surviving entries after the ReLU shift.
This property ensures that the attention mechanism maintains sharp, distinct
peaks even with longer sequences, thereby preserving the model's ability to
focus on the most relevant tokens. This offers distinct advantages over
projection-based sparse mechanisms (e.g., Sparsemax \citep{martins2016softmax}),
particularly in terms of computational efficiency and hardware compatibility, as
discussed in Appendix~\ref{app:sparsemax_comparison}.
From a thermodynamic perspective (see Appendix~\ref{app:log_space}), this process can
be viewed as active entropy minimisation, with $p$ acting as an inverse
temperature coefficient. Furthermore, geometric analysis (Appendix~\ref{app:graph_theory})
suggests that LSSAR prevents `manifold drift' by anchoring representation updates
to a local semantic neighbourhood.
\begin{table*}[tb]
    \centering
    \small
    \caption{Comparison of Softmax-attention and state-of-the-art Softmax-free attention
        mechanisms across different sequence lengths. Bold font indicates the best
        result for each sequence length. }
    \begin{tabular}{lcccc}
        \toprule
        Attention Mechanism                                                    & 1K              & 2K              & 4K              & 8K              \\
        \midrule
        \multicolumn{5}{c}{Base attention mechanisms (no re-weighting)}                                                                    \\
        Softmax                                                                & 3.1911          & 4.1662          & 5.4513          & 6.2823          \\
        SSMax \citep{nakanishi2025scalable}                 & \textbf{3.1705} & 4.0915          & 5.4606          & 6.4384          \\
        Sigmoid \citep{ramapuram2024theory}                                    & 3.1935          & 7.4554          & 11.8355         & 14.4995         \\
        Sigmoid \citep{ramapuram2024theory}($b=-\log{\mathbf{N}}$)             & 3.1930          & 6.5830          & 8.4679          & 9.5939          \\
        Sigmoid \citep{ramapuram2024theory}($b=-\log{\mathbf{N}}, l_{1}$-norm) & 3.1849          & 4.3470          & 5.5544          & 6.1846          \\
        Sigmoid (\cref{tab:activationcomp})                                    & 3.2000          & 4.3811          & 5.8465          & 6.5701          \\
        ReLU \citep{wortsman2023replacing}                                     & 3.2143          & 6.2662          & 8.4982          & 10.3460         \\
        ReLU \citep{li2022robust}                                              & 3.2006          & 4.5192          & 5.6924          & 6.4561          \\
        ReLU \citep{shen2023study}                                             & 3.2155          & 6.5573          & 8.9072          & 10.7266         \\
        LSSA                                                                   & 3.1905          & 4.1301          & 5.2960          & 5.9403          \\
        \midrule
        \multicolumn{5}{c}{Re-weighted variants ($p=3$)}                                                                                    \\
        Softmax                                                                & 3.1879          & 4.0277          & 5.2842          & 6.2339          \\
        Sigmoid \citep{ramapuram2024theory}($b=-\log{\mathbf{N}}, l_{1}$-norm) & 3.1841          & 3.7387          & 4.9286          & 5.7450          \\
        LSSAR                                                                  & 3.1782 & 4.2383          & 5.4056          & 6.3007          \\
        \midrule
        \multicolumn{5}{c}{Re-weighted variants ($p=15$)}                                                                                   \\
        Softmax                                                                & 5.3878          & 5.9491          & 6.5276          & 7.0183          \\
        Sigmoid \citep{ramapuram2024theory}($b=-\log{\mathbf{N}}, l_{1}$-norm) & 3.2171          & 3.3499          & 3.6108          & 3.8587          \\
        LSSAR                                                                  & 3.1905          & \textbf{3.1930} & \textbf{3.2291} & \textbf{3.3171} \\
        \bottomrule
    \end{tabular}

    \label{tab:softmaxfreeatt}
\end{table*}

\section{Experiments}
\label{sec:experiment} The experiments were conducted using 8 NVIDIA A100
80GB GPUs. We utilised the GPT-2 small architecture \citep[124 million parameters;][]{radford2019language},
replacing the original absolute position embeddings with Rotary Position Embeddings
\citep[RoPE;][]{su2024roformer}.

All models were trained using a sequence length of 1024 on the FineWeb-10B
dataset \citep{penedo2024fine}, which consists of 10.2 billion training
tokens distributed across 18,865 training steps, along with 0.1 billion validation
tokens. Full details of the optimisation setup, batch construction, and numerical
precision are provided in Appendix~\ref{app:training_config}.

Due to page limitations, we provide extensive supplementary experiments and
analysis in Appendix~\ref{app:further_experiments}. These include: (1) a decomposition
study of the Softmax function (\cref{tab:decomsoft}), which empirically
confirms the central role of $l_1$-normalisation; (2) a comparison of different
activation functions (\cref{tab:activationcomp}), demonstrating that Softplus
performs best among the tested alternatives; (3) additional LSSAR+PI experiments in Appendix~\ref{app:PI}, showing that LSSAR
can be combined with position interpolation; (4) an analysis of the
re-weighting parameter $p$ in Appendix~\ref{sect:varyp} and an adaptive
position-dependent $p_i$ schedule in Appendix~\ref{app:adaptive_p}; and (5) computational analysis and scaling experiments with
filtered data in Appendix~\ref{app:comp_analysis}
and Appendix~\ref{app:scaling_experiments}, respectively.

\subsection{Comparison with Softmax-Free Attention Methods}
We compared the proposed LSSA and LSSAR with leading Softmax-free methods using the same GPT-2-124M model. These attention functions include two that use
Sigmoid (one proposed in \citet{ramapuram2024theory} and the other produced when using Sigmoid for $\phi$ in \cref{equ:newatt} as used in \cref{tab:activationcomp}), and three ReLU-based
attention methods from \citep{wortsman2023replacing,li2022robust,shen2023study}.
The original Sigmoid attention \citep{ramapuram2024theory} computes attention scores as $\text{Sigmoid}(x - b)$, where the bias term $b = -\log{L}$ shifts the input distribution to prevent saturation. We evaluated two
additional variants. The first variant alters the hyperparameter $b$ from $-\log
    {L}$ to $-\log{\mathbf{N}}$. The second variant maintains this modification
while subsequently applying the $l_{1}$-norm to each row of the attention matrix, ensuring each row sums to one. We additionally include Scalable-Softmax (SSMax)
\citep{nakanishi2025scalable} as a direct baseline for methods that modify the
effective attention temperature. SSMax is trained under the same setting, with
its learnable scaling parameter initialised as $s=0.43$. The experimental findings
are detailed in \cref{tab:softmaxfreeatt}.

At an inference sequence length of 1K, all attention variants perform
similarly, with only slight differences in validation loss. Specifically, SSMax
achieves the lowest validation loss, while Sigmoid
\citep{ramapuram2024theory}($b=-\log{\mathbf{N}}, l_{1}$-norm) and LSSA also
perform slightly better than the standard Softmax attention. As sequence length
increases, performance differences become more evident. The standard Softmax
attention experiences performance decline with longer sequences but remains
relatively stable compared to most other methods.

Sigmoid attention \citep{ramapuram2024theory} shows significant performance decline at longer sequences, with validation loss rising sharply at 8K tokens. Theoretically, this method employs a shifted activation Sigmoid$(x-b)$, where the bias term $b$ effectively shifts the input distribution leftwards. This is designed to prevent the inputs from falling into the right-hand saturation region of the Sigmoid function, thereby maintaining sensitivity to large scores. However, our results indicate that this element-wise shift alone is insufficient. The modified Sigmoid \citep{ramapuram2024theory}($b=-\log{\mathbf{N}}$) shows some improvement, suggesting that the proposed hyperparameter $\log{\mathbf{N}}$ is more effective than $\log{L}$, yet the model still faces challenges with longer sequences. Crucially, incorporating the $l_{1}$-norm in Sigmoid \citep{ramapuram2024theory}($b=-\log{\mathbf{N}}, l_{1}$-norm) dramatically recovers performance, aligning
it closely with standard Softmax attention, underscoring the importance of $l_{1}$-norm in attention mechanisms.

ReLU-based attention variants \citep{wortsman2023replacing,li2022robust,shen2023study}
also experience notable performance decline with longer sequences, though not
as drastically as the basic Sigmoid attention. This indicates that while
ReLU-based methods may be effective for shorter sequences, they might not be
ideal for managing longer-range dependencies. We conjecture that this degradation may be attributable to the `dead neuron' phenomenon: as sequence length increases, a larger proportion of off-diagonal attention scores fall below zero and are suppressed, potentially severing gradient pathways for distant tokens. A theoretical analysis of this mechanism is provided in \cref{sec:design}.

To demonstrate the effectiveness of the proposed re-weighting mechanism, the
attention scores generated by Softmax, Sigmoid \citep{ramapuram2024theory}($b
    =-\log{\mathbf{N}}, l_{1}$-norm), and LSSA were re-weighted with $p=3$ and
$p =15$. As shown in \cref{tab:softmaxfreeatt}, the length extrapolation
ability provided by the re-weighting can improve the performance of all the
tested attention mechanisms. For Softmax, performance was improved slightly
using $p=3$, but was harmed when using $p=15$. For the other attention
mechanisms, $p=3$ was most effective for sequence lengths of 1K, while
$p=15$ was best for the longer sequences. For $p=15$, LSSAR achieves the best
long-context performance among the tested variants, with validation loss
remaining comparatively flat up to 8K tokens. The gradient analysis in
Appendix~\ref{app:gradient_analysis} explains this contrast: under stronger
sharpening, Softmax exhibits exponential gradient decay in the high-confidence
regime, whereas LSSAR exhibits polynomial gradient decay and therefore retains
a more effective optimisation signal. This quantitative robustness is further corroborated
by the qualitative attention map visualisations in Appendix~\ref{app:attmap}
(see \cref{fig:attmap}), where LSSAR displays a more balanced distribution
across tokens with sharper, non-collapsing peaks and markedly reduced attention
sink compared to standard Softmax, confirming its enhanced ability to
preserve salient long-range dependencies.

The above experiments use standard RoPE without modifying the positional
encoding. Since attention re-weighting and positional interpolation operate on
different parts of the Transformer, they can also be combined. We therefore
include additional LSSAR+PI experiments in Appendix~\ref{app:PI}, showing that
LSSAR remains effective under continued long-context finetuning with position
interpolation.

\begin{figure}[tb]
    \centering
    \includegraphics[width=\linewidth]{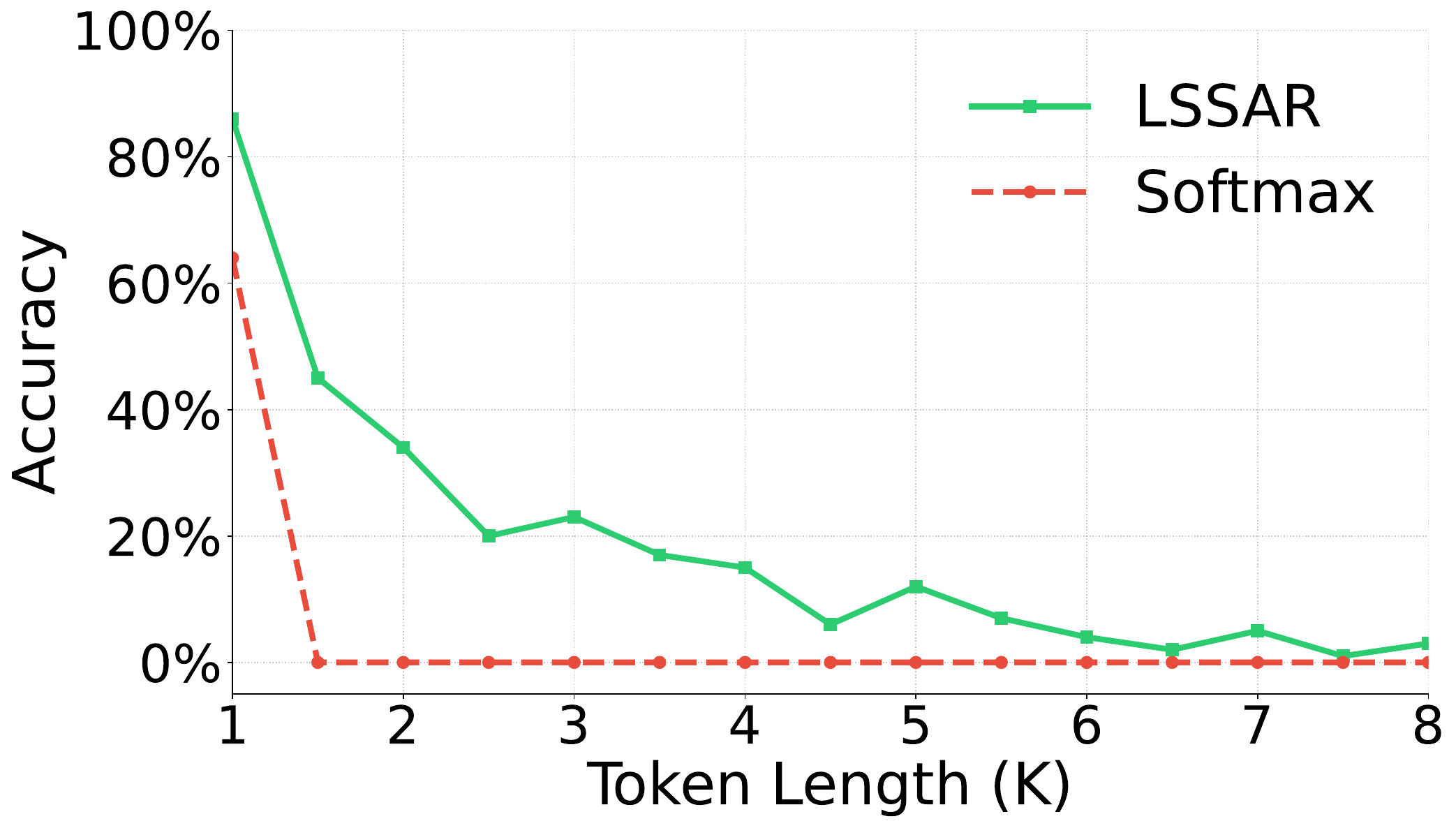}
    \caption{Passkey retrieval accuracy for LSSAR ($p=15$) and standard
        Softmax attention. Accuracy is averaged over 100 trials with the passkey
        placed at random positions within the sequence.}
    \label{fig:passkey_accuracy}
\end{figure}

\begin{table*}
    [tb]
    \centering
    \caption{Zero-shot performance of the models with Softmax attention and
        LSSAR on downstream tasks. The best scores in each column indicated in bold.}
    \begin{tabular}{lccccccc}
        \toprule         & ARC-E          & ARC-C          & HellaSwag      & PIQA           & MMLU  & SciQ          & SummScreen     \\
        \midrule Softmax & 39.77          & \textbf{23.72} & 32.42          & 64.09          & 22.97 & 60.6          & 1.682          \\
        LSSAR ($p=15$)   & \textbf{40.57} & 22.61          & \textbf{33.03} & \textbf{65.34} & 22.97 & \textbf{62.1} & \textbf{6.309} \\
        \bottomrule
    \end{tabular}
    \label{tab:downstream}
\end{table*}

\begin{table}[tb]
    \centering
    \caption{Symbolic regression results comparing different models. Since the
        solar mass is fixed in this setting, the fitted constant corresponds to
        $Gm_{\text{sun}}$.}\tabcolsep0.5pt
    \begin{tabular}{lcc}
        \toprule
        Model         & Recovered Function                                                                                         & Fitted Constant        \\
        \midrule
        Oracle (KNN)  & $F \propto \frac{m_1}{r^2}$                                                                                & 25.85                  \\[3pt]
        GPT (Softmax) & $F \propto \frac{\left( \sin \left( \frac{1}{\sin(r - 0.24)} \right) + 1.45 \right)}{\frac{1}{r} + m_2}  $ & N/A                    \\[7pt]
        GPT (LSSAR)   & $F \propto \frac{m_1}{r^2}$                                                                                & 27.02                  \\
        \midrule
        Ground Truth  & $F \propto \frac{m_1}{r^2}$                                                                                & $4\pi^2 \approx 39.48$ \\
        \bottomrule
    \end{tabular}
    \label{tab:symbolic_regression}
\end{table}

\subsection{Long Context Passkey Retrieval}
To offer a more direct and challenging evaluation of length extrapolation,
we employed the long-context passkey retrieval task \citep{mohtashami2023random}.
This needle-in-a-haystack test is specifically designed to assess a model's
ability to identify a single, crucial, piece of information within a long
and distracting context.

We compared LSSAR ($p=15$) against the standard Softmax attention baseline. The
results, shown in \cref{fig:passkey_accuracy}, reveal a critical failure in the
baseline model. While Softmax attention achieves 64\% accuracy within its training
length, its performance catastrophically collapses to 0\% for all tested lengths
of 1.5K tokens and beyond. This demonstrates its inability to overcome attention
smoothing, effectively losing the passkey as the context grows. In stark contrast,
LSSAR not only achieves a much higher accuracy of 86\% at the 1K training length
but also demonstrates remarkable robustness in extrapolation. Its accuracy
degrades gracefully from 45\% at 1.5K tokens, 20\% at 4K tokens, and it maintains
non-zero accuracy even at 8K tokens. This sustained, non-zero performance provides
clear evidence that the re-weighting mechanism successfully sharpens the model's
focus, allowing it to pinpoint the relevant information even in sequences
far exceeding its training context and overcoming a critical limitation of
standard Softmax attention. Complementary passkey results for LSSAR combined
with Position Interpolation are reported in \cref{tab:lssar_pi_passkey} in
Appendix~\ref{app:PI}. Additional experiments on passkey retrieval with
different model scales and filtered FineWeb-Edu data are provided in
\cref{fig:edu_passkey} in Appendix~\ref{app:scaling_experiments}.

\subsection{Downstream Evaluation}
\label{sec:ablation}
To demonstrate the practical efficacy of LSSAR, we evaluate its zero-shot
performance against Softmax attention on six standard benchmarks: ARC \citep{clark2018think},
HellaSwag \citep{zellers2019hellaswag}, PIQA \citep{bisk2020piqa}, MMLU \citep{hendrycksmeasuring},
SciQ \citep{welbl2017crowdsourcing}, and SummScreen \citep{chen2021summscreen},
using the \texttt{lm-evaluation-harness} framework \citep{eval-harness}. For
ARC-E, ARC-C, HellaSwag, PIQA, and SciQ, we report normalised accuracy; for
MMLU, we use standard accuracy; and for SummScreen, we adopt the ROUGE-1 score
to measure summarisation quality.

LSSAR outperforms Softmax attention in five of these tasks, and has
performance competitive with that of Softmax attention in the other two (\cref{tab:downstream}).
Notably, LSSAR excels on the long-context summarisation benchmark, SummScreen,
achieving a nearly fourfold improvement score over Softmax attention. This
highlights LSSAR’s ability to synthesize information over sequences of up to
2K tokens, aligning with its design for robust length extrapolation. Further
downstream evaluation results on different model scales can be found in
Appendix~\ref{app:scaling_experiments}. Additional analysis on the LAMBADA benchmark, which tests long-range contextual understanding, is provided in Appendix~\ref{app:lambada}.

\subsection{Symbolic Regression on Physical Structure}
\label{sec:symbolic_regression}

To evaluate whether LSSAR helps models capture physically meaningful structure,
we conducted symbolic regression experiments following the force-prediction
protocol of \citet{vafa2025inductive}. We trained a 109M parameter GPT-2 model
on 10M simulated orbital trajectory sequences, replacing the standard Softmax
attention with our proposed LSSAR mechanism, while keeping all other
architectural and training hyperparameters unchanged. The model was pretrained
to predict planetary positions and subsequently fine-tuned on a force prediction
task, where the output was the gravitational force magnitude $F$ given the
planetary state (mass $m_1$, solar mass $m_2$, and distance $r$).

Following fine-tuning, PySR \citep{cranmer2023interpretable} was used to perform
symbolic regression on the model-predicted force values, rather than on
ground-truth equations, searching for interpretable mathematical expressions
that best fit the model's predictions. The true gravitational law follows
Newton's formula:
\begin{equation}
    F = G \cdot \frac{m_1 \cdot m_2}{r^2}
\end{equation}
where $m_2=m_{\text{sun}}$ is fixed in the solar-system setting, so
$G \cdot m_2=G \cdot m_{\text{sun}}=4\pi^2 \approx 39.48$ in astronomical
units.

As shown in \cref{fig:force_earth}, LSSAR generates force vectors that are more
consistent with the Newtonian directionality than the Softmax baseline.
Symbolic regression identifies an inverse-square form,
$F = 27.02 \times m_1/r^2$, which captures the distance dependence and linear
mass dependence. Recovering this inverse-square structure is the central
evidence that LSSAR helps the model infer physically meaningful relations from
orbital trajectories. In contrast, standard GPT fails to match the physical
structure, producing nonsensical equations
\citep{vafa2025inductive}.

As shown in \cref{tab:symbolic_regression}, LSSAR's fitted constant (27.02) is
closer to the theoretical value ($4\pi^2 \approx 39.48$) than the Oracle (KNN)
baseline with direct access to true force labels (25.85). The fitted constant
still differs from the theoretical value by approximately 31.6\%, which may be
affected by data noise, optimisation randomness, model capacity, and the limited
force-prediction finetuning setup. We therefore view the coefficient mismatch as
a limitation of numerical calibration, while the recovered inverse-square
dependence remains the key evidence of physical-structure recovery. The Oracle
baseline learns to interpolate from true gravitational force labels, whereas
LSSAR observes orbital trajectory sequences during pretraining and is fine-tuned
on force prediction. This comparison suggests that the LSSAR attention mechanism
can provide a useful inductive bias for recovering compact physical structure.

To further contextualise these results, \citet{vafa2025inductive} also tested
state-of-the-art large language models, including o3, Claude 4 Sonnet, and
Gemini 2.5 Pro, on the same force-prediction task using in-context learning.
Those models did not recover the inverse-square relation, producing trivial
expressions such as $F \propto m_1$ or $F \propto 1/(m_2 - 0.50)$.

Overall, the LSSAR-based model yields an interpretable inverse-square form,
whereas the Softmax baseline and the in-context LLM baselines fail to match this
physical structure. These results indicate that LSSAR's attention mechanism
induces inductive biases aligned with compact physical structure in this
controlled setting.  A mechanistic interpretation of
why LSSAR may improve symbolic-regression outcomes is provided in
Appendix~\ref{app:depth_physical_law}.
Visualisations of force
predictions for other planets in the solar system are provided in
Appendix~\ref{app:symbolic_regression_planets}.

\section{Discussion}
\label{sec:discussion} Due to limited computing resources, our main
experiments evaluate LSSAR on GPT-2-124M models trained at a sequence length of
1024 and tested up to 16K tokens. The results in \cref{fig:attcompare} show that
LSSAR maintains a relatively stable validation loss over this extrapolation
range, suggesting that the proposed attention mechanism improves the intrinsic
length-generalisation behaviour of the Transformer. Additional experiments on
45M and 355M models are provided in Appendix~\ref{app:scaling_experiments},
where LSSAR shows consistent improvements over the Softmax baseline across the
tested model sizes. Furthermore, we discuss implications for reasoning models
in Appendix~\ref{app:reasoning}.

While our PyTorch implementation incurs substantial computational overhead,
this should be treated as a current implementation constraint rather than a
settled efficiency result. In the current implementation, LSSAR is noticeably
slower at inference and uses substantially more memory during training than
standard attention. Our analysis in Appendix~\ref{app:comp_analysis} suggests
that LSSAR may be amenable to future fused implementations, but we do not yet
provide an optimised kernel or an efficiency result comparable to standard
high-performance attention implementations.

\section{Limitations}
We emphasise several limitations of the present study. First, the empirical
evidence currently comes from models in the 45M to 355M parameter range and
from training runs that are much shorter than those used to train
production-scale language models. The present results should therefore be
interpreted as mechanism-level evidence rather than as definitive large-scale
validation. 

Second, the principal extrapolation experiments train at sequence length 1024
and evaluate zero-shot at 2K, 4K, 8K, and 16K. This setting is useful for
isolating intrinsic length generalisation, but it does not correspond to a
production training pipeline in which long-context examples are gradually
introduced during training or in which the model is further adapted at the
target context length. The reported zero-shot results should thus be read as
controlled extrapolation measurements rather than as a complete long-context
recipe.

Third, LSSAR remains an exact $O(L^2)$ attention mechanism rather than a
linear-attention or state-space alternative. Its present implementation is
therefore intended to test the attention mechanism itself, not to provide a
drop-in efficiency improvement over highly optimised attention kernels.

Fourth, the stability of LSSAR relative to standard Softmax attention in much
deeper networks remains untested, especially because Softplus and its
derivative can saturate in certain regimes. Larger-scale validation,
deeper-network analysis, and kernel-level optimisation therefore remain
important directions for future work.

\section{Related Work}
\textbf{Softmax-free Attention.} Previous research has investigated Softmax-free
attention by substituting the Softmax function with the ReLU activation
\citep{li2022robust, shen2023study, wortsman2023replacing, hron2020infinite, bai2024transformers, fu2024can},
SquaredReLU activation \citep{hua2022transformer}, and Sigmoid activation
\citep{ramapuram2024theory}, as well as examining purely linear attention
\citep{zhen2022cosformer, tsai2019transformer, katharopoulos2020transformers, han2023flatten, arora2024simple, lu2021soft}.
However, none of these approaches outperform the original Softmax attention.
In contrast, the proposed LSSA demonstrates enhanced numerical stability and
superior performance across various token lengths.

It is also important to distinguish our work from two other lines of research
for long-context modeling. The first improves computational efficiency by
approximating the attention matrix, leading to linear attention mechanisms such
as Performer \citep{choromanski2020rethinking}. The second line of research proposes
entirely new architectures, such as introducing sparse attention patterns as
in LongFormer \citep{beltagy2020longformer}, or replacing the attention
paradigm altogether with State Space Models like Mamba \citep{gu2024mamba}.
In contrast to these approaches, LSSAR remains within the exact, quadratic-time,
attention paradigm. Our contribution is not a new architecture or an
approximation, but a direct, drop-in replacement for the standard attention
mechanism designed to fundamentally improve its numerical stability and length
extrapolation capabilities within the conventional Transformer framework.

Among these linear-attention variants, FlattenTransformer is particularly
relevant because it also uses amplification to focus attention weights
\citep{han2023flatten}. The mechanism, however, is different: FlattenTransformer
amplifies query-key features before applying a linear attention approximation,
whereas LSSAR sharpens the exact pairwise attention scores after the first-stage
mapping and row-wise normalization. This distinction motivates our focus on
attention re-weighting in the exact attention space.

\textbf{Attention Re-weighting.} The non-linear re-weighting mechanism
introduced by softmax attention ($l_{1}$-normalization) has been shown to
concentrate the distribution of attention weights, thereby stabilising the training
process \citep{aueb2016one,gao2017properties,jang2016categorical,zhen2022cosformer}.
Our empirical findings further demonstrate its essential role in maintaining
the performance of LLMs. Moreover, we introduce a novel perspective: a non-linear
positivity transformation followed by $l_{1}$-normalization.
Inspired by the classic normalization-ReLU structure, the proposed re-weighting
mechanism masks less relevant tokens and amplifies the relevant ones, which boosts
the length extrapolation ability of underlying models. This provides the
deep learning community with a deeper understanding of the attention
mechanism within transformers.

\textbf{Length Extrapolation.} Positional embeddings \citep{su2024roformer,chen2023extending,chi2022kerple,kiyono2021shape,golovneva2024contextual,he2024two,huang2020improve,li2023functional,likhomanenko2021cape,liu2023scaling,wang2024resonance,zheng2024cape}
play a vital role in transformer architectures by providing essential
information about the positions of tokens within a sequence, which is considered
a key factor in enhancing length extrapolation \citep{kazemnejad2023impact}.
Among these embeddings, RoPE is particularly noteworthy; it forms the foundation
of many modern LLMs, including GPT-4 \citep{achiam2023gpt}, Llama3 \citep{dubey2024llama},
Deepseek-v3 \citep{liu2024deepseek}, DeepSeek-R1 \citep{deepseekai2025deepseekr1}
and Qwen3 \citep{yang2025qwen3}. Unlike RoPE-based extrapolation techniques that
compensate for sequence length limitations through positional embedding adjustments
\citep{chen2023extending,li2023functional,kaiokendev2023things,bloc97_2023_ntk,bloc97_2023_correction,emozilla2023dynamic},
LSSAR redefines the core attention mechanism itself. The re-weighting
operation is not merely an auxiliary technique but a structural enhancement to the
attention computation, ensuring sharp peaks and stable gradients by design.
This indicates that a well-designed attention mechanism like LSSAR is
essential for RoPE-based LLMs to enhance their length extrapolation, suggesting
that LSSAR may serve as a general strategy to bolster the length extrapolation
abilities of most contemporary LLMs.

\section{Conclusion}
This paper introduced two novel improvements to attention mechanisms. The
first normalizes scores using LSSA, a mechanism built on the Softplus function
and a dynamic length scale factor, which outperforms standard Softmax. The second
applies a unique re-weighting mechanism that sharpens the attention distribution.
The combined model, LSSAR, demonstrates a remarkable ability to extrapolate
to longer sequences while maintaining numerical stability and mitigating the
attention sink phenomenon. Downstream evaluations confirm that these architectural
improvements translate into superior performance on a range of tasks.

Beyond conventional benchmarks, our symbolic regression experiments reveal a
deeper insight: LSSAR enables models to recover the inverse-square structure of
Newtonian gravity from orbital trajectories, while standard Softmax attention
fails to match this physical structure. This demonstrates that appropriate attention
mechanisms are crucial for foundation models to develop physically grounded world
models.

\ificmlshowauthors
    \section*{Acknowledgements}

    We acknowledge financial support from the Special Project of Nanyang Normal University (No. 2024ZX033) and the Natural Science Foundation of Henan (Nos. 252300420979 and 262300422563).

    We also thank King's College London for the use of the CREATE research computing facility~\cite{CREATE}. Furthermore, the authors acknowledge the use of resources provided by the Isambard-AI National AI Research Resource (AIRR). Isambard-AI is operated by the University of Bristol and is funded by the UK Government’s Department for Science, Innovation and Technology (DSIT) via UK Research and Innovation; and the Science and Technology Facilities Council [ST/AIRR/I-A-I/1023]~\cite{mcintosh2024isambard}. Access to these resources was granted under the AIRR Gateway project (Project ID: BYYG-VXGF-P).
\fi

\section*{Impact Statement}

This paper presents LSSAR, a novel attention mechanism designed to improve the numerical stability and length extrapolation capabilities of large language models. As a contribution to foundational models, our work has broad implications for the field of machine learning.

\textbf{Positive Impacts.} LSSAR's improved length extrapolation enables more efficient processing of long documents, potentially reducing computational costs and energy consumption compared to methods that require retraining on longer sequences. The enhanced numerical stability may also improve the reliability of LLM deployments in safety-critical applications. Furthermore, our symbolic regression experiments suggest that LSSAR can help models learn more physically grounded representations, which may be useful for scientific modelling applications.

Beyond these direct benefits, LSSAR implements a \textit{coarse-to-fine causal filtering} mechanism (see Appendix~\ref{app:depth_physical_law} for theoretical details): the Shift-ReLU stage performs coarse-grained noise elimination by categorically removing tokens below the population average, while the subsequent WTA dynamics perform fine-grained causal selection among surviving candidates. This hierarchical filtering has two promising implications. First, as discussed in Appendix~\ref{app:reasoning}, it is particularly relevant for reasoning models that rely on Chain-of-Thought (CoT) processes. By progressively distilling relevant premises from noise, LSSAR may reduce the redundant self-verification loops observed in current reasoning systems, potentially compacting reasoning chains and improving token efficiency. Second, this capability opens avenues for AI safety research: (a) the coarse filtering stage may enhance robustness against adversarial attacks that exploit attention dispersion by eliminating obviously irrelevant distractors; and (b) the sparse, interpretable attention patterns resulting from fine filtering (see Appendix~\ref{app:attmap}) could facilitate AI alignment efforts by making the model's causal reasoning more transparent. We believe these directions warrant further investigation.

\textbf{Potential Concerns.} As with any advancement in LLM capabilities, LSSAR could be misused to generate harmful content more effectively over longer contexts. However, we believe this risk is not fundamentally different from those posed by existing LLM technologies, and the same mitigation strategies (content filtering, responsible deployment practices) remain applicable.

\bibliography{example_paper}

@misc{stable_transformer,
  author={Qi, Xianbiao and Ye, Jiaquan and He, Yelin and Li, Chun-Guang and Zi, Bojia and Dai, Xili and Zou, Qin and Xiao, Rong},
  title={Stable-Transformer: Towards a Stable Transformer Training},
  year={2024},
  url={https://openreview.net/forum?id=lkRjnNW0gb},
}

@inproceedings{
  zhen2022cosformer,
  title={{cosFormer}: Rethinking Softmax In Attention},
  author={Zhen Qin and Weixuan Sun and Hui Deng and Dongxu Li and Yunshen Wei and Baohong Lv and Junjie Yan and Lingpeng Kong and Yiran Zhong},
  booktitle={International Conference on Learning Representations},
  year={2022}
}

@article{tsai2019transformer,
 author = {Tsai, Yao-Hung Hubert and Bai, Shaojie and Yamada, Makoto and Morency, Louis-Philippe and Salakhutdinov, Ruslan},
 journal = {arXiv:1908.11775},
 title = {Transformer dissection: a unified understanding of transformer's attention via the lens of kernel},
 year = {2019}
}

@inproceedings{katharopoulos2020transformers,
 author = {Katharopoulos, Angelos and Vyas, Apoorv and Pappas, Nikolaos and Fleuret, Fran{\c{c}}ois},
 booktitle = {International conference on machine learning},
 organization = {PMLR},
 pages = {5156--5165},
 title = {Transformers are {RNNs}: Fast autoregressive transformers with linear attention},
 year = {2020}
}

@inproceedings{han2023flatten,
 author = {Han, Dongchen and Pan, Xuran and Han, Yizeng and Song, Shiji and Huang, Gao},
 booktitle = {Proceedings of the IEEE/CVF international conference on computer vision},
 pages = {5961--5971},
 title = {Flatten transformer: Vision transformer using focused linear attention},
 year = {2023}
}

@article{arora2024simple,
 author = {Arora, Simran and Eyuboglu, Sabri and Zhang, Michael and Timalsina, Aman and Alberti, Silas and Zinsley, Dylan and Zou, James and Rudra, Atri and R{\'e}, Christopher},
 journal = {arXiv:2402.18668},
 title = {Simple linear attention language models balance the recall-throughput tradeoff},
 year = {2024}
}

@article{radford2019language,
 author = {Radford, Alec and Wu, Jeffrey and Child, Rewon and Luan, David and Amodei, Dario and Sutskever, Ilya and others},
 journal = {OpenAI blog},
 number = {8},
 pages = {9},
 title = {Language models are unsupervised multitask learners},
 volume = {1},
 year = {2019}
}

@article{su2024roformer,
 author = {Su, Jianlin and Ahmed, Murtadha and Lu, Yu and Pan, Shengfeng and Bo, Wen and Liu, Yunfeng},
 journal = {Neurocomputing},
 pages = {127063},
 publisher = {Elsevier},
 title = {Roformer: Enhanced transformer with rotary position embedding},
 volume = {568},
 year = {2024}
}

@inproceedings{henry2020query,
 author = {Henry, Alex and Dachapally, Prudhvi Raj and Pawar, Shubham Shantaram and Chen, Yuxuan},
 booktitle = {Findings of the Association for Computational Linguistics: EMNLP 2020},
 pages = {4246--4253},
 title = {Query-key normalization for transformers},
 year = {2020}
}

@inproceedings{houlsby2023scaling,
 author = {Dehghani, Mostafa and Djolonga, Josip and Mustafa, Basil and Padlewski, Piotr and Heek, Jonathan and Gilmer, Justin and Steiner, Andreas Peter and Caron, Mathilde and Geirhos, Robert and Alabdulmohsin, Ibrahim and others},
 booktitle = {International conference on machine learning},
 organization = {PMLR},
 pages = {7480--7512},
 title = {Scaling vision transformers to 22 billion parameters},
 year = {2023}
}

@inproceedings{liu2022swin,
 author = {Liu, Ze and Hu, Han and Lin, Yutong and Yao, Zhuliang and Xie, Zhenda and Wei, Yixuan and Ning, Jia and Cao, Yue and Zhang, Zheng and Dong, Li and others},
 booktitle = {Proceedings of the IEEE/CVF conference on computer vision and pattern recognition},
 pages = {12009--12019},
 title = {Swin transformer v2: Scaling up capacity and resolution},
 year = {2022}
}

@misc{su2021viewing,
 author = {Su, Jianlin},
 title = {Viewing the scale operation of attention from the perspective of entropy invariance},
 year = {2021}
}

@article{chiang2022overcoming,
 author = {Chiang, David and Cholak, Peter},
 journal = {arXiv:2202.12172},
 title = {Overcoming a theoretical limitation of self-attention},
 year = {2022}
}

@article{howard2017mobilenets,
 author = {Howard, Andrew G and Zhu, Menglong and Chen, Bo and Kalenichenko, Dmitry and Wang, Weijun and Weyand, Tobias and Andreetto, Marco and Adam, Hartwig},
 journal = {arXiv:1704.04861},
 title = {Mobilenets: Efficient convolutional neural networks for mobile vision applications},
 year = {2017}
}

@article{hendrycks2016gaussian,
 author = {Hendrycks, D},
 journal = {arXiv:1606.08415},
 title = {Gaussian Error Linear Units (Gelus)},
 year = {2016}
}

@techreport{rumelhart1986learning,
  title={Learning Internal Representations by Error Propagation},
  author={Rumelhart, David E. and Hinton, Geoffrey E. and Williams, Ronald J.},
  journal={Nature},
  volume={323},
  number={6088},
  pages={533--536},
  year={1986}
}

@article{misra2019mish,
 author = {Misra, Diganta},
 journal = {arXiv:1908.08681},
 title = {Mish: A self regularized non-monotonic activation function},
 year = {2019}
}

@article{so2021searching,
 author = {So, David and Ma{\'n}ke, Wojciech and Liu, Hanxiao and Dai, Zihang and Shazeer, Noam and Le, Quoc V},
 journal = {Advances in neural information processing systems},
 pages = {6010--6022},
 title = {Searching for efficient transformers for language modeling},
 volume = {34},
 year = {2021}
}

@inproceedings{zheng2015improving,
 author = {Zheng, Hao and Yang, Zhanlei and Liu, Wenju and Liang, Jizhong and Li, Yanpeng},
 booktitle = {2015 International joint conference on neural networks (IJCNN)},
 organization = {IEEE},
 pages = {1--4},
 title = {Improving deep neural networks using softplus units},
 year = {2015}
}

@article{shen2023study,
 author = {Shen, Kai and Guo, Junliang and Tan, Xu and Tang, Siliang and Wang, Rui and Bian, Jiang},
 journal = {arXiv:2302.06461},
 title = {A study on relu and softmax in transformer},
 year = {2023}
}

@article{wortsman2023replacing,
 author = {Wortsman, Mitchell and Lee, Jaehoon and Gilmer, Justin and Kornblith, Simon},
 journal = {arXiv:2309.08586},
 title = {Replacing softmax with relu in vision transformers},
 year = {2023}
}

@article{ramapuram2024theory,
 author = {Ramapuram, Jason and Danieli, Federico and Dhekane, Eeshan and Weers, Floris and Busbridge, Dan and Ablin, Pierre and Likhomanenko, Tatiana and Digani, Jagrit and Gu, Zijin and Shidani, Amitis and others},
 journal = {arXiv:2409.04431},
 title = {Theory, analysis, and best practices for sigmoid self-attention},
 year = {2024}
}

@inproceedings{li2022robust,
 author = {Li, Zhiyuan and Bhojanapalli, Srinadh and Zaheer, Manzil and Reddi, Sashank and Kumar, Sanjiv},
 booktitle = {International Conference on Machine Learning},
 organization = {PMLR},
 pages = {12656--12684},
 title = {Robust training of neural networks using scale invariant architectures},
 year = {2022}
}

@article{shah2024flashattention,
 author = {Shah, Jay and Bikshandi, Ganesh and Zhang, Ying and Thakkar, Vijay and Ramani, Pradeep and Dao, Tri},
 journal = {Advances in Neural Information Processing Systems},
 pages = {68658--68685},
 title = {Flashattention-3: Fast and accurate attention with asynchrony and low-precision},
 volume = {37},
 year = {2024}
}

@inproceedings{hron2020infinite,
 author = {Hron, Jiri and Bahri, Yasaman and Sohl-Dickstein, Jascha and Novak, Roman},
 booktitle = {International Conference on Machine Learning},
 organization = {PMLR},
 pages = {4376--4386},
 title = {Infinite attention: NNGP and NTK for deep attention networks},
 year = {2020}
}

@article{bai2024transformers,
 author = {Bai, Yu and Chen, Fan and Wang, Huan and Xiong, Caiming and Mei, Song},
 journal = {Advances in neural information processing systems},
 pages = {57125--57211},
 title = {Transformers as statisticians: Provable in-context learning with in-context algorithm selection},
 volume = {36},
 year = {2023}
}

@article{fu2024can,
 author = {Fu, Hengyu and Guo, Tianyu and Bai, Yu and Mei, Song},
 journal = {Advances in Neural Information Processing Systems},
 pages = {11912--11951},
 title = {What can a single attention layer learn? a study through the random features lens},
 volume = {36},
 year = {2023}
}

@inproceedings{hua2022transformer,
 author = {Hua, Weizhe and Dai, Zihang and Liu, Hanxiao and Le, Quoc},
 booktitle = {International conference on machine learning},
 organization = {PMLR},
 pages = {9099--9117},
 title = {Transformer quality in linear time},
 year = {2022}
}

@article{lu2021soft,
 author = {Lu, Jiachen and Yao, Jinghan and Zhang, Junge and Zhu, Xiatian and Xu, Hang and Gao, Weiguo and Xu, Chunjing and Xiang, Tao and Zhang, Li},
 journal = {Advances in Neural Information Processing Systems},
 pages = {21297--21309},
 title = {Soft: Softmax-free transformer with linear complexity},
 volume = {34},
 year = {2021}
}

@article{dubey2024llama,
  title={The {Llama} 3 herd of models},
  author={Dubey, Abhimanyu and Jauhri, Abhinav and Pandey, Abhinav and Kadian, Abhishek and Al-Dahle, Ahmad and Letman, Aiesha and Mathur, Akhil and Schelten, Alan and Yang, Amy and Fan, Angela and others},
  journal={arXiv:2407.21783},
  year={2024}
}

@article{achiam2023gpt,
 author = {Achiam, Josh and Adler, Steven and Agarwal, Sandhini and Ahmad, Lama and Akkaya, Ilge and Aleman, Florencia Leoni and Almeida, Diogo and Altenschmidt, Janko and Altman, Sam and Anadkat, Shyamal and others},
 journal = {arXiv:2303.08774},
 title = {Gpt-4 technical report},
 year = {2023}
}

@article{kazemnejad2023impact,
 author = {Kazemnejad, Amirhossein and Padhi, Inkit and Natesan Ramamurthy, Karthikeyan and Das, Payel and Reddy, Siva},
 journal = {Advances in Neural Information Processing Systems},
 pages = {24892--24928},
 title = {The impact of positional encoding on length generalization in transformers},
 volume = {36},
 year = {2023}
}

@article{chen2023extending,
 author = {Chen, Shouyuan and Wong, Sherman and Chen, Liangjian and Tian, Yuandong},
 journal = {arXiv:2306.15595},
 title = {Extending context window of large language models via positional interpolation},
 year = {2023}
}

@inproceedings{chi2022kerple,
 author = {Ta-Chung Chi and Ting-Han Fan and Peter J. Ramadge and Alexander Rudnicky},
 booktitle = {Advances in Neural Information Processing Systems},
 pages = {8386--8399},
 title = {KERPLE: Kernelized Relative Positional Embedding for Length Extrapolation},
 volume = {35},
 year = {2022}
}

@article{kiyono2021shape,
 author = {Kiyono, Shun and Kobayashi, Sosuke and Suzuki, Jun and Inui, Kentaro},
 journal = {arXiv:2109.05644},
 title = {Shape: Shifted absolute position embedding for transformers},
 year = {2021}
}

@article{golovneva2024contextual,
 author = {Golovneva, Olga and Wang, Tianlu and Weston, Jason and Sukhbaatar, Sainbayar},
 journal = {arXiv:2405.18719},
 title = {Contextual Position Encoding: Learning to Count What's Important},
 year = {2024}
}

@article{he2024two,
 author = {He, Zhenyu and Feng, Guhao and Luo, Shengjie and Yang, Kai and Wang, Liwei and Xu, Jingjing and Zhang, Zhi and Yang, Hongxia and He, Di},
 journal = {arXiv:2401.16421},
 title = {Two stones hit one bird: Bilevel positional encoding for better length extrapolation},
 year = {2024}
}

@article{huang2020improve,
 author = {Huang, Zhiheng and Liang, Davis and Xu, Peng and Xiang, Bing},
 journal = {arXiv:2009.13658},
 title = {Improve transformer models with better relative position embeddings},
 year = {2020}
}

@article{li2023functional,
 author = {Li, Shanda and You, Chong and Guruganesh, Guru and Ainslie, Joshua and Ontanon, Santiago and Zaheer, Manzil and Sanghai, Sumit and Yang, Yiming and Kumar, Sanjiv and Bhojanapalli, Srinadh},
 journal = {arXiv:2310.04418},
 title = {Functional interpolation for relative positions improves long context transformers},
 year = {2023}
}

@article{likhomanenko2021cape,
 author = {Likhomanenko, Tatiana and Xu, Qiantong and Synnaeve, Gabriel and Collobert, Ronan and Rogozhnikov, Alex},
 journal = {Advances in Neural Information Processing Systems},
 pages = {16079--16092},
 title = {Cape: Encoding relative positions with continuous augmented positional embeddings},
 volume = {34},
 year = {2021}
}

@article{liu2023scaling,
 author = {Liu, Xiaoran and Yan, Hang and Zhang, Shuo and An, Chenxin and Qiu, Xipeng and Lin, Dahua},
 journal = {arXiv:2310.05209},
 title = {Scaling laws of rope-based extrapolation},
 year = {2023}
}

@article{wang2024resonance,
 author = {Wang, Suyuchen and Kobyzev, Ivan and Lu, Peng and Rezagholizadeh, Mehdi and Liu, Bang},
 journal = {arXiv:2403.00071},
 title = {Resonance rope: Improving context length generalization of large language models},
 year = {2024}
}

@article{zheng2024cape,
  author = {Chuanyang Zheng and Yihang Gao and Han Shi and Minbin Huang and Jingyao Li and Jing Xiong and Xiaozhe Ren and Michael Ng and Xin Jiang and Zhenguo Li and Yu Li},
  title = {{CAPE}: Context-Adaptive Positional Encoding for Length Extrapolation},
  journal = {arXiv:2405.14722},
  year = {2024}
}

@article{liu2024deepseek,
 author = {Liu, Aixin and Feng, Bei and Xue, Bing and Wang, Bingxuan and Wu, Bochao and Lu, Chengda and Zhao, Chenggang and Deng, Chengqi and Zhang, Chenyu and Ruan, Chong and others},
 journal = {arXiv:2412.19437},
 title = {Deepseek-v3 technical report},
 year = {2024}
}

@article{aueb2016one,
 author = {Michalis K. Titsias},
 journal = {Advances in Neural Information Processing Systems},
 title = {One-vs-each approximation to softmax for scalable estimation of probabilities},
 volume = {29},
 year = {2016}
}

@article{gao2017properties,
 author = {Gao, Bolin and Pavel, Lacra},
 journal = {arXiv:1704.00805},
 title = {On the properties of the softmax function with application in game theory and reinforcement learning},
 year = {2017}
}

@article{jang2016categorical,
 author = {Jang, Eric and Gu, Shixiang and Poole, Ben},
 journal = {arXiv:1611.01144},
 title = {Categorical reparameterization with gumbel-softmax},
 year = {2016}
}

@article{penedo2024fine,
 author = {Penedo, Guilherme and Kydl{\'\i}{\v{c}}ek, Hynek and Lozhkov, Anton and Mitchell, Margaret and Raffel, Colin A and Von Werra, Leandro and Wolf, Thomas and others},
 journal = {Advances in Neural Information Processing Systems},
 pages = {30811--30849},
 title = {The fineweb datasets: Decanting the web for the finest text data at scale},
 volume = {37},
 year = {2024}
}

@article{deepseekai2025deepseekr1,
 author = {Guo, Daya and Yang, Dejian and Zhang, Haowei and Song, Junxiao and Zhang, Ruoyu and Xu, Runxin and Zhu, Qihao and Ma, Shirong and Wang, Peiyi and Bi, Xiao and others},
 journal = {arXiv:2501.12948},
 title = {Deepseek-r1: Incentivizing reasoning capability in llms via reinforcement learning},
 year = {2025}
}

@misc{kaiokendev2023things,
author = {kaiokendev},
title = {Things I'm learning while training superhot},
year = {2023},
url = {https://kaiokendev.github.io/til#extending-context-to-8k},
}

@misc{bloc97_2023_correction,
  author = {bloc97},
  title = {Add {NTK}-Aware interpolation "by parts" correction},
  year = {2023},
  url = {https://github.com/jquesnelle/scaled-rope/pull/1},
  note = {GitHub Pull Request}
}

@misc{bloc97_2023_ntk,
  author = {bloc97},
  title = {{NTK}-Aware Scaled RoPE allows LLaMA models to have extended (8k+) context size without any fine-tuning and minimal perplexity degradation},
  year = {2023},
  url = {https://www.reddit.com/r/LocalLLaMA/comments/14lz7j5/ntkaware_scaled_rope_allows_llama_models_to_have/},
  note = {Reddit post}
}

@misc{emozilla2023dynamic,
  author = {emozilla},
  title = {Dynamically Scaled RoPE further increases performance of long context LLaMA with zero fine-tuning},
  year = {2023},
  url = {https://www.reddit.com/r/LocalLLaMA/comments/14mrgpr/dynamically_scaled_rope_further_increases/},
  note = {Reddit post}
}

@article{clark2018think,
 author = {Clark, Peter and Cowhey, Isaac and Etzioni, Oren and Khot, Tushar and Sabharwal, Ashish and Schoenick, Carissa and Tafjord, Oyvind},
 journal = {arXiv:1803.05457},
 title = {Think you have solved question answering? try arc, the ai2 reasoning challenge},
 year = {2018}
}

@article{zellers2019hellaswag,
 author = {Zellers, Rowan and Holtzman, Ari and Bisk, Yonatan and Farhadi, Ali and Choi, Yejin},
 journal = {arXiv:1905.07830},
 title = {Hellaswag: Can a machine really finish your sentence?},
 year = {2019}
}

@inproceedings{bisk2020piqa,
 author = {Bisk, Yonatan and Zellers, Rowan and Gao, Jianfeng and Choi, Yejin and others},
 booktitle = {Proceedings of the AAAI conference on artificial intelligence},
 number = {05},
 pages = {7432--7439},
 title = {Piqa: Reasoning about physical commonsense in natural language},
 volume = {34},
 year = {2020}
}

@article{hendrycksmeasuring,
 author = {Hendrycks, Dan and Burns, Collin and Basart, Steven and Zou, Andy and Mazeika, Mantas and Song, Dawn and Steinhardt, Jacob},
 journal = {arXiv:2009.03300},
 title = {Measuring massive multitask language understanding},
 year = {2020}
}

@article{welbl2017crowdsourcing,
 author = {Welbl, Johannes and Liu, Nelson F and Gardner, Matt},
 journal = {arXiv:1707.06209},
 title = {Crowdsourcing multiple choice science questions},
 year = {2017}
}

@inproceedings{chen2021summscreen,
 author = {Chen, Mingda and Chu, Zewei and Wiseman, Sam and Gimpel, Kevin},
 booktitle = {Proceedings of the 60th Annual Meeting of the Association for Computational Linguistics (Volume 1: Long Papers)},
 pages = {8602--8615},
 title = {Summscreen: A dataset for abstractive screenplay summarization},
 year = {2022}
}

@misc{eval-harness,
 author = {Gao, Leo and Tow, Jonathan and Abbasi, Baber and Biderman, Stella and Black, Sid and DiPofi, Anthony and Foster, Charles and Golding, Laurence and Hsu, Jeffrey and Le Noac'h, Alain and Li, Haonan and McDonell, Kyle and Muennighoff, Niklas and Ociepa, Chris and Phang, Jason and Reynolds, Laria and Schoelkopf, Hailey and Skowron, Aviya and Sutawika, Lintang and Tang, Eric and Thite, Anish and Wang, Ben and Wang, Kevin and Zou, Andy},
 doi = {10.5281/zenodo.12608602},
 month = {07},
 publisher = {Zenodo},
 title = {A framework for few-shot language model evaluation},
 url = {https://zenodo.org/records/12608602},
 version = {v0.4.3},
 year = {2024}
}

@article{velivckovic2024softmax,
  title={Softmax is not Enough (for Sharp Size Generalisation)},
  author={Veli{\v{c}}kovi{\'c}, Petar and Perivolaropoulos, Christos and Barbero, Federico and Pascanu, Razvan},
  journal={arXiv:2410.01104},
  year={2024}
}

@article{mohtashami2023random,
 author = {Mohtashami, Amirkeivan and Jaggi, Martin},
 journal = {Advances in Neural Information Processing Systems},
 pages = {54567--54585},
 title = {Random-access infinite context length for transformers},
 volume = {36},
 year = {2023}
}

@article{yang2025qwen3,
 author = {Yang, An and Li, Anfeng and Yang, Baosong and Zhang, Beichen and Hui, Binyuan and Zheng, Bo and Yu, Bowen and Gao, Chang and Huang, Chengen and Lv, Chenxu and others},
 journal = {arXiv:2505.09388},
 title = {Qwen3 technical report},
 year = {2025}
}

@article{xiaoefficient,
 author = {Xiao, Guangxuan and Tian, Yuandong and Chen, Beidi and Han, Song and Lewis, Mike},
 journal = {arXiv:2309.17453},
 title = {Efficient streaming language models with attention sinks},
 year = {2023}
}

@article{dao2022flashattention,
 author = {Dao, Tri and Fu, Dan and Ermon, Stefano and Rudra, Atri and R{\'e}, Christopher},
 journal = {Advances in neural information processing systems},
 pages = {16344--16359},
 title = {Flashattention: Fast and memory-efficient exact attention with io-awareness},
 volume = {35},
 year = {2022}
}

@article{dao2023flashattention,
 author = {Dao, Tri},
 journal = {arXiv:2307.08691},
 title = {Flashattention-2: Faster attention with better parallelism and work partitioning},
 year = {2023}
}

@article{choromanski2020rethinking,
 author = {Choromanski, Krzysztof and Likhosherstov, Valerii and Dohan, David and Song, Xingyou and Gane, Andreea and Sarlos, Tamas and Hawkins, Peter and Davis, Jared and Mohiuddin, Afroz and Kaiser, Lukasz and others},
 journal = {arXiv:2009.14794},
 title = {Rethinking attention with performers},
 year = {2020}
}

@article{beltagy2020longformer,
 author = {Beltagy, Iz and Peters, Matthew E and Cohan, Arman},
 journal = {arXiv:2004.05150},
 title = {Longformer: The long-document transformer},
 year = {2020}
}

@inproceedings{gu2024mamba,
 author = {Gu, Albert and Dao, Tri},
 booktitle = {First conference on language modeling},
 title = {Mamba: Linear-time sequence modeling with selective state spaces},
 year = {2024}
}

@article{gunasekar2023textbooks,
 author = {Gunasekar, Suriya and Zhang, Yi and Aneja, Jyoti and Mendes, Caio C{\'e}sar Teodoro and Del Giorno, Allie and Gopi, Sivakanth and Javaheripi, Mojan and Kauffmann, Piero and de Rosa, Gustavo and Saarikivi, Olli and others},
 journal = {arXiv:2306.11644},
 title = {Textbooks are all you need},
 year = {2023}
}

@inproceedings{biderman2023pythia,
 author = {Biderman, Stella and Schoelkopf, Hailey and Anthony, Quentin Gregory and Bradley, Herbie and O’Brien, Kyle and Hallahan, Eric and Khan, Mohammad Aflah and Purohit, Shivanshu and Prashanth, USVSN Sai and Raff, Edward and others},
 booktitle = {International Conference on Machine Learning},
 organization = {PMLR},
 pages = {2397--2430},
 title = {Pythia: A suite for analyzing large language models across training and scaling},
 year = {2023}
}

@inproceedings{paperno2016lambada,
 author = {Paperno, Denis and Kruszewski, Germ{\'a}n and Lazaridou, Angeliki and Pham, Ngoc-Quan and Bernardi, Raffaella and Pezzelle, Sandro and Baroni, Marco and Boleda, Gemma and Fern{\'a}ndez, Raquel},
 booktitle = {Proceedings of the 54th annual meeting of the association for computational linguistics (volume 1: Long papers)},
 pages = {1525--1534},
 title = {The LAMBADA dataset: Word prediction requiring a broad discourse context},
 year = {2016}
}

@article{deepseekv32,
 author = {Liu, Aixin and Mei, Aoxue and Lin, Bangcai and Xue, Bing and Wang, Bingxuan and Xu, Bingzheng and Wu, Bochao and Zhang, Bowei and Lin, Chaofan and Dong, Chen and others},
 journal = {arXiv:2512.02556},
 title = {Deepseek-v3. 2: Pushing the frontier of open large language models},
 year = {2025}
}

@inproceedings{martins2016softmax,
 author = {Martins, Andre and Astudillo, Ramon},
 booktitle = {International conference on machine learning},
 organization = {PMLR},
 pages = {1614--1623},
 title = {From softmax to sparsemax: A sparse model of attention and multi-label classification},
 year = {2016}
}

@article{peters2019sparse,
 author = {Peters, Ben and Niculae, Vlad and Martins, Andr{\'e} FT},
 journal = {arXiv:1905.05702},
 title = {Sparse sequence-to-sequence models},
 year = {2019}
}

@article{wu2025emergence,
  title={On the emergence of position bias in transformers},
  author={Wu, Xinyi and Wang, Yifei and Jegelka, Stefanie and Jadbabaie, Ali},
  journal={arXiv:2502.01951},
  year={2025}
}

@article{bengio2013representation,
  title={Representation learning: A review and new perspectives},
  author={Bengio, Yoshua and Courville, Aaron and Vincent, Pascal},
  journal={IEEE transactions on pattern analysis and machine intelligence},
  volume={35},
  number={8},
  pages={1798--1828},
  year={2013},
  publisher={IEEE}
}

@article{xie2025mhc,
  title={mHC: Manifold-Constrained Hyper-Connections},
  author={Xie, Zhenda and Wei, Yixuan and Cao, Huanqi and Zhao, Chenggang and Deng, Chengqi and Li, Jiashi and Dai, Damai and Gao, Huazuo and Chang, Jiang and Zhao, Liang and others},
  journal={arXiv:2512.24880},
  year={2025}
}

@article{olshausen1996emergence,
  title={Emergence of simple-cell receptive field properties by learning a sparse code for natural images},
  author={Olshausen, Bruno A and Field, David J},
  journal={Nature},
  volume={381},
  number={6583},
  pages={607--609},
  year={1996},
  publisher={Nature Publishing Group UK London}
}

@article{vafa2025inductive,
  title={What Has a Foundation Model Found? Using Inductive Bias to Probe for World Models},
  author={Vafa, Keyon and Chang, Peter G and Rambachan, Ashesh and Mullainathan, Sendhil},
  booktitle={International Conference on Machine Learning},
  pages={60727--60747},
  year={2025},
  organization={PMLR}
}

@article{cranmer2023interpretable,
  title={Interpretable machine learning for science with {PySR} and {SymbolicRegression.jl}},
  author={Cranmer, Miles},
  journal={arXiv:2305.01582},
  year={2023}
}

@article{heeger1992normalization,
  title={Normalization of cell responses in cat striate cortex},
  author={Heeger, David J},
  journal={Visual neuroscience},
  volume={9},
  number={2},
  pages={181--197},
  year={1992},
  publisher={Cambridge University Press}
}

@article{douglas1995recurrent,
  title={Recurrent excitation in neocortical circuits},
  author={Douglas, Rodney J and Koch, Christof and Mahowald, Misha and Martin, Kevan AC and Suarez, Humbert H},
  journal={Science},
  volume={269},
  number={5226},
  pages={981--985},
  year={1995},
  publisher={American Association for the Advancement of Science}
}

@article{carandini2012normalization,
  title={Normalization as a canonical neural computation},
  author={Carandini, Matteo and Heeger, David J},
  journal={Nature reviews neuroscience},
  volume={13},
  number={1},
  pages={51--62},
  year={2012},
  publisher={Nature Publishing Group UK London}
}

@misc{CREATE,
  title = {{King's Computational Research, Engineering and Technology Environment (CREATE)}},
  author = {{King's College London}},
  year = {2022},
  url = {https://doi.org/10.18742/rnvf-m076},
  note = {Retrieved March 2, 2022},
  doi = {10.18742/rnvf-m076}
}

@incollection{mcintosh2024isambard,
  title={Isambard-ai: a leadership-class supercomputer optimised specifically for artificial intelligence},
  author={McIntosh-Smith, Simon and Alam, Sadaf and Woods, Christopher},
  booktitle={Proceedings of the Cray User Group},
  pages={44--54},
  year={2024}
}

@article{nakanishi2025scalable,
  title={Scalable-Softmax Is Superior for Attention},
  author={Nakanishi, Ken M.},
  journal={arXiv:2501.19399},
  year={2025}
}
\bibliographystyle{icml2026}

\newpage
\appendix
\onecolumn
\renewcommand{\appendixname}{Appendix~\Alph{section}}
\renewcommand{\thetable}{A.\arabic{table}}
\renewcommand{\thefigure}{A.\arabic{figure}}
\renewcommand{\theequation}{A.\arabic{equation}}
\setcounter{equation}{0}

\section{Theoretical Analysis}

\subsection{Detailed Algorithm of LSSAR}
\label{app:algorithm}

To help readers better understand the theoretical analysis in this appendix without needing to repeatedly refer back to the main text for equations, we provide the detailed algorithmic procedure of LSSAR in Algorithm~\ref{alg:lssar_detailed}. This self-contained reference explicitly presents all mathematical operations involved in each stage.

\begin{center}
    \begin{minipage}{0.6\textwidth}
        \begin{algorithm}[H]
            \caption{Detailed Computation of LSSAR}
            \label{alg:lssar_detailed}
            \begin{algorithmic}[1]
                \Require $\mathbf{Q}, \mathbf{K} \in \mathbb{R}^{L \times d}$: Query and Key matrices
                \Require $\mathbf{M'} \in \{0, 1\}^{L \times L}$: Causal mask
                \Require $p > 0$: Power parameter
                \Ensure $\mathbf{A} \in \mathbb{R}^{L \times L}$: Sparse attention matrix
                \Statex
                \Statex \textbf{Stage 1: Normalisation (LSSA)}
                \State $\mathbf{Q}_{i} \leftarrow \mathbf{Q}_i / \|\mathbf{Q}_i\|_2$ \Comment{$l_2$-normalise queries}
                \State $\mathbf{K}_{i} \leftarrow \mathbf{K}_i / \|\mathbf{K}_i\|_2$ \Comment{$l_2$-normalise keys}
                \State $\mathbf{N}_{ij} \leftarrow i$ \Comment{$\mathbf{N}$: token count matrix}
                \State $\mathbf{S} \leftarrow (\log d \cdot \log \mathbf{N}) \otimes \mathbf{Q}\mathbf{K}^T$ \Comment{$\mathbf{S}$: length-scaled scores}
                \State $\mathbf{A} \leftarrow \operatorname{Softplus}(\mathbf{S}) \otimes \mathbf{M'}$ \Comment{Apply activation and mask}
                \State $\mathbf{A}_{i} \leftarrow \mathbf{A}_i / \|\mathbf{A}_i\|_1$ \Comment{First $l_1$-normalisation}
                \Statex
                \Statex \textbf{Stage 2: Sharpening (Re-weighting)}
                \State $\mathbf{O}_{ij} \leftarrow \mathbf{1}_{[i > 3]}$ \Comment{$\mathbf{O}$: offset (0 for first 3 rows)}
                \State $\mathbf{A} \leftarrow \operatorname{ReLU}^p(\mathbf{A} \cdot \mathbf{N} - \mathbf{O})$ \Comment{Shift-ReLU sharpening}
                \State $\mathbf{A}_{i} \leftarrow \mathbf{A}_i / \|\mathbf{A}_i\|_1$ \Comment{Second $l_1$-normalisation}
                \Statex
                \State \textbf{return} $\mathbf{A}$
            \end{algorithmic}
        \end{algorithm}
    \end{minipage}
\end{center}

\subsection{From Soft Inhibition to Sparse Coding}
\label{app:neuroscience}

Beyond the mathematical optimisation dynamics, the role of the $l_{1}$-norm in attention mechanisms can be reinterpreted through the lens of computational neuroscience. In biological neural networks, energy efficiency is a fundamental constraint; the brain consumes a disproportionate amount of the body's metabolic energy, yet individual neurons operate under a regime of sparse coding \citep{olshausen1996emergence}. Most neurons remain silent at any given moment, firing action potentials only when specific, salient stimuli are present. This principle suggests that an ideal attention mechanism should not merely distribute weights, but actively suppress irrelevant signals to conserve computational `energy' and maximise the signal-to-noise ratio.

The role of Softmax attention can be rigorously understood through the lens of Divisive Normalisation, a canonical neural computation proposed by \citet{heeger1992normalization} to describe the nonlinear response properties of neurons in the primary visual cortex (V1). The Divisive Normalisation model computes a neuron's response as $R_i = \frac{f(E_i)}{\sigma + \sum_j f(E_j)}$, where $E_i$ is the excitatory input and $f(\cdot)$ is a nonlinear gain function. Comparing this with the Softmax formulation in \cref{equ:softmax}, we observe a striking mathematical correspondence: both express a ratio where the numerator is a nonlinearly transformed activation of a single neuron (or position), and the denominator is the aggregate of all competing neurons' activations. This reveals that Softmax Attention is essentially a special case of neural competition, where the excitatory input undergoes an exponential nonlinear gain transformation $f(x) = e^{x}$, followed by lateral inhibition mediated by the $l_{1}$-norm denominator. Biologically, this mirrors the function of inhibitory interneurons, which suppress the activity of surrounding neurons, making the most active neurons more salient. However, Softmax fundamentally creates a `dense code': due to the nature of the exponential function, no weight is ever exactly zero. In neuroscientific terms, this corresponds to a system with a high spontaneous firing rate or background noise. While this ensures differentiability, it violates the principle of metabolic efficiency and fails to filter out off-manifold noise, leading to the susceptibility to distractors discussed in previous sections.

LSSAR advances this paradigm significantly, functioning as a unified computational model of cortical microcircuits that integrates multiple biological mechanisms into a single differentiable operator:

\textbf{Stage 1: Gain Control via Divisive (Shunting) Inhibition.} The first stage directly mirrors the classic Divisive Normalisation model. It stabilises signal magnitude by normalising each neuron's response relative to the pooled activity of its neighbours, effectively implementing shunting inhibition. Crucially, the $l_1$-norm in this stage serves more than preliminary gain control; its primary function is to enforce \textit{global competition} among all neurons. By coupling all inputs via the denominator, it ensures that every neuron, even those destined for suppression by subsequent thresholding via Shift-ReLU, contributes to the gradient calculation during backpropagation. This gradient coupling is the key mechanism that prevents the dead neuron problem: if a suppressed neuron needs to become active to lower the loss, the gradient flows back through the normalisation term to push its pre-activation value up, creating a robust learning dynamic without requiring slow homeostatic regulation.

\textbf{Stage 2: Noise Filtering and Decision Sharpening.} The second stage integrates two complementary biological mechanisms. First, the Shift-ReLU mechanism functionally mimics the rheobase or voltage threshold in biological neurons: a neuron does not fire unless its input current exceeds a dynamic baseline, represented here by the mean activity. This implements \textit{subtractive inhibition} \citep{carandini2012normalization}, where a uniform inhibitory signal (the local average) is subtracted from all inputs before a hard threshold is applied. By assigning exact zero probability to tokens falling below this threshold, LSSAR transitions from a dense, noisy representation to a true sparse code. This parallels the `all-or-none' nature of biological action potentials, where sub-threshold stimuli result in silence. The Shift-ReLU mechanism ensures true metabolic efficiency via sparsity: if the input stimulus is weak or uniform, the mean-shift operation effectively suppresses all activity to zero, simulating the silence of a neuron in the absence of a driving stimulus.

Second, the power operation ($\operatorname{ReLU}^p$) followed by re-normalisation implements \textit{recurrent lateral inhibition} in a feed-forward manner. In biological circuits, ambiguity among competing stimuli is resolved through iterative cycles of mutual inhibition until a winner emerges,the well-known Winner-Take-All (WTA) dynamics \citep{douglas1995recurrent}. Our power parameter $p$ mathematically simulates the steady-state distribution of this recurrent process: a higher $p$ corresponds to more rounds of recurrent competition, sharpening the decision without the computational cost of iterative unrolling. The second $l_1$-norm, applied after the power operation, functions as the convergence mechanism for this recurrent competition, ensuring the output remains a valid probability distribution representing the WTA outcome.

The separation of the two $l_1$-norm stages is motivated primarily by engineering considerations rather than biological constraints. The first $l_1$-norm establishes gradient connectivity and global competition \textit{before} the decision-making process begins, while the second $l_1$-norm implements the competitive dynamics that sharpen the final decision. This architectural choice allows LSSAR to maintain robust gradient flow (preventing dead neurons) while still achieving sharp, sparse attention distributions, which single-stage mechanisms cannot achieve. While this design incidentally mirrors certain aspects of cortical processing (e.g., the temporal separation between gain control and decision-making), the primary justification is optimisation stability rather than biological plausibility.

Beyond the engineering rationale, there is an interesting parallel to the semi-saturation constant $\sigma$ in the standard Divisive Normalisation model, which handles low-stimulus regimes. In LSSAR, the Shift-ReLU mechanism serves a functionally analogous role as a hard dynamic threshold: when input stimuli are weak or uniform, the mean-shift operation suppresses activity to zero, while the power parameter $p$ allows each attention head to tune its selectivity or sharpening based on task demands. However, this analogy should be understood as an observation rather than a design principle.

This transition from dense to sparse coding offers distinct advantages in high-dimensional sensory processing, akin to how the visual cortex uses sparse coding to disentangle explanatory factors from noise. LSSAR's hard sparsity acts as a denoising filter, preventing the aggregation of numerous weak, irrelevant signals, identified as `manifold drift' that confuse standard Softmax models (see Appendix~\ref{app:graph_theory} for details). From this perspective, LSSAR is not merely an engineering improvement but a step towards a more biologically plausible attention mechanism that unifies gain control, noise filtering, and competitive decision-making within a single differentiable framework.

\subsubsection{Coarse-to-Fine Causal Filtering and Symbolic Regression}
\label{app:depth_physical_law}

The biological mechanisms described above can be unified into a coherent
framework of \textit{coarse-to-fine causal filtering}, which provides a
plausible explanation for why LSSAR can improve symbolic-regression outcomes in
our experiments (\cref{sec:symbolic_regression}). In particular, the recovered
inverse-square dependence suggests that this filtering process can help isolate
the compact physical structure present in orbital trajectories.

\textbf{Coarse Filtering via Shift-ReLU.} The Shift-ReLU mechanism performs \textit{coarse-grained noise elimination}. By subtracting the mean activity and applying a hard threshold, this stage categorically removes tokens whose relevance falls below the population average. In the context of orbital trajectory sequences, this may correspond to down-weighting redundant or weakly informative time steps, local trajectory fluctuations, and numerical noise that do not strongly constrain the force relation. This coarse filtering reduces the search space for the subsequent refinement stage, transforming a dense, noisy attention distribution into a sparse set of candidate causal variables.

\textbf{Fine Filtering via WTA Dynamics.} The power operation implements \textit{fine-grained causal selection} among the surviving candidates. As discussed earlier, this operation mathematically simulates $p$ iterations of recurrent lateral inhibition, progressively amplifying the strongest signals while suppressing weaker competitors. Physical relationships, such as the inverse-square dependence $F \propto m/r^2$, are characterised by sparse functional structure: the gravitational force depends on a small set of variables (mass and distance), not on the myriad of irrelevant contextual features present in raw trajectory data. The WTA dynamics may help filter candidate dependencies by amplifying compact, structured relations while suppressing spurious correlations that survived the coarse filtering stage.

This coarse-to-fine causal filtering mirrors how the brain's cortical circuits perform perceptual inference: an initial rapid feedforward sweep eliminates obviously irrelevant stimuli (analogous to Shift-ReLU), followed by iterative recurrent processing that resolves ambiguity among competing hypotheses (analogous to the power operation). Standard Softmax attention, with its dense, non-sparse distributions, lacks both filtering stages: it aggregates information from all tokens indiscriminately, injecting noise that obscures sparse underlying relationships.

\subsection{Comparison with Projection-Based Sparse Mechanisms}
\label{app:sparsemax_comparison}

In the main text, we addressed the limitations of Softmax regarding attention
smoothing and numerical instability. It is worth noting that a distinct
family of probability mapping functions, most notably Sparsemax \citep{martins2016softmax}
and Entmax \citep{peters2019sparse}, has also been proposed to induce
sparsity in attention distributions. While these methods share our motivation
of mitigating noise and improving interpretability by assigning zero probability
to irrelevant tokens, they rely on a fundamentally different mathematical
foundation: using Euclidean projection onto the probability simplex. This creates
specific challenges for large-scale language modelling compared to
our proposed LSSAR mechanism.

The primary distinction lies in the computational complexity and hardware
efficiency. Sparsemax and its generalised form, Entmax, require computing a threshold
that involves sorting or iterative bisection of the input vector elements.
This sorting operation typically has a complexity of
$\mathcal{O}(L \log L)$ or $\mathcal{O}(L)$, and is computationally expensive
on modern accelerators (GPUs/TPUs) compared to matrix multiplications and element-wise
operations. More critically, the requirement for global sorting or thresholding
across the sequence dimension hinders compatibility with I/O-aware optimisations
such as FlashAttention. FlashAttention relies on tiling techniques that
process blocks of the attention matrix in fast SRAM without materialising
the full matrix. Since Sparsemax requires global knowledge of the row to determine
the projection threshold, it cannot be easily fused into tiled kernels. In contrast,
our LSSAR mechanism relies on element-wise operations (Softplus, ReLU, and
power functions) and row-wise normalisation. These operations are local or
reducible, suggesting that LSSAR may be amenable to future tiled or fused
attention implementations. However, we have not yet demonstrated such an
optimised kernel, and the additional operations still introduce non-trivial
computational overhead in the current implementation.

Furthermore, the training dynamics of LSSAR differ significantly from projection-based methods. Both Sparsemax and direct ReLU-based attention mechanisms introduce exact zeros into the probability distribution, creating ``dead zones'' where the gradient is strictly zero. This `dead neuron' problem fundamentally impedes gradient flow: once a token's score falls below the threshold, it loses all gradient feedback and becomes permanently isolated from the optimisation process.

LSSAR resolves this problem through its two-stage architecture. The key insight is that the $l_1$-normalisation in Stage 1 establishes gradient connectivity \textit{before} the sparsity-inducing operations in Stage 2. By coupling all tokens through the shared denominator, even tokens that are subsequently suppressed to zero continue to receive gradient feedback, enabling their potential re-activation in future training steps. This design allows LSSAR to achieve sparse attention distributions while maintaining the robust gradient landscape essential for stable training. A detailed biological interpretation of this mechanism as divisive and subtractive inhibition is provided in Appendix~\ref{app:neuroscience}.

\subsection{Log-Space Formulation and Thermodynamic Interpretation}
\label{app:log_space}
From an engineering perspective, we observe that the proposed re-weighting mechanism
can be reformulated in the logarithmic space, revealing a direct equivalence
to a temperature-scaled Softmax function. This insight allows for
substantial
acceleration by leveraging highly optimised standard kernels. Let
us define
the transformation $u_{i}= p \cdot \log(x_{i})$, where cases of
$x_{i}=0$ are
handled by assigning $u_{i}= -\infty$. Applying the standard Softmax
function to the transformed scores $\mathbf{u}$ yields:
\begin{equation}
    \text{Softmax}(\mathbf{u})_{i}= \frac{e^{u_i}}{\sum_{j}e^{u_j}}= \frac{e^{p
                \log x_i}}{\sum_{j}e^{p \log x_j}}
\end{equation}
Utilising the logarithmic identity $e^{a \log b}= b^{a}$, the expression
simplifies
to the original power formulation:
\begin{equation}
    \frac{(e^{\log x_i})^{p}}{\sum_{j}(e^{\log x_j})^{p}}= \frac{x_{i}^{p}}{\sum_{j}x_{j}^{p}}
\end{equation}
This derivation proves that Stage 2 is mathematically identical to applying
Softmax
to the logarithm of the ReLU outputs scaled by $p$.

To rigorously justify the interpretation of $p$ as an \textit{inverse
    temperature coefficient}, we can juxtapose our log-space formulation with
the
standard Boltzmann distribution (or Softmax with temperature $T$) used
in statistical
mechanics and machine learning:
\begin{align}
    \text{Standard Boltzmann:}\quad & P(i) \propto \exp\left( \frac{z_{i}}{T}\right)    \\
    \text{LSSAR Log-Space:}\quad    & P(i) \propto \exp\left( p \cdot \log x_{i}\right)
\end{align}
By mapping the logarithmic scores $\log x_{i}$ to the energy states (logits)
$z_{i}$, the structural alignment becomes evident:
\begin{equation}
    p \equiv \frac{1}{T}
\end{equation}
While
standard
attention mechanisms often operate at a fixed temperature (e.g., $T= \sqrt{d}$), the LSSAR framework employs $p$ to dynamically control the
sharpness
of the distribution. A large power parameter (e.g., $p=15$) corresponds
to a
regime of extremely low temperature ($T \approx 0.06$). As
$p \to \infty$ (or
$T \to 0$), the system approaches its ground state, where
the Softmax function
asymptotically converges to the $\text{argmax}$
operation. Consequently, the
re-weighting mechanism can be understood as a
process of active entropy minimisation,
forcing the model to make decisive,
low-uncertainty selections of relevant tokens.
This thermodynamic view explains
why increasing $p$ fundamentally
strengthens the model's ability to maintain
focus over long sequences, preventing
the probability mass from diluting
into a uniform distribution (high entropy)
as the context window expands.

It is essential to highlight that the effectiveness of the re-weighting
stage
relies not only on the power parameter $p$ but also on the preceding
ReLU activation.
The shift operation ($\mathbf{A}\otimes \mathbf{N}- \mathbf{O}$)
followed by
ReLU functions as a hard noise filter or a dynamic hard mask. By
truncating the
tail of the distribution and assigning exact zero probability
to tokens falling
below the heuristic baseline, this step explicitly eliminates
the majority
of irrelevant context before the sharpening process begins.

This hard masking significantly alleviates the burden on the power operation.
In a standard Softmax (infinite support), the temperature parameter must be
extremely
low (i.e. $p$ extremely high) to suppress the accumulated mass of thousands
of irrelevant tokens to a negligible level. In LSSAR, since the ReLU has
already
removed the noise tokens, the power parameter $p$ serves a more refined
purpose: it only needs to distinguish between the remaining potentially
relevant
candidates. This synergy allows LSSAR to achieve sharp, sparse
attention distributions
with a moderate finite $p$ (e.g., $p=15$), avoiding the
numerical
instability and gradient vanishing problems often associated with the
extreme temperature scaling required by standard attention mechanisms to
achieve
similar sparsity.

\label{app:graph_theory}
\subsection{Graph-Theoretic and Geometric Perspectives}

This section provides a rigorous theoretical justification for the
superiority of LSSAR over standard Softmax attention. Our analysis proceeds from
two complementary perspectives. First, we employ the graph-theoretic framework
proposed by \citep{wu2025emergence} to demonstrate how LSSAR structurally
eliminates the inherent position bias found in Transformers. Second, we
interpret this structural correction through the lens of the manifold
hypothesis to explain the model's enhanced length extrapolation capabilities.

\subsubsection{Structural Bias Elimination via Graph Topology}

Recent theoretical advancements by \citep{wu2025emergence} have modelled the
flow of information in Transformers as a directed graph $\mathcal{G}$, where
tokens represent nodes and attention scores represent weighted edges. Under
the standard causal masking setting, the first token in a sequence acts as a
unique `centre node': the only node that is a predecessor to all other tokens.
\citep{wu2025emergence} proved that when the attention mechanism utilises
the Softmax function, which assigns a strictly positive probability ($A_{ij}>
    0$) to all allowed connections, the attention distribution inevitably
collapses towards this centre node as the network depth increases. This phenomenon,
often termed the `attention sink', is not necessarily driven by semantic relevance
but is a structural artefact of the full connectivity within the causal graph.
Consequently, deep Transformers exhibit a systematic position bias,
disproportionately attending to the initial tokens regardless of the input
context.

LSSAR fundamentally resolves this structural pathology by altering the
topology of the attention graph. By introducing the ReLU
in the re-weighting stage, LSSAR enforces hard sparsity, allowing
attention weights $A_{ij}$ to be exactly zero. This operation effectively severs
the edges in the attention graph that fall below the adaptive mean-based threshold.
Mathematically, this violates the strictly positive connectivity assumption required
for the convergence results in \citep{wu2025emergence}. By dynamically disconnecting
the edges between current tokens and irrelevant predecessors (including the
initial token when it serves no semantic purpose), LSSAR prevents the
probability mass from accumulating at the sequence start. Thus, the model is
liberated from the structural position bias, allowing it to distribute attention
solely based on semantic relevance.

\subsubsection{Geometric Interpretation: Preventing Manifold Drift}

While the graph-theoretic view explains the elimination of bias, the manifold
hypothesis \citep{bengio2013representation} offers a geometric explanation
for the robustness of LSSAR, which parallels recent findings in architectural
design such as Manifold-Constrained Hyper-Connections (mHC) \citep{xie2025mhc}.
The mHC framework addresses the instability of residual connections in deep
networks by projecting the weight matrices onto the Birkhoff polytope (the
manifold of doubly stochastic matrices). This geometric constraint strictly
preserves the norm and ensures that the residual mapping remains close to
an identity function, effectively preventing signal explosion and drift during
propagation.

Analogously, LSSAR enforces a topological manifold constraint on the
attention graph to prevent what we term `manifold drift'. In standard Softmax
attention, the mechanism aggregates information from the entire context window.
As the sequence length extends, the cumulative contribution of numerous
irrelevant tokens grows significantly. This off-manifold noise
pulls the aggregated hidden states away from the intrinsic semantic manifold.
This corresponds to the unconstrained instability observed in hyper-connections,
where unrestricted mixing leads to representation collapse.

By introducing the Shift-ReLU mechanism, LSSAR acts as a manifold-aware projection
operator. The hard sparsity constraint functions similarly to the boundary
conditions of the Birkhoff polytope in mHC: it strictly filters out off-manifold
noise vectors that fall below the adaptive threshold. By constructing a sparse,
dynamic $k$-Nearest Neighbours ($k$-NN) graph rather than a fully connected
dense graph, LSSAR ensures that the representation update is derived exclusively
from a local neighbourhood of semantically relevant tokens. This preservation
of local topology ensures that the hidden states remain firmly anchored to the
semantic manifold, irrespective of the sequence length. Consequently, LSSAR
achieves robust length extrapolation by fundamentally preventing the representations
from drifting into undefined regions of the latent space, just as mHC ensures
stability by constraining residuals to a well-defined geometric region.

\subsection{Asymptotic Gradient Behaviour}
\label{app:gradient_analysis}

In this section, we provide a rigorous analysis of the optimisation dynamics by examining the asymptotic behaviour of the gradients in the saturation regime. We demonstrate that the standard Softmax function suffers from exponential gradient decay, which fundamentally conflicts with the sharpening objective of the re-weighting mechanism. In contrast, we show that the LSSAR formulation exhibits polynomial gradient decay, maintaining a robust learning signal even under high sharpening factors.

To analyse the gradient flow, we trace the derivative from the re-weighted output back to the input logit $x$. The LSSAR pipeline introduces an intermediate filtering stage (Shift-ReLU) between the generator and the power operation. Let $a(x)$ be the normalised generator output, $r(a) = \text{ReLU}(a - \mu)$ be the filtered score, and $y = r^p$ be the final sharpened output. The total gradient is given by the chain rule:
\begin{equation}
    \frac{\partial y}{\partial x} = \frac{\partial y}{\partial r} \cdot \frac{\partial r}{\partial a}\cdot \frac{\partial a}{\partial x}
\end{equation}

We focus our analysis on the `winner' token regime (where $x_i \to \infty$). In this active regime, the token survives the filtering stage ($a > \mu$), meaning the ReLU operates in its linear region. Consequently, the filter gradient is $\frac{\partial r}{\partial a} \approx 1$. The dynamics are thus dominated by the interaction between the amplifier ($p \cdot r^{p-1}$) and the source gradient.

For the standard Softmax mechanism, the generator is the Softmax function. The derivative of the Softmax function with respect to its input is given by $\frac{\partial a_i}{\partial x_i} = a_i(1 - a_i)$. To rigorously substantiate the exponential decay of this source gradient, we expand the derivation of the term $(1 - a_i)$. By isolating the contribution of the target token $x_i$ from the aggregate score of all other tokens, denoted as $C = \sum_{j \neq i} e^{x_j}$, the expression for the complement probability becomes:
\begin{equation}
    1 - a_i = 1 - \frac{e^{x_i}}{e^{x_i} + C} = \frac{(e^{x_i} + C) - e^{x_i}}{e^{x_i} + C} = \frac{C}{e^{x_i} + C}
\end{equation}
In the regime where the model assigns high confidence to the winner token (i.e., as $x_i \to \infty$), the exponential term $e^{x_i}$ grows significantly larger than the constant interference term $C$. Consequently, the denominator is dominated by $e^{x_i}$, allowing us to approximate the expression as $\frac{C}{e^{x_i}} = C \cdot e^{-x_i}$. Since $\lim_{x_i \to \infty} a_i = 1$, the magnitude of the full derivative $a_i(1 - a_i)$ is entirely dominated by this decay term:
\begin{equation}
    \frac{\partial a_i}{\partial x_i} \approx 1 \cdot (C \cdot e^{-x_i}) \propto e^{-x_i}
\end{equation}
This exponential decay presents a fatal obstacle for the re-weighting mechanism. The total gradient becomes the product of a polynomial term (from the power operation) and an exponential decay term. Since exponential decay dominates polynomial growth for any finite $p$, the total gradient vanishes rapidly as the model becomes confident. This phenomenon, which we term \textit{exponential saturation}, effectively locks the weights and prevents further optimisation.

In stark contrast, LSSAR employs Softplus followed by $l_{1}$-normalisation. The derivative of the Softplus function is the Sigmoid function, $\sigma(x) = (1 + e^{-x})^{-1}$. As $x_i \to \infty$, $\sigma(x_i)$ asymptotically approaches 1. This upper bound signifies that the generator itself does not vanish or saturate in the high-confidence regime, unlike the exponential tail of Softmax. Consequently, $\text{Softplus}(x_i)$ approaches linearity ($\approx x_i$), and the normalised score approximates a harmonic function $a_i \approx x_i / (x_i + K)$, where $K$ represents the contribution of other tokens. The derivative of this function follows a polynomial decay:
\begin{equation}
    \frac{\partial a_i}{\partial x_i} \approx \frac{1 \cdot (x_i + K) - x_i \cdot 1}{(x_i + K)^2} = \frac{K}{(x_i + K)^2} \propto x_i^{-2}
\end{equation}
Crucially, because the intermediate Shift-ReLU operation behaves linearly for active tokens, it preserves this polynomial characteristic. The total gradient is thus the product of a polynomial growth term (from the power operation) and a polynomial decay term (from the source). Unlike the Softmax case, these terms can balance each other or even result in net gradient amplification. This structural property ensures that the optimisation pathway remains open even in the high-confidence regime, explaining why LSSAR can be stably trained with high sharpening factors where standard Softmax architectures fail.

\section{Further Experiments}

\label{app:further_experiments}
\label{app:training_config}

All models were trained using a sequence length of 1024 on the FineWeb-10B
dataset. We used GPT-2-124M (12 layers, 12 heads, and hidden size 768) with
RoPE, Adam with $\beta_1=0.9$ and $\beta_2=0.95$, a learning rate of $6\times10^{-4}$,
700 warmup steps, cosine decay, weight decay 0.1, and gradient clipping at 1.0.
Training used 8 NVIDIA A100 80GB GPUs, a micro-batch size of 4 per GPU,
gradient accumulation of 16, and PyTorch Automatic Mixed Precision (AMP) with
bfloat16 input precision. The resulting batch size was 524,288 tokens per
step, and the full run lasted 18,865 iterations, corresponding to approximately
9.89B training tokens.

Unless otherwise specified, all experiments outside the dedicated physical-law
and scaling analyses use this FineWeb-10B training setup with the GPT-2-124M
architecture. This includes the ablation studies, Softmax-free attention
comparisons, long-context extrapolation experiments, and attention-score
visualisations.

All baselines used the same training and evaluation setup; only the attention
mechanism was changed.

\subsection{Ablation Study}
\label{app:ablation}

\subsubsection{Modified Attention Scores}
\label{app:modified_attention}

To investigate the necessity of non-negative attention scores, we conducted experiments with modified Softmax outputs. As shown in \cref{tab:modifiedatt}, inverting the Softmax outputs by multiplying them by negative one resulted in a negligible change in validation loss, suggesting that non-negativity alone is not the determining factor. We also re-centred the attention scores by subtracting the row-wise mean, creating a mix of positive and negative values\footnote{To prevent training instability due to the absence of non-zero values in the initial rows, the first three rows were left unmodified.}. This led to a minor increase in loss. However, when we subsequently scaled these re-centred scores by the $l_{1}$-norm, performance was almost fully restored. Together with the decomposition study in \cref{tab:decomsoft}, these results suggest that Softmax's effectiveness comes from the combination of a nonlinear transformation and subsequent normalisation, with the $l_{1}$-norm playing a central role in preserving performance.

\subsubsection{Softmax Decomposition}
\begin{table}
    \centering
    \caption{Validation loss values for training the GPT-2 model from scratch with the standard and modified attention scores.}
    \begin{tabular}{lcccc}
        \toprule
             & Softmax & Inverse & Re-centred & Re-centred \& $l_{1}$-norm \\
        \midrule
        Loss & 3.1911  & 3.1915  & 3.2008     & 3.1954                     \\
        \bottomrule
    \end{tabular}
    \label{tab:modifiedatt}
\end{table}
\begin{table}
    \caption{Validation loss for GPT-2 model with and without the Softmax
        components: $e^{x}$ and $l_{1}$-norm.}
    \centering
    \begin{tabular}{ccccc}
        \toprule Scenario & I      & II         & III        & IV         \\
        \hline
        $e^{x}$           &        & \checkmark &            & \checkmark \\
        $l_{1}$-norm      &        &            & \checkmark & \checkmark \\
        \hline
        Loss              & 3.4138 & 3.247      & 3.3297     & 3.1911     \\
        \bottomrule
    \end{tabular}
    \label{tab:decomsoft}
\end{table}
As illustrated in \cref{equ:softmax}, the Softmax operation can be
decomposed into a non-linear transformation ($e^{x}$), followed by the $l_{1}$-norm.
In this section, we examine the importance of each component for LLMs by training
the GPT-2 model from scratch with and without each component.

The results presented in \cref{tab:decomsoft} provide quantitative evidence
for our theoretical analysis detailed in \cref{sec:design}. First, the
comparison between Scenario III and Scenario IV highlights the critical role
of the non-linear transformation. Scenario III, which applies the $l_{1}$-norm
directly to raw attention scores (effectively a form of linear attention), results
in a higher validation loss than Scenario IV. This indicates that without
the expansive mapping provided by the exponential function, the model struggles
to distinguish between relevant and irrelevant tokens, leading to an overly
flat attention distribution.

Most critically, the comparison between Scenario II and Scenario IV empirically
substantiates the vital role of the $l_{1}$-norm as a mechanism for \textit{lateral
    inhibition} and \textit{gradient coupling}. While better than the linear
baseline, Scenario II still lags behind the full Softmax. This performance
gap demonstrates that non-linearity alone is insufficient. Without the $l_{1}$-norm,
the attention mechanism lacks the `zero-sum' constraint required to induce
competition among tokens. Consequently, the superior convergence observed in
Scenario IV confirms that the gradient coupling provided by the $l_{1}$-norm
is indispensable for stabilising the optimisation process and preventing the
isolation of token representations. These findings justify our design choice
in LSSA to retain both a strong non-linear mapping (Softplus) and the $l_{1}$-norm
to preserve these essential dynamical properties.

\subsubsection{Comparison with Different Activation Functions}
\label{sec:compareact}

We modify the standard attention mechanism by introducing several novel
attention variants for comparative analysis. Specifically, these variants
are created by substituting $\phi$ in \cref{equ:newatt} with
several alternative activation functions: ReLU \citep{rumelhart1986learning},
ReLU$^{2}$ \citep{so2021searching}, ReLU6 \citep{howard2017mobilenets}, GeLU
\citep{hendrycks2016gaussian}, Sigmoid \citep{rumelhart1986learning},
Softplus \citep{zheng2015improving}, and Mish \citep{misra2019mish}.


The results presented in \cref{tab:activationcomp} indicate that all
activation functions, except for Softplus, result in poorer validation loss
values compared to the standard attention mechanism represented by $e^{x}$. This
performance disparity can be rigorously explained through our theoretical framework
regarding gradient coupling and mapping intensity.

First, ReLU-based functions (ReLU, ReLU$^{2}$, ReLU6) violate the principle of
\textbf{non-zero participation}. By mapping negative inputs strictly to zero,
these functions effectively remove a large portion of tokens from the
$l_{1}$-norm denominator. As discussed in \cref{sec:design}, this severs
the gradient pathway for these suppressed tokens, creating `dead neuron' that
cannot be optimised via lateral inhibition. Consequently, the model loses the
ability to recover potentially relevant information from the negative regime,
leading to suboptimal convergence.

Second, while Sigmoid is strictly positive, it is bounded and saturates at both
ends of the input range. When inputs enter these saturation regions, the gradient
becomes small, making the attention mechanism less sensitive to variations in
the latent representation. In contrast, Softplus and $e^{x}$ remain unbounded on
the positive side, which helps preserve gradient flow before normalisation.

In contrast, both Softplus and $e^{x}$ satisfy the two critical conditions
for effective attention: they are strictly positive (ensuring global
gradient coupling via the $l_{1}$-norm) and unbounded (preventing saturation).
Softplus, in particular, achieves the lowest validation loss. This slight advantage
over $e^{x}$ may be attributed to its asymptotic linearity in the positive domain,
which provides a more stable gradient flow compared to the exponential
growth of $e^{x}$, thereby balancing the need for feature separation with numerical
stability.
\begin{table*}[tb]
    \centering
    \caption{Validation loss for various activation functions employed in
        the attention mechanism. Note that $\phi=e^{x}$ corresponds to the conventional
        Softmax attention.}
    \begin{tabular}{ccccccccc}
        \toprule & ReLU   & ReLU$^{2}$ & ReLU6  & GeLU   & Sigmoid & Softplus        & Mish   & $e^{x}$ \\
        \hline
        Loss     & 3.2006 & 3.2494     & 3.2039 & 3.2051 & 3.2000  & \textbf{3.1901} & 3.2001 & 3.1911  \\
        \bottomrule
    \end{tabular}
    \label{tab:activationcomp}
\end{table*}

\subsubsection{Varying $p$ for Re-weighting Mechanism}
\label{sect:varyp}
\begin{figure}[tb]
    \centering
    \centering
    \includegraphics[width=0.5\linewidth]{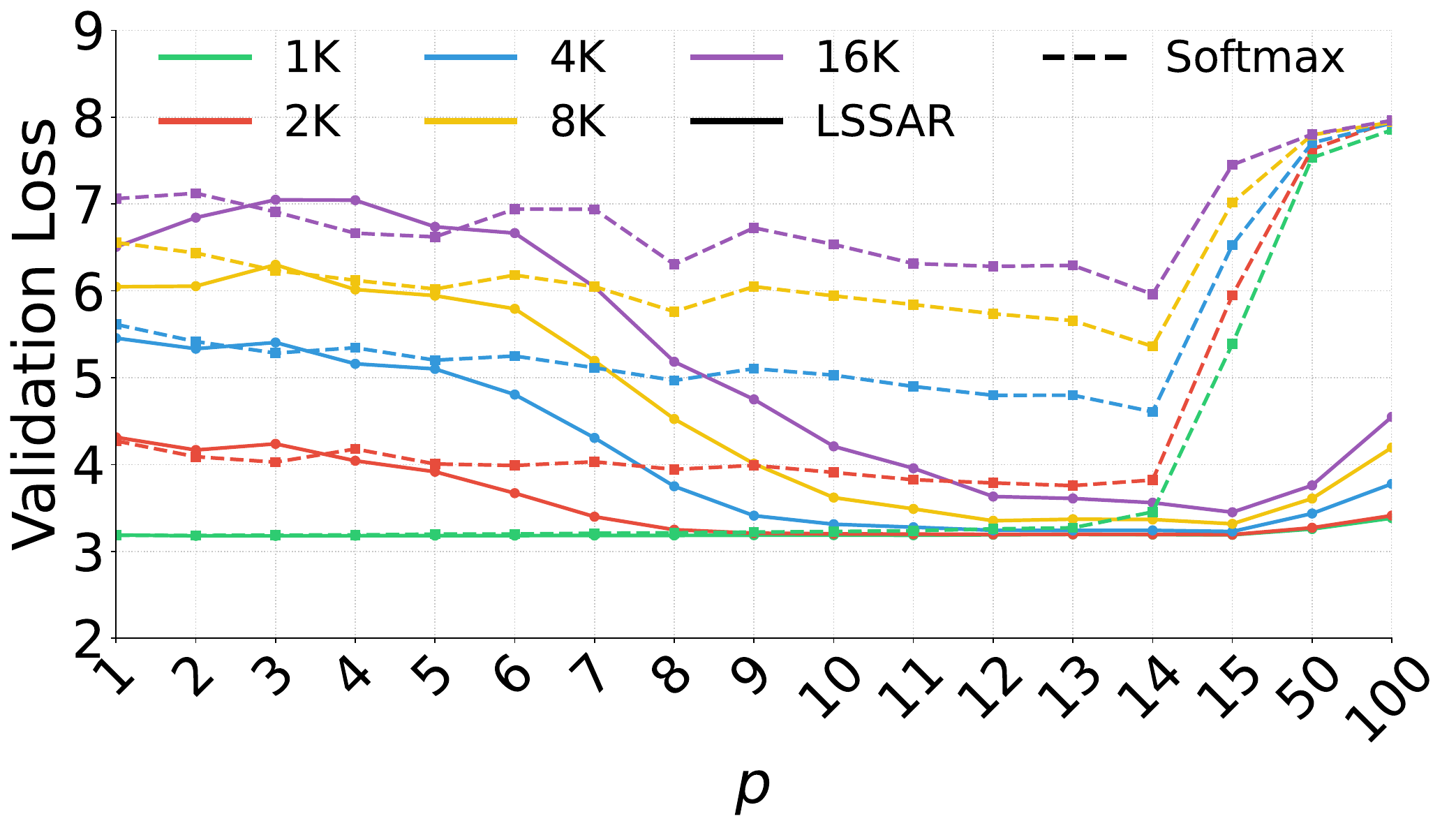}
    \caption{Comparison of LSSA and Softmax attention with varying
        values of $p$ for the re-weighting mechanism across different
        sequence lengths.}
    \label{fig:attcompare}
\end{figure}
To gain deeper insights into the proposed re-weighting mechanism, we
evaluated it using LSSA and Softmax attention across a range of $p$. As illustrated
in \cref{fig:attcompare}, LSSA and Softmax attention exhibit distinct
behaviours to changes in the re-weighting parameter $p$. While Softmax attention
performs comparably to LSSA at lower $p$ values, it suffers from severe
gradient explosion issues as $p$ increases to $p=15$ and beyond. This instability
manifests as a sharp rise in validation loss across all sequence lengths,
rendering the model practically unusable at these higher $p$ values.

In contrast, LSSA remains stable over the tested range of $p$ values. Increasing $p$ initially sharpens the re-weighted attention distribution and improves validation loss, with the best performance in this sweep observed around $p=15$. However, we do not view this value as a quantity derived from first principles. Very large $p$ can over-concentrate probability mass on a small number of entries, so $p$ should be treated as a tunable hyperparameter selected by validation performance for the target model and context length.

In practice, the preferred value of $p$ depends on both the training sequence length and the model architecture. In our experiments, longer training lengths tended to require larger $p$ values. As provisional starting points for tuning, we suggest $p \approx 13$ for length 1024, $p \approx 18$ for length 2048, and $p \approx 25$ for length 4096. These values are empirical heuristics rather than a universal closed-form rule. Additional adaptive-$p_i$ results are provided in Appendix~\ref{app:adaptive_p}.

Crucially, while Softmax suffers from exponential gradient decay (gradients $\propto e^{-x}$) that causes optimisation to stall at high $p$, LSSAR maintains polynomial gradient decay ($\propto x^{-2}$) due to Softplus's bounded derivative (Sigmoid $\leq 1$). This ensures stable gradient flow even with aggressive sharpening (see Appendix~\ref{app:gradient_analysis}). Although validation loss increases gradually beyond $p=15$ due to over-concentration on maximum values, LSSA handles large $p$ without numerical instability.

\subsection{Combined with Position Interpolation}
\label{app:PI}

To evaluate whether the proposed attention mechanism remains effective when
combined with a standard positional long-context technique, we conducted an
additional series of experiments using LSSAR together with Position
Interpolation (PI). Starting from the 1024-token LSSAR checkpoint ($p=15$), we
performed continued finetuning with PI enabled at target context lengths of 2K,
4K, and 8K on the same FineWeb-10B dataset.

The continued finetuning setup was kept simple in order to isolate the effect
of combining PI with the proposed attention mechanism. For each target length,
we trained for 2000 additional steps using a learning rate of $6\times10^{-5}$,
a 100-step warmup schedule, no further decay, and a total batch size of 524,288
tokens.

\begin{table}[tb]
    \centering
    \caption{Validation loss comparison between the base LSSAR model and the
        corresponding LSSAR+PI model after continued finetuning at each target
        context length. Lower is better.}
    \label{tab:lssar_pi_valloss}
    \begin{tabular}{lcc}
        \toprule
        Target length & Base LSSAR ($p=15$) & LSSAR ($p=15$) + PI \\
        \midrule
        2K            & 3.1930              & \textbf{3.1636}     \\
        4K            & 3.2291              & \textbf{3.1560}     \\
        8K            & 3.3171              & \textbf{3.1638}     \\
        \bottomrule
    \end{tabular}
\end{table}

\begin{table}[tb]
    \centering
    \caption{Passkey retrieval accuracy comparison between the base LSSAR model
        and the corresponding LSSAR+PI model after continued finetuning at each
        target context length. Accuracy is averaged over 100 trials. Higher is
        better.}
    \label{tab:lssar_pi_passkey}
    \begin{tabular}{lcc}
        \toprule
        Target length & Base LSSAR ($p=15$) & LSSAR ($p=15$) + PI \\
        \midrule
        2K            & 34\%                & \textbf{83\%}       \\
        4K            & 15\%                & \textbf{48\%}       \\
        8K            & 3\%                 & \textbf{21\%}       \\
        \bottomrule
    \end{tabular}
\end{table}

\paragraph{Validation Loss.}
The validation-loss results are reported in \cref{tab:lssar_pi_valloss}. At
all three target lengths, the LSSAR+PI model achieves lower validation loss than
the corresponding base LSSAR checkpoint. The improvement becomes more pronounced
as the target context length increases, indicating that PI complements the
proposed attention mechanism in the continued long-context finetuning regime.

\paragraph{Passkey Retrieval.}
The Passkey Retrieval results are reported in \cref{tab:lssar_pi_passkey}.
Consistent with the validation-loss comparison, combining PI with LSSAR yields
substantial gains at every target length. The largest relative improvements are
observed on the passkey task, where the 2K, 4K, and 8K accuracies increase from
$34\%$ to $83\%$, from $15\%$ to $48\%$, and from $3\%$ to $21\%$,
respectively. Although the absolute accuracy at 8K remains modest, this is
likely attributable to the limited finetuning budget of 2000 steps from a
1024-token checkpoint rather than to an incompatibility between the two methods.

Taken together, these results provide direct empirical evidence that LSSAR
remains effective when combined with PI, and that the gains from the proposed
attention mechanism are preserved under a more practical long-context
finetuning setup.

\subsection{Adaptive Position-Dependent $p_i$ Analysis}
\label{app:adaptive_p}

To further investigate whether the sharpening parameter should vary with
sequence position, we evaluated a position-dependent adaptive rule for query
row $i$:
\begin{equation}
    p_i =
    \begin{cases}
        15,                                                                          & i \le L_{\text{train}}, \\
        15 \cdot \left(1 + \alpha \left(\sqrt{i/L_{\text{train}}} - 1\right)\right), & i > L_{\text{train}},
    \end{cases}
\end{equation}
where $L_{\text{train}} = 1024$ and $\alpha$ controls the extrapolation
strength.

\begin{figure*}[tb]
    \centering
    \begin{minipage}[b]{0.49\textwidth}
        \centering
        \includegraphics[width=\linewidth]{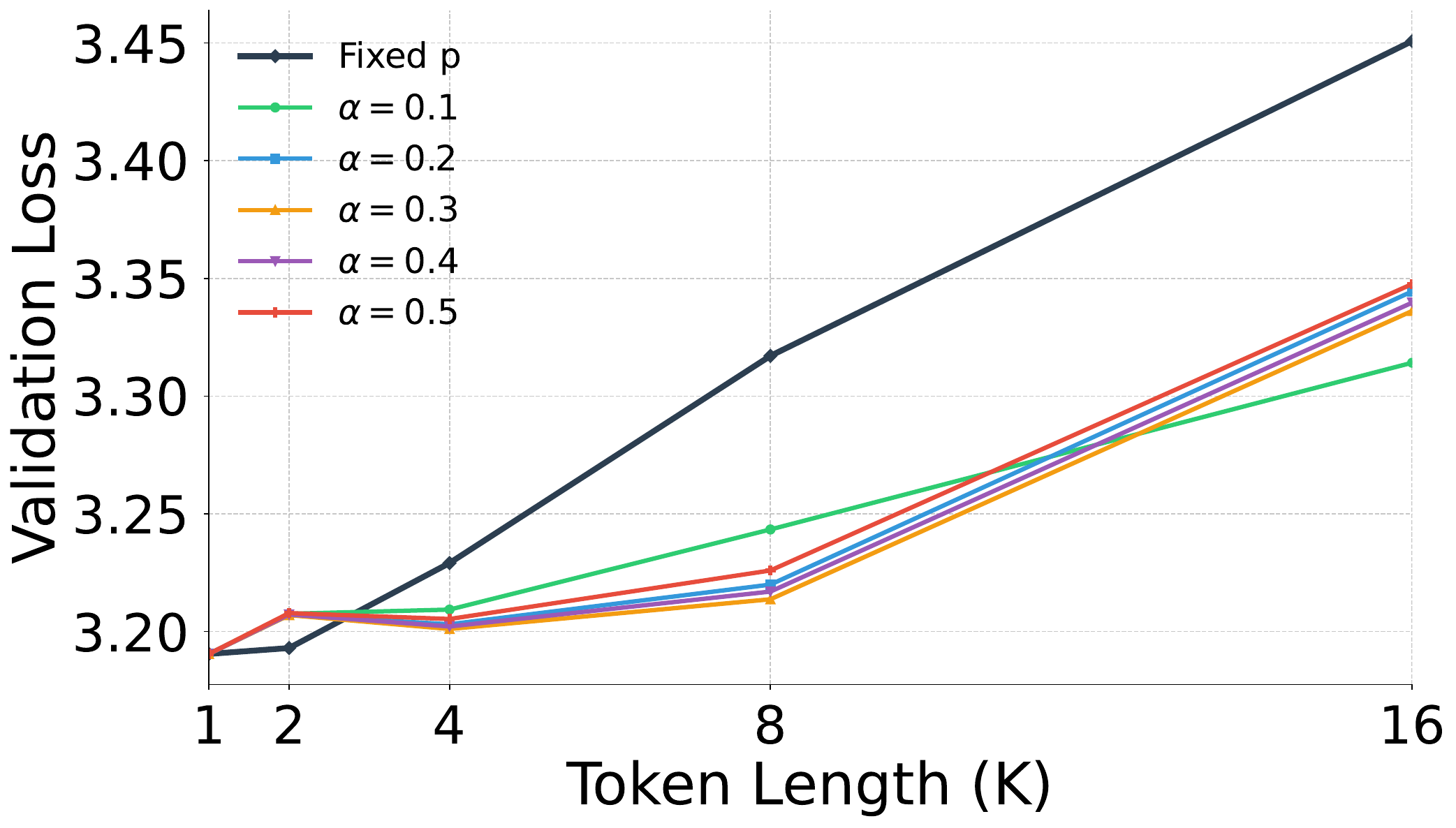}

    \end{minipage}
    \hfill
    \begin{minipage}[b]{0.49\textwidth}
        \centering
        \includegraphics[width=\linewidth]{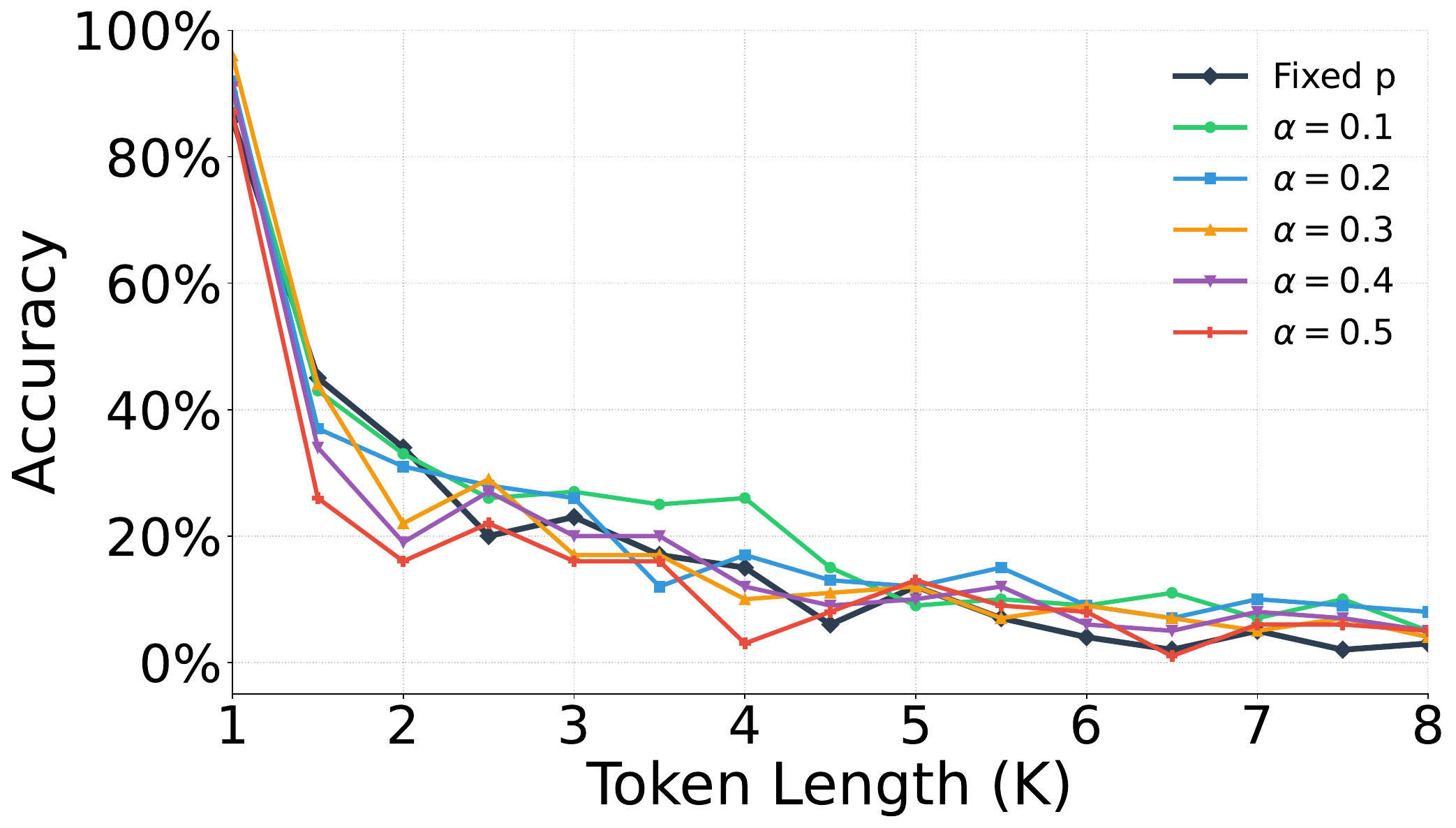}

    \end{minipage}
    \caption{Adaptive position-dependent $p_i$ analysis. Left: validation-loss
        extrapolation under different adaptive schedules. Right: passkey
        retrieval accuracy under the same schedules.}
    \label{fig:adaptive_p}
\end{figure*}

We evaluated $\alpha \in \{0.1, 0.2, 0.3, 0.4, 0.5\}$ and compared the
resulting models with the fixed-$p=15$ baseline. The results are shown in
\cref{fig:adaptive_p}. At extreme context lengths, the adaptive schedule
improves validation loss. For example, at $16\times$ extrapolation,
$\alpha=0.1$ reduces the validation loss from 3.45 to 3.31. This indicates
that allowing $p_i$ to increase gradually beyond the training horizon can
partially improve perplexity-style extrapolation.

However, the corresponding Passkey Retrieval results degrade in a manner
similar to the fixed-$p=15$ setting, indicating that adaptive sharpening alone
does not eliminate the dispersion issue. Taken together, these results suggest
that a length-dependent $p_i$ schedule can improve validation-loss
extrapolation, but does not by itself resolve the retrieval trade-off. We
therefore present this adaptive analysis as a practical extension of the main
hyperparameter study rather than as a complete solution.

\subsection{Computational Analysis}
\label{app:comp_analysis}

To evaluate the computational costs of our proposed methods, we conducted a
benchmark analysis. The results, detailed in Table~\ref{tab:comp_analysis}, compare
the performance of GPT-2-124M models with LSSA and LSSAR against the standard
Softmax attention baseline. For a fair comparison, all methods were implemented
using standard PyTorch functions without leveraging specialised fused CUDA kernels
(e.g., FlashAttention). Experiments were conducted on a single NVIDIA A100
GPU using bfloat16 precision with a batch size of 4 and a sequence length of
1024.

\begin{table}[tb]
    \centering
    \caption{Computational overhead comparison for different attention
        mechanisms. }
    \label{tab:comp_analysis}
    \begin{tabular}{@{}lcccc@{}}
        \toprule \multirow{2}{*}{Attention Mechanism} & \multicolumn{2}{c}{Training} & \multicolumn{2}{c}{Evaluation}                           \\
        \cmidrule(lr){2-3} \cmidrule(lr){4-5}         & Time (ms)                    & Memory (MB)                    & Time (ms) & Memory (MB) \\
        \midrule Standard Attention                   & 120.48                       & 9609.45                        & 49.26     & 2324.71     \\
        LSSA                                          & 169.92                       & 12071.23                       & 58.56     & 2325.58     \\
        LSSAR ($p=15$)                                & 250.32                       & 16685.57                       & 81.15     & 2325.58     \\
        \bottomrule
    \end{tabular}
\end{table}

The results in Table~\ref{tab:comp_analysis} show that the current PyTorch
implementation of LSSAR incurs substantial overhead relative to the standard
Softmax baseline. In our benchmark, LSSAR ($p=15$) increases evaluation latency
from 49.26 ms to 81.15 ms and training memory from 9609.45 MB to 16685.57 MB.
Its evaluation memory footprint remains nearly identical to the Softmax
baseline, suggesting that the largest memory overhead arises in the backward
pass, where the PyTorch Autograd engine must cache additional intermediate
tensors for operations such as $\text{ReLU}^{p}$. 

To examine whether an optimised implementation may be feasible, we analyse
LSSAR's theoretical complexity. From a computational standpoint, the
complexity of standard attention is dominated by two matrix multiplications,
resulting in $O(L^{2}d)$ floating-point operations (FLOPs). The additional
operations in LSSAR (norms, element-wise functions) are of a lower order ($O(
    Ld)$
or $O(L^{2})$), meaning that the asymptotic computational complexity of
LSSAR remains identical to standard attention.

The more critical aspect for future optimisation is memory (I/O) complexity. The primary bottleneck in
naive attention is the memory bandwidth required to read and write the large
$L \times L$ attention matrix to and from High-Bandwidth Memory (HBM). I/O-aware
algorithms like the FlashAttention family \citep{dao2022flashattention,dao2023flashattention,shah2024flashattention}
solve this by computing the output in tiles without ever materialising the full
matrix in HBM, reducing memory access complexity from $O(L^{2})$ to the
optimal $\boldsymbol{O(Ld)}$. Crucially, all additional operations in LSSAR
are local (element-wise or row-wise). This locality means they can be applied
to a sub-block (tile) of the attention matrix within fast on-chip SRAM.
Consequently, these operations may be amenable to fusion into the main loop of a
tiled attention algorithm, although we have not yet demonstrated such a fused
kernel in practice.

In summary, the locality of the additional operations suggests that LSSAR may be
compatible with future tiled or fused implementations, but we have not yet
demonstrated an optimised kernel or efficiency comparable to standard
high-performance attention. The present discussion should therefore be viewed as
preliminary feasibility analysis rather than a confirmed engineering result.

\subsection{Visualisation of Attention Scores}
\label{app:attmap}
This section provides a visual comparison of attention
maps generated by standard Softmax attention and our proposed LSSAR
mechanism. Following the default configuration in Appendix~\ref{app:training_config},
the attention maps are extracted from the final layer of GPT-2-124M models
trained on the FineWeb-10B dataset with a sequence length of 1024 tokens. The final layer is
chosen for this analysis as its attention patterns are most indicative of the
model's high-level understanding and directly influence the final output.
Comparing these maps offers a clear view of how each mechanism synthesises
information across the entire sequence.

As shown in \cref{fig:attmap}, the attention maps for all 12 heads produced by
standard Softmax attention (left panel) exhibit the attention sink phenomenon.
This is where the first token disproportionately attracts attention from
other tokens, irrespective of its semantic importance. This issue is
particularly pronounced in longer sequences, where the first token can dominate
the attention distribution, leading to suboptimal model performance. In contrast,
the attention maps generated by LSSAR (right panel) display a more balanced
distribution of attention across tokens. Notably, in the 4x3 grid of LSSAR
heads, those in the second column do not exhibit the attention sink phenomenon.
This demonstrates that the proposed method effectively mitigates this
problem, enabling the model to focus on more relevant tokens throughout the sequence.
This visual evidence corroborates our quantitative findings, demonstrating that
LSSAR not only enhances length extrapolation capabilities but also improves the
overall quality of the attention distributions in transformer models.

\setcounter{figure}{0}
\setcounter{table}{0}
\begin{figure}[tb]
    \centering
    \begin{minipage}[b]{0.48\linewidth}
        \centering
        \includegraphics[width=\linewidth]{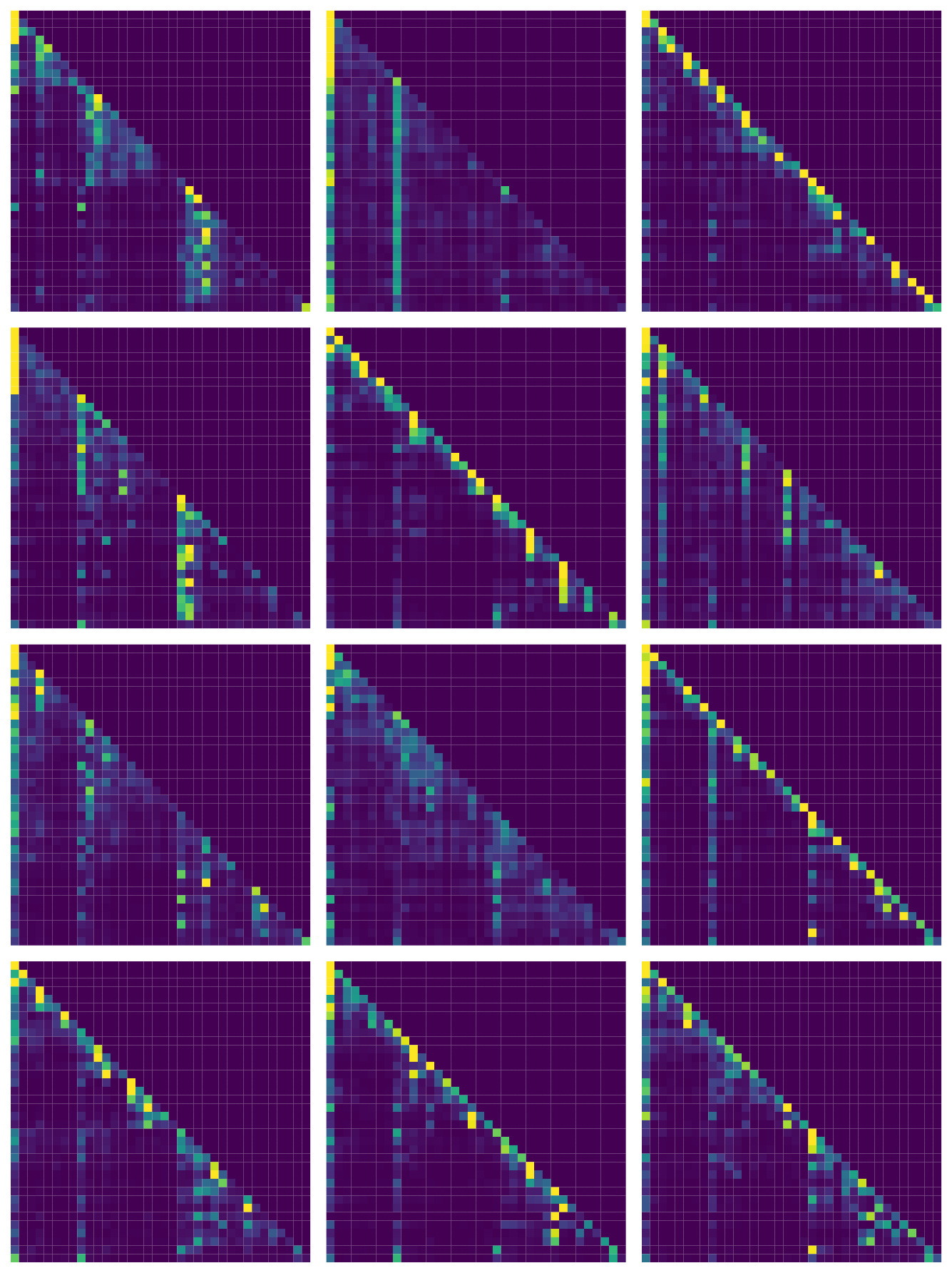}
    \end{minipage}
    \hfill
    \begin{minipage}[b]{0.48\linewidth}
        \centering
        \includegraphics[width=\linewidth]{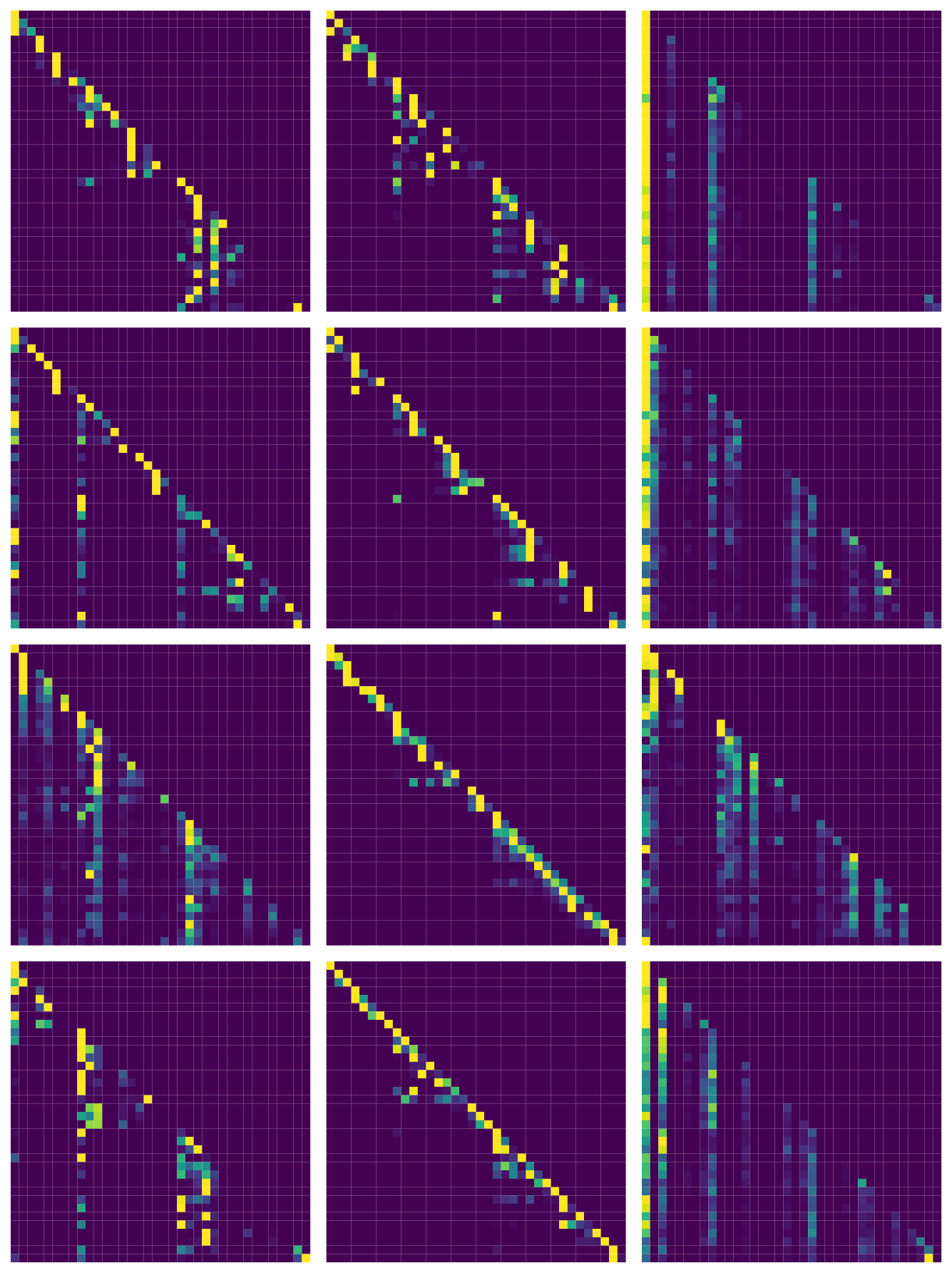}
    \end{minipage}
    \caption{Comparison of attention maps from the last layer of GPT-2-124M,
        showing standard Softmax attention (left) versus LSSAR with $p=15$ (right).
        Each panel displays the 12 attention heads in a 4x3 grid. For visualisation
        purposes, attention scores are clamped to the range [0, 0.5]. The input
        text is: \textit{Working from home can be great most days. I enjoy the flexibility
            and not having to commute in traffic. But sometimes I miss the office
            interactions with my coworkers and team meetings.}}
    \label{fig:attmap}
\end{figure}

\subsection{Experiments for Scaling with Filtered Data}
\label{app:scaling_experiments}
\begin{figure}[tb]
    \centering
    \begin{minipage}[b]{0.32\linewidth}
        \centering
        \includegraphics[width=\linewidth]{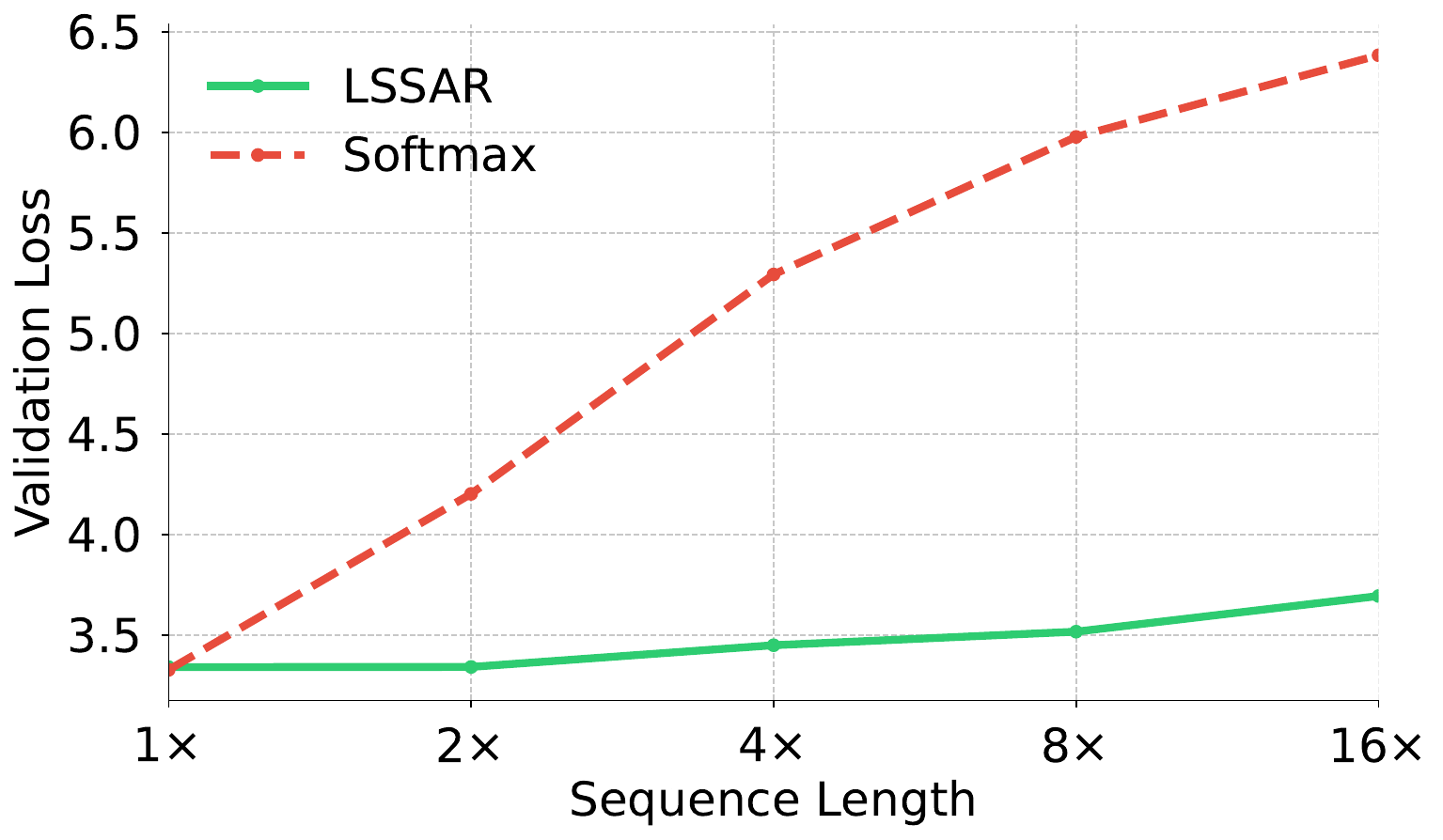}
    \end{minipage}
    \hfill
    \begin{minipage}[b]{0.32\linewidth}
        \centering
        \includegraphics[width=\linewidth]{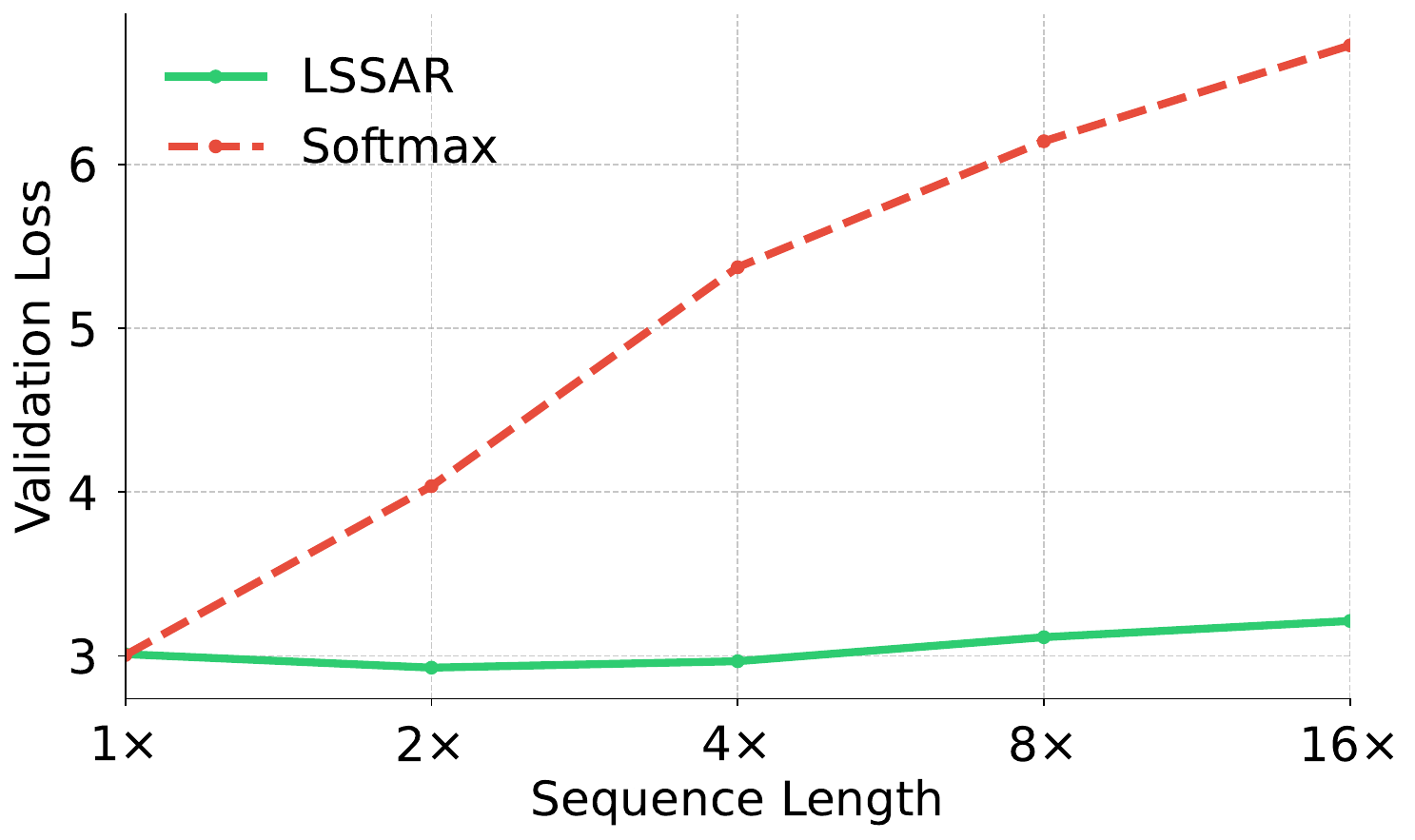}
    \end{minipage}
    \hfill
    \begin{minipage}[b]{0.32\linewidth}
        \centering
        \includegraphics[width=\linewidth]{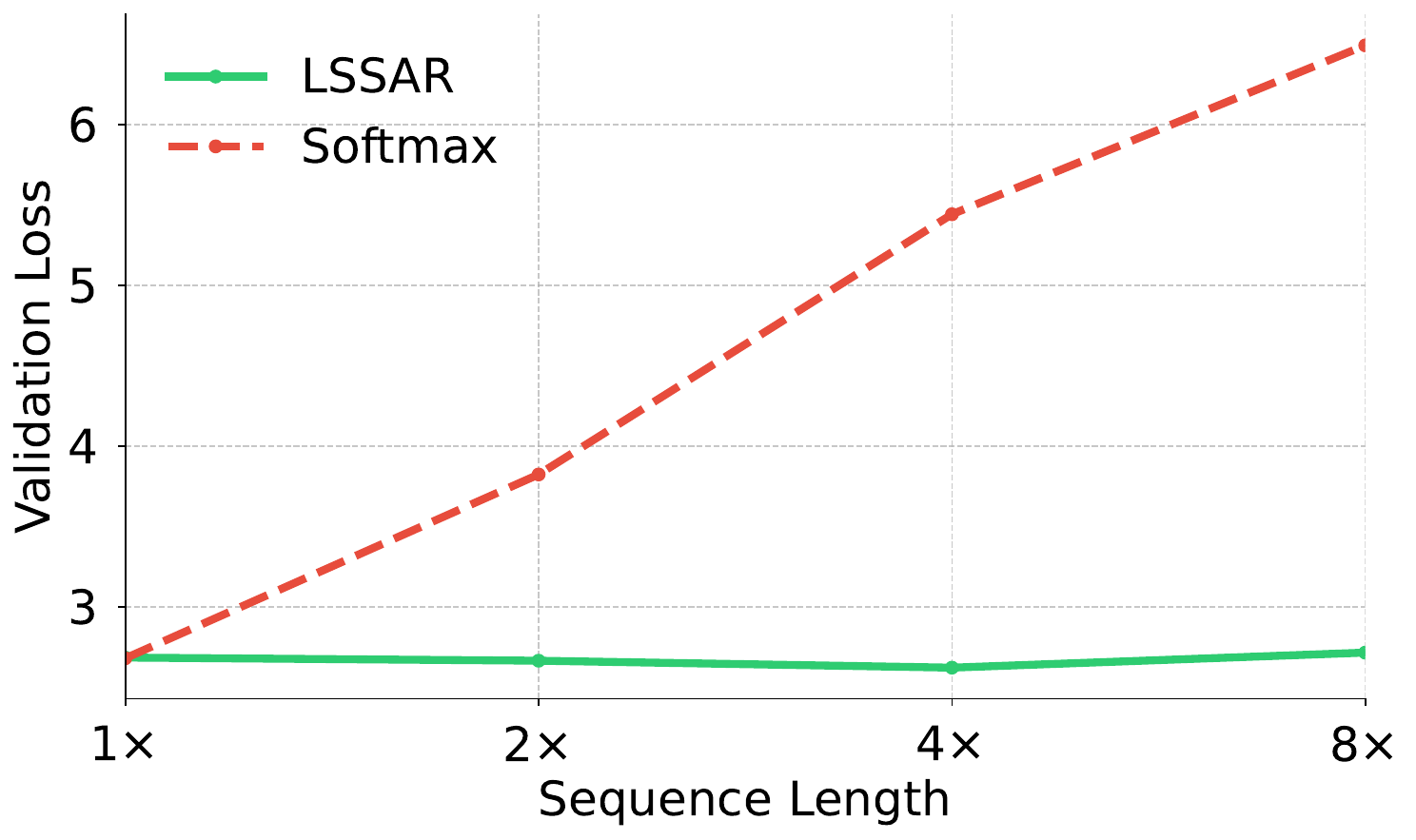}
    \end{minipage}
    \caption{Comparison of Softmax attention and LSSAR($p=15$) with validation
        loss extrapolation for GPT-2-45M (left), GPT-2-124M (middle) and GPT-2-355M
        (right).}
    \label{fig:edu_val_loss}
\end{figure}

\begin{figure}[tb]
    \centering
    \begin{minipage}[b]{0.32\linewidth}
        \centering
        \includegraphics[width=\linewidth]{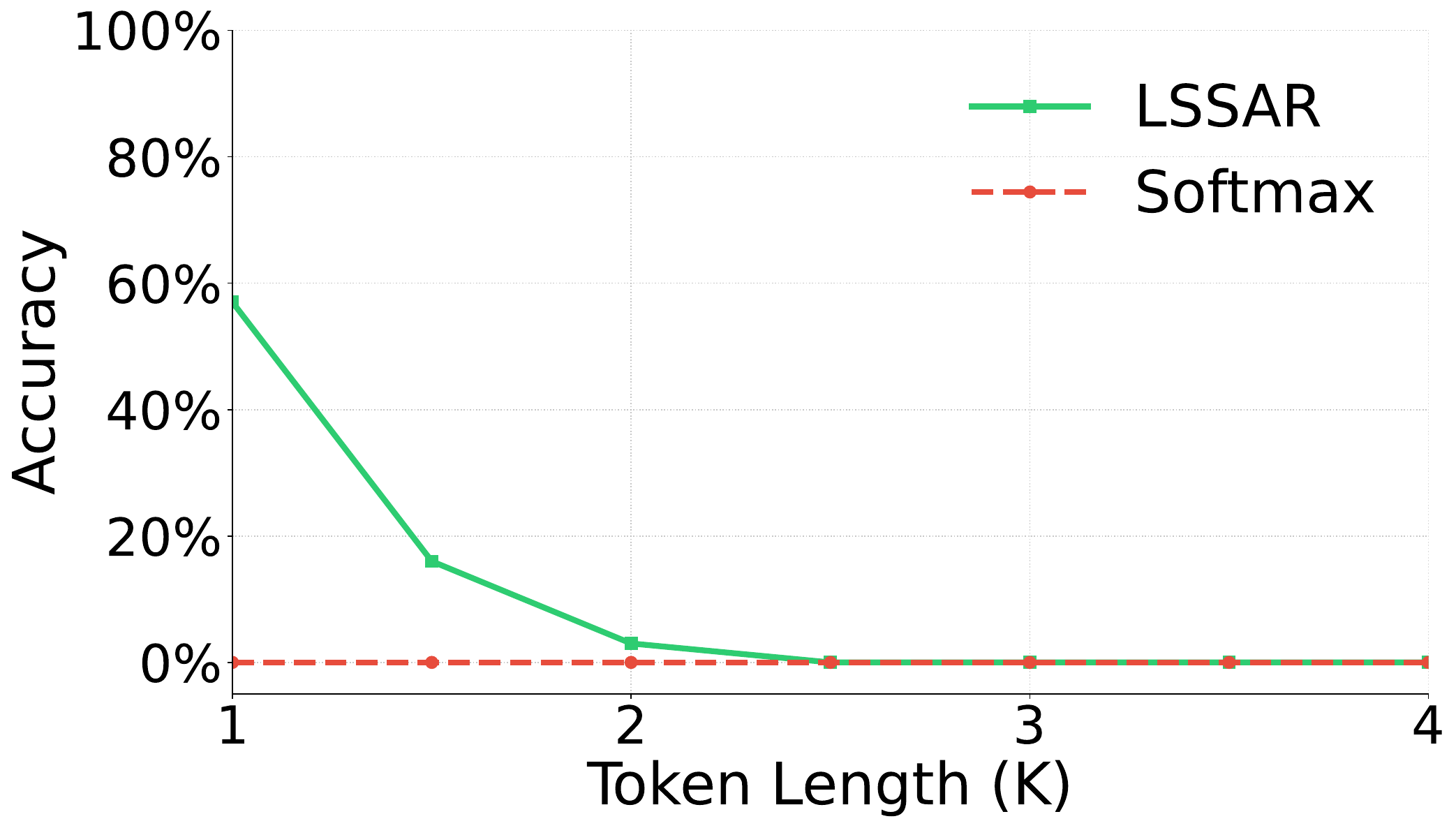}
    \end{minipage}
    \hfill
    \begin{minipage}[b]{0.32\linewidth}
        \centering
        \includegraphics[width=\linewidth]{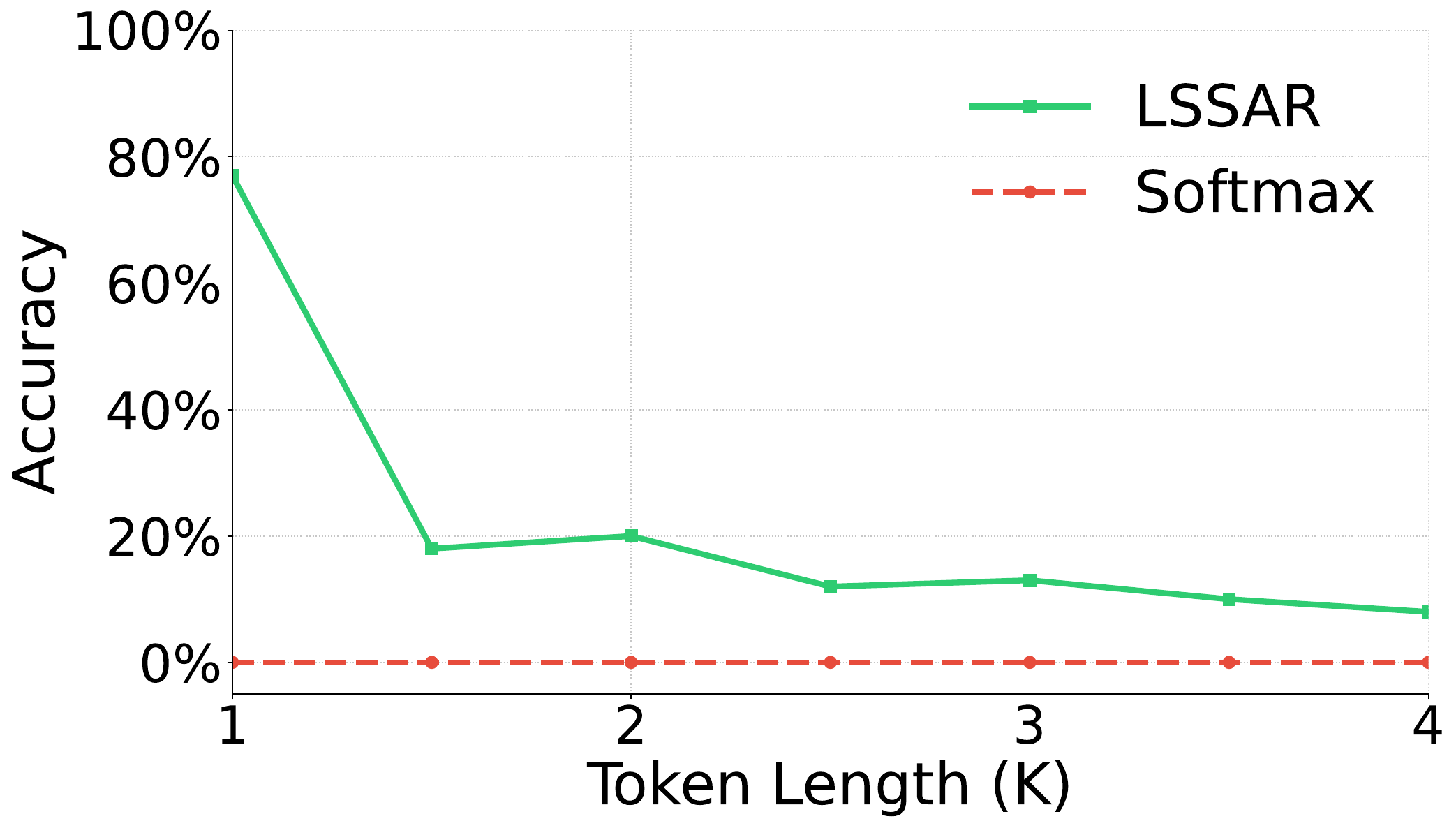}
    \end{minipage}
    \hfill
    \begin{minipage}[b]{0.32\linewidth}
        \centering
        \includegraphics[width=\linewidth]{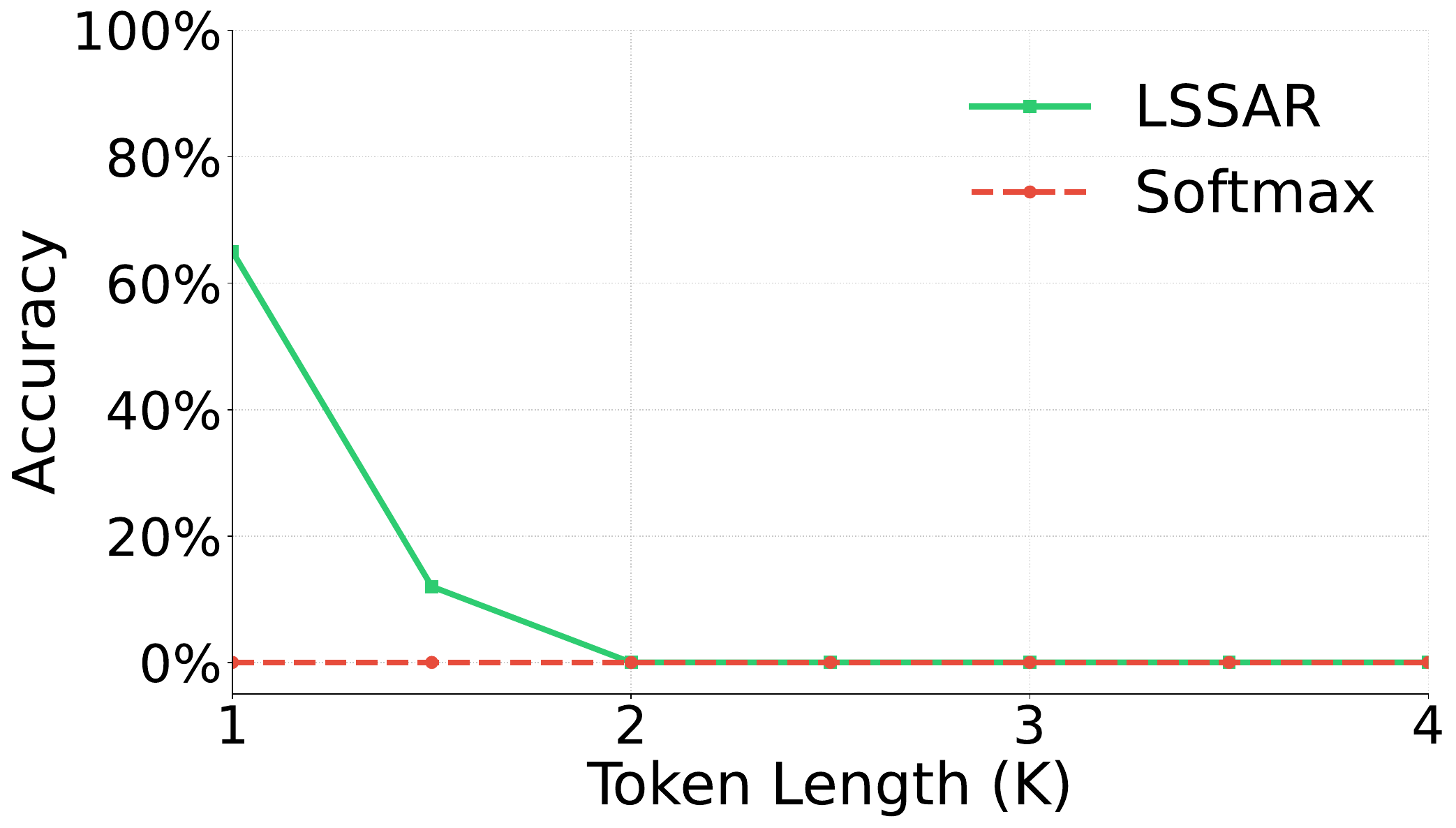}
    \end{minipage}
    \caption{Comparison of Softmax attention and LSSAR($p=15$) with passkey retrieval
        accuracy for GPT-2-45M (left), GPT-2-124M (middle) and GPT-2-355M (right).
        Accuracy was averaged over 100 trials with the passkey placed at random positions
        within the sequence.}
    \label{fig:edu_passkey}
\end{figure}
To evaluate the robustness of the proposed LSSAR mechanism across different
model scales and data distributions, we conducted a new series of experiments
using the FineWeb-Edu dataset \citep{penedo2024fine}. We specifically selected
FineWeb-Edu over the standard FineWeb dataset used in the main text because it
is rigorously filtered using Llama-3-70B, retaining only content with high
educational value and logical coherence. This selection serves a dual purpose.
On the one hand, it eliminates the potential influence of noisy training data
on downstream task performance, ensuring that the evaluation reflects the
intrinsic capabilities of the attention mechanism. On the other hand,
training on such a clean, high-quality dataset tends to drive models to focus
sharply on semantic and syntactic dependencies \citep{gunasekar2023textbooks}.
This characteristic paradoxically increases the difficulty of the Passkey Retrieval
task, as the model must attend to a ``passkey'' token that acts as semantic
noise within a highly coherent context. In this rigorous setting, the
ability of an attention mechanism to distinguish and retrieve the passkey
becomes a definitive test of its precision and extrapolation capabilities.

We trained three model configurations from scratch to investigate scaling
behaviours. The first is a 6-layer GPT-2-45M model with a configuration suggested
by the Pythia suite \citep{biderman2023pythia} for analysing scaling
behaviours. The rest are the standard GPT-2 architectures with 124M and 355M
parameters respectively. All models were modified to incorporate RoPE to
ensure a fair comparison with the state-of-the-art extrapolation baseline, and
were trained on the FineWeb-Edu 100B dataset with a sequence length of 1024
tokens.

The GPT-2-124M experiments follow the default training configuration in
Appendix~\ref{app:training_config}, except that the training corpus is replaced
by FineWeb-Edu 100B. The GPT-2-45M and GPT-2-355M runs use the same sequence
length, total batch size of 524,288 tokens, RoPE, weight decay, AMP/bfloat16
precision, and evaluation protocol. They differ from the GPT-2-124M setting only
in the model scale and the corresponding optimisation schedule: GPT-2-45M uses
a 6-layer configuration, a per-GPU batch size of 64, a learning rate of
$1\times10^{-3}$, and 9,000 training steps; GPT-2-355M uses a 24-layer
configuration, a per-GPU batch size of 32, a learning rate of $3\times10^{-4}$,
and 60,000 training steps.

\paragraph{Validation Loss Extrapolation.}
We first evaluated the language modelling performance on sequence lengths
extending far beyond the training context. As illustrated in Figure~\ref{fig:edu_val_loss},
the standard Softmax attention exhibits a significant degradation in
validation loss as the sequence length increases, failing to extrapolate effectively
even when trained on high-quality data. In contrast, LSSAR maintains a stable
and nearly constant validation loss across all tested lengths for both the 45M
and 355M models, demonstrating that its entropy invariance property holds
true regardless of model scale or data quality.

\paragraph{Passkey Retrieval Robustness.}
The results for the Passkey Retrieval task are presented in Figure~\ref{fig:edu_passkey}.
Consistent with our hypothesis regarding high-quality training data, the standard
Softmax baseline fails completely, yielding 0\% accuracy across all tested sequence
lengths, including the training window itself. The model's strong bias towards
coherent semantic structures prevents it from attending to the random passkey,
treating it effectively as noise. In stark contrast, LSSAR successfully
overcomes this limitation, achieving substantial retrieval accuracy within the
training length (57\% for GPT-2-45M, 77\% for GPT-2-124M and 65\% for GPT-2-355M)
and maintaining functional capabilities into the extrapolation regime. This
result confirms that the proposed re-weighting mechanism effectively
sharpens the attention distribution, enabling the model to capture critical high-entropy
information even when trained on ``textbook-quality'' data that discourages such
behaviour.

\paragraph{Downstream Task Performance.}
We evaluated zero-shot performance on standard downstream benchmarks. The results
are detailed in Table~\ref{tab:edu_downstream}. LSSAR demonstrates superior
performance compared to the Softmax baseline across the majority of tasks
for all model scales (45M, 124M, and 355M). While Softmax shows marginal advantages
in isolated cases (e.g., ARC-C and MMLU for the 124M model), LSSAR achieves
significant gains in tasks requiring long-range dependency and knowledge
retrieval. Notably, on the SummScreen benchmark, which requires processing long
contexts for summarisation, LSSAR achieves substantial improvements over
Softmax (approximately $2.7\times$ on 124M and $3.9\times$ on 355M). These gains
confirm that the architectural improvements of LSSAR translate into tangible
benefits for complex reasoning and summarisation tasks, without sacrificing
generic capabilities.

\begin{table}[h]
    \centering
    \caption{Zero-shot performance on downstream tasks for models trained on
        FineWeb-Edu. Best scores are bolded.}
    \label{tab:edu_downstream} \resizebox{\textwidth}{!}{
        \begin{tabular}{llccccccc}
            \toprule Model                       & Attention     & ARC-E          & ARC-C          & HellaSwag      & PIQA           & MMLU           & SciQ           & SummScreen      \\
            \midrule \multirow{2}{*}{GPT-2-45M}  & Softmax       & 45.58          & 16.72          & 27.34          & 58.98          & 22.94          & 70.40          & 0.8100          \\
                                                 & LSSAR($p=15$) & \textbf{46.04} & \textbf{18.34} & \textbf{27.49} & \textbf{59.79} & \textbf{22.97} & \textbf{75.00} & \textbf{2.1932} \\
            \midrule \multirow{2}{*}{GPT-2-124M} & Softmax       & 39.35          & \textbf{19.88} & 26.85          & 57.29          & \textbf{24.06} & 55.20          & 2.3313          \\
                                                 & LSSAR($p=15$) & \textbf{44.49} & 19.11          & \textbf{29.46} & \textbf{65.89} & 22.96          & \textbf{70.60} & \textbf{6.2825} \\
            \midrule \multirow{2}{*}{GPT-2-355M} & Softmax       & 59.30          & 23.46          & 30.63          & 66.49          & 22.98          & 78.90          & 2.4506          \\
                                                 & LSSAR($p=15$) & \textbf{62.50} & \textbf{26.96} & \textbf{34.81} & \textbf{68.34} & \textbf{23.94} & \textbf{84.00} & \textbf{9.5083} \\
            \bottomrule
        \end{tabular}
    }
\end{table}
\paragraph{Visualisation of Attention Scores.}
Finally, we visualise the attention scores of all models using the same
setting as in Appendix~\ref{app:attmap}. The results are presented in \cref{fig:attmap_d6},
\cref{fig:attmap_d12} and \cref{fig:attmap_d24}. The attention maps produced
by standard Softmax attention exhibit the attention sink phenomenon. In
contrast, the attention maps generated by LSSAR display a more balanced distribution
of attention across tokens. Notably, in the 4x4 grid of LSSAR heads with GPT-2-355M,
those in the first and third columns do not exhibit the attention sink
phenomenon. This demonstrates that the proposed method effectively mitigates
this problem, enabling the model to focus on more relevant tokens throughout
the sequence. This visual evidence corroborates our quantitative findings,
demonstrating that LSSAR not only enhances length extrapolation capabilities
but also improves the overall quality of the attention distributions in
transformer models.

\begin{figure}[tbp]
    \centering
    \begin{subfigure}
        [b]{\textwidth}
        \centering
        \includegraphics[width=\textwidth]{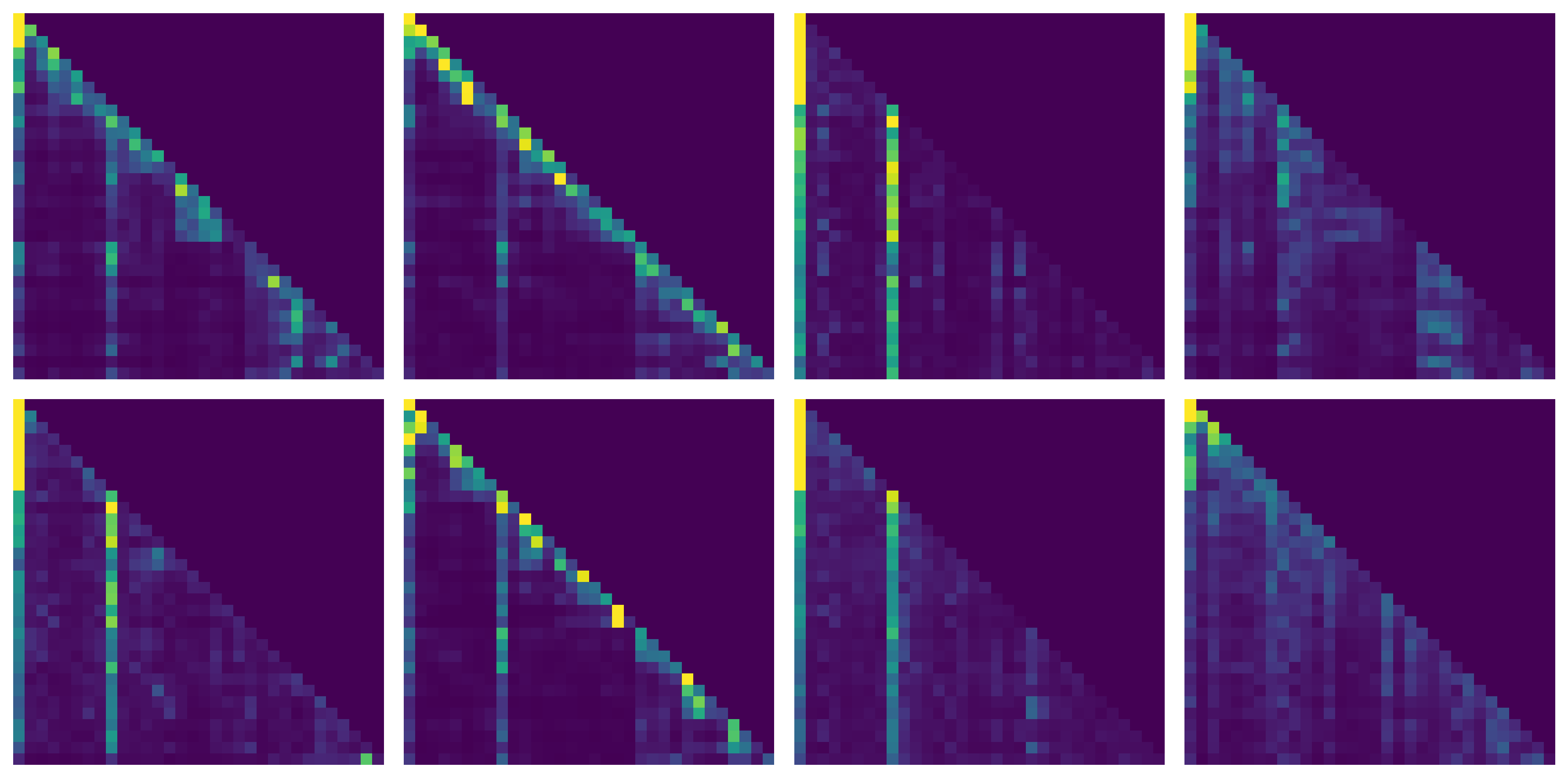}
        \caption{Softmax attention}
    \end{subfigure}
    \hfill
    \begin{subfigure}
        [b]{\textwidth}
        \centering
        \includegraphics[width=\textwidth]{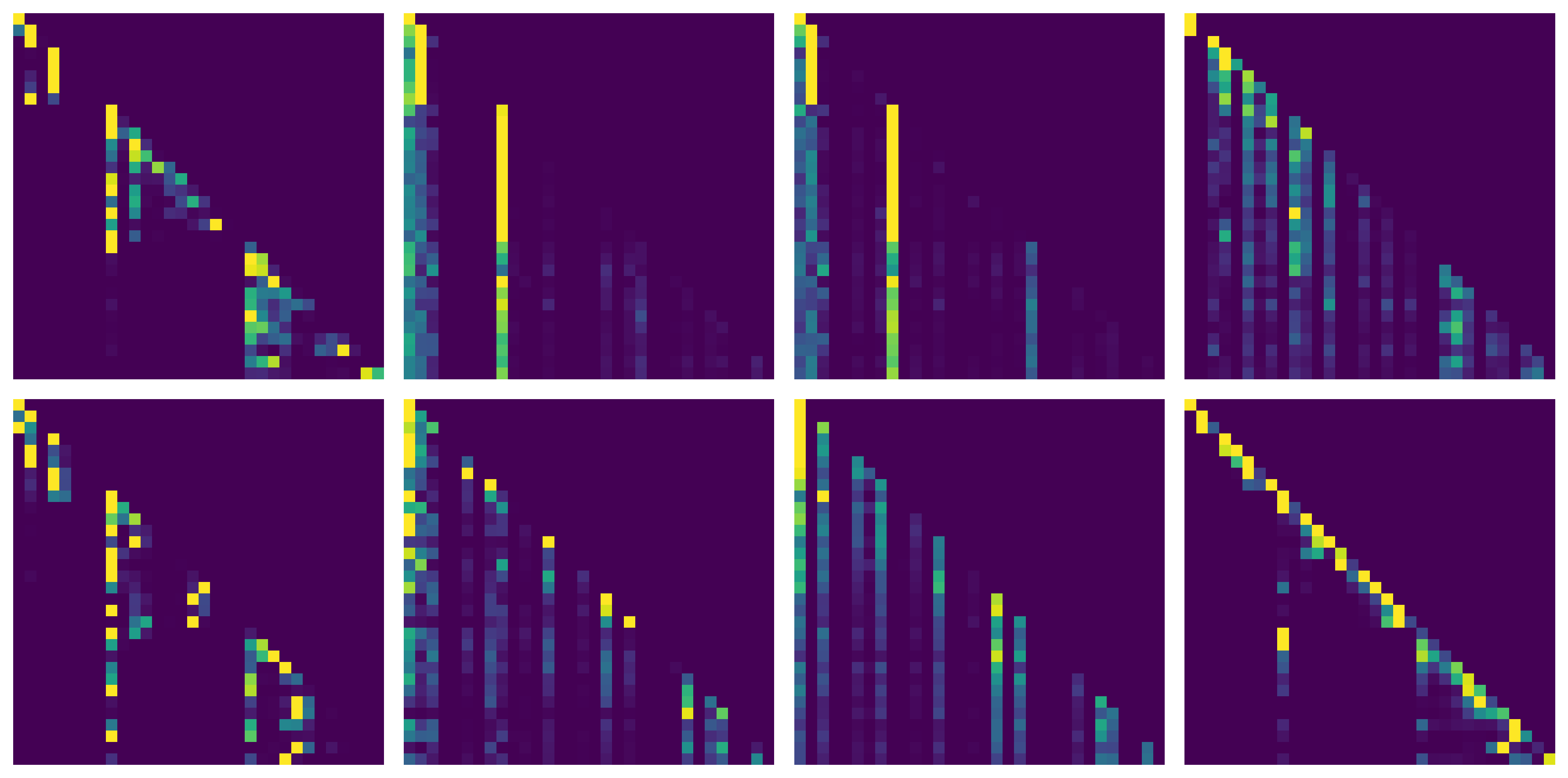}
        \caption{LSSAR($p=15$)}
    \end{subfigure}
    \caption{Comparison of attention maps from the last layer of GPT-2-45M, showing
        standard Softmax attention (above) versus LSSAR with $p=15$ (below).
        Each panel displays the 8 attention heads in a 4x2 grid. For visualisation
        purposes, attention scores are clamped to the range [0, 0.5].}
    \label{fig:attmap_d6}
\end{figure}

\begin{figure}[tbp]
    \centering
    \begin{subfigure}
        [b]{\textwidth}
        \centering
        \includegraphics[width=\textwidth]{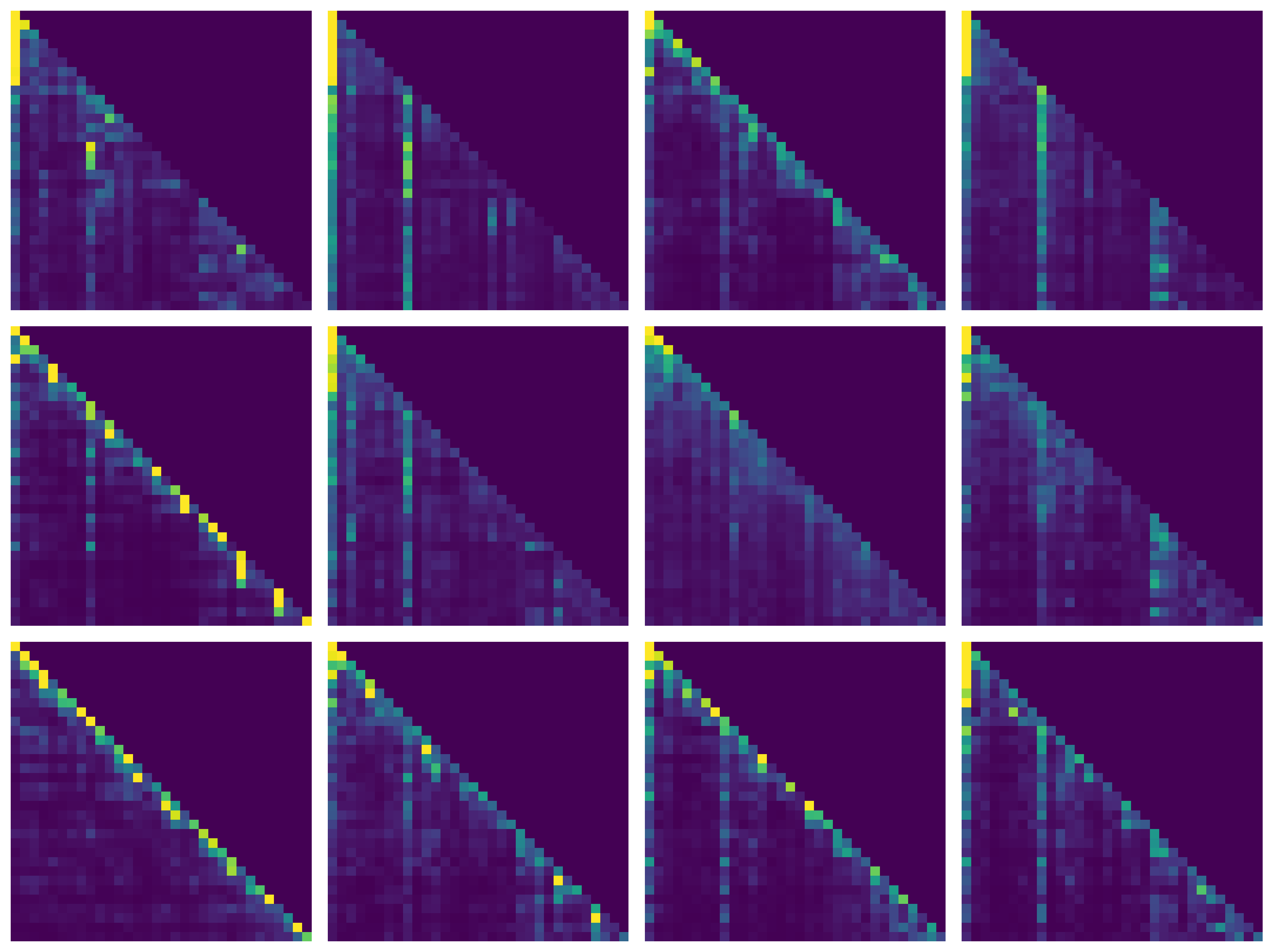}
        \caption{Softmax attention}
    \end{subfigure}
    \caption{Comparison of attention maps from the last layer of GPT-2-124M, (continued on next page).}
\end{figure}

\begin{figure}[tbp]
    \ContinuedFloat
    \begin{subfigure}
        [b]{\textwidth}
        \centering
        \includegraphics[width=\textwidth]{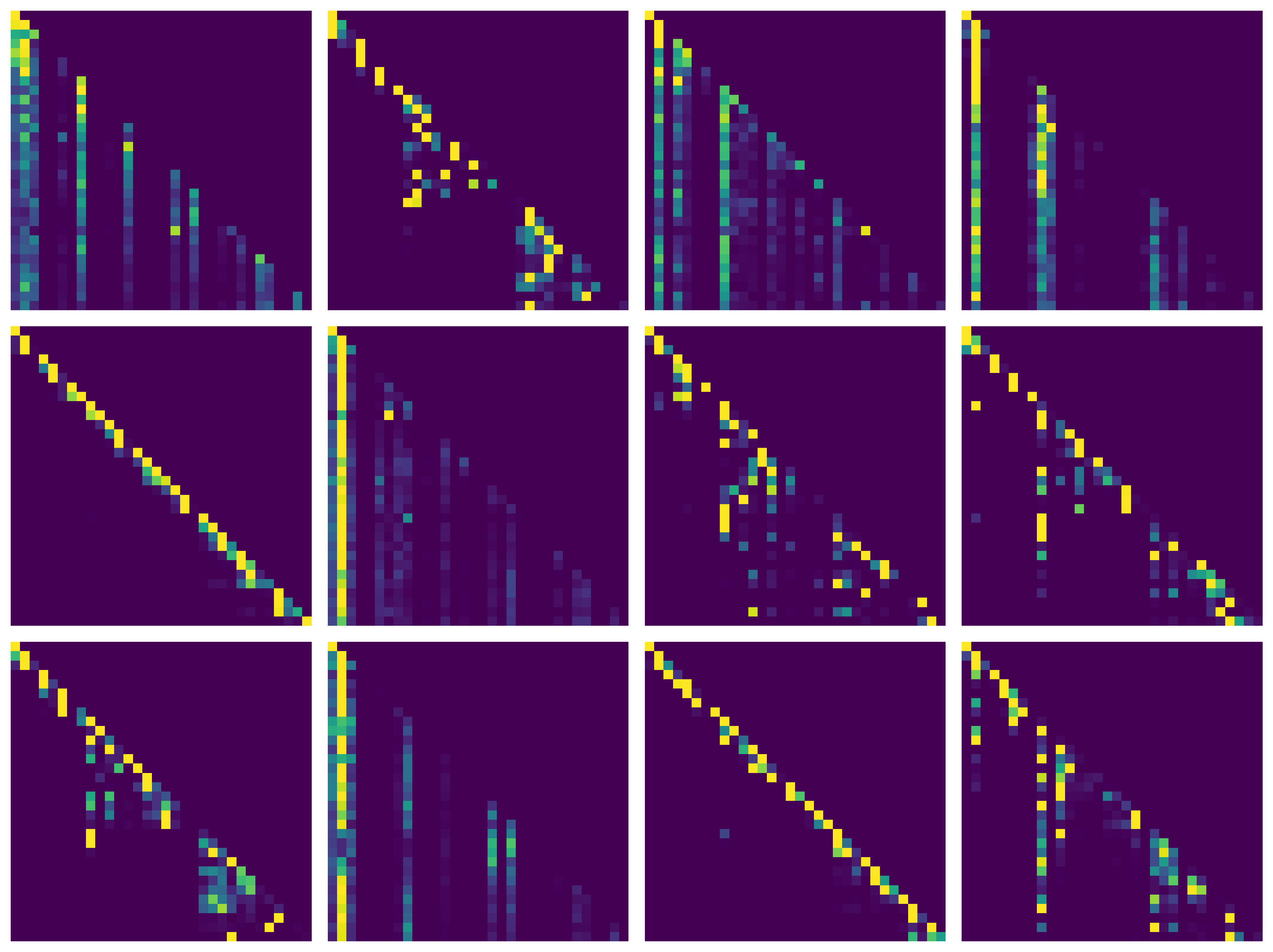}
        \caption{LSSAR($p=15$)}
    \end{subfigure}
    \caption{Comparison of attention maps from the last layer of GPT-2-124M,
        showing standard Softmax attention (above) versus LSSAR with $p=15$ (below).
        Each panel displays the 12 attention heads in a 4x3 grid. For visualisation
        purposes, attention scores are clamped to the range [0, 0.5]. }
    \label{fig:attmap_d12}
\end{figure}

\begin{figure}[tbp]
    \centering
    \begin{subfigure}
        [b]{\textwidth}
        \centering
        \includegraphics[width=\textwidth]{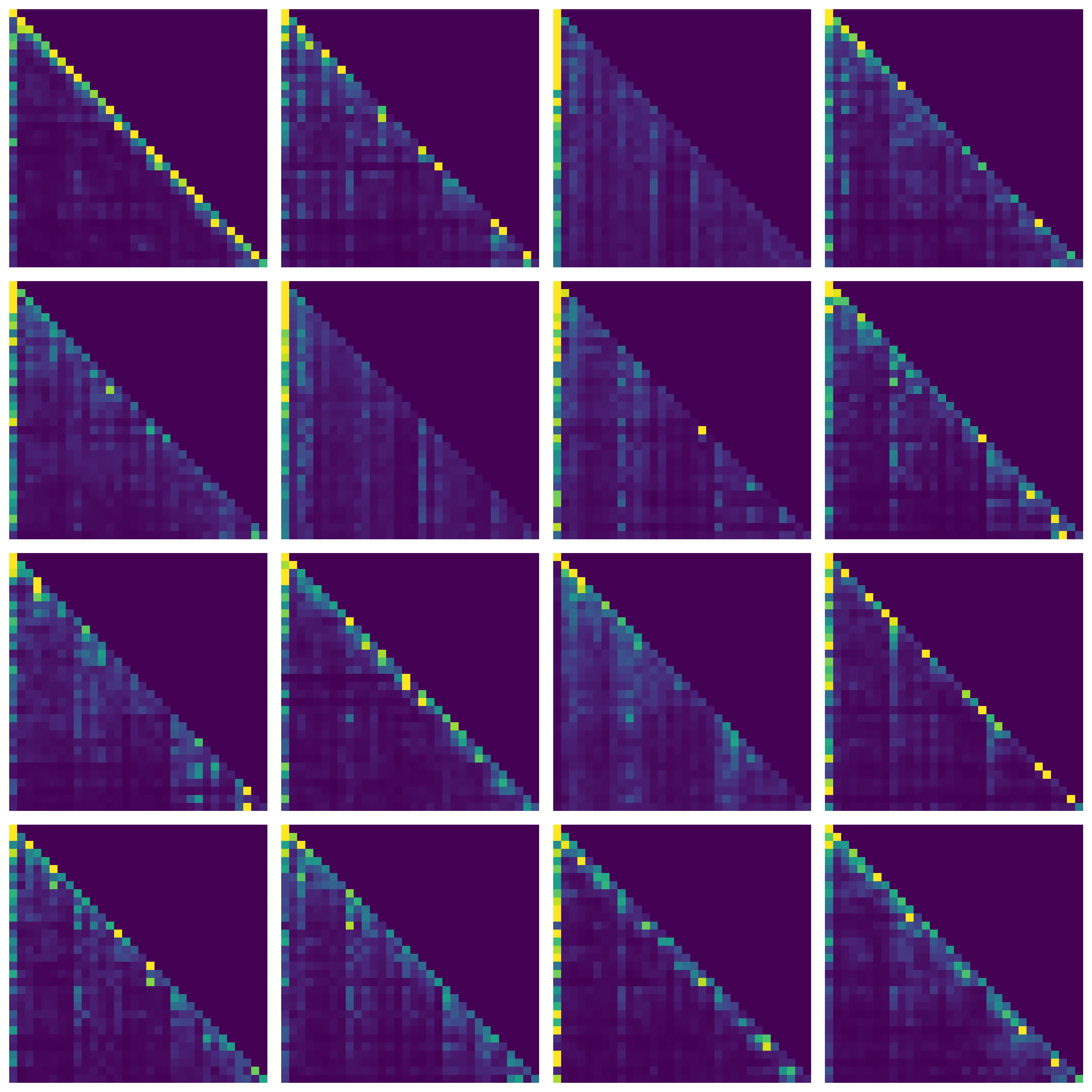}
        \caption{Softmax attention}
    \end{subfigure}
    \caption{Comparison of attention maps from the last layer of GPT-2-355M
        (continued on next page).}
\end{figure}

\begin{figure}[tbp]
    \ContinuedFloat
    \centering
    \begin{subfigure}
        [b]{\textwidth}
        \centering
        \includegraphics[width=\textwidth]{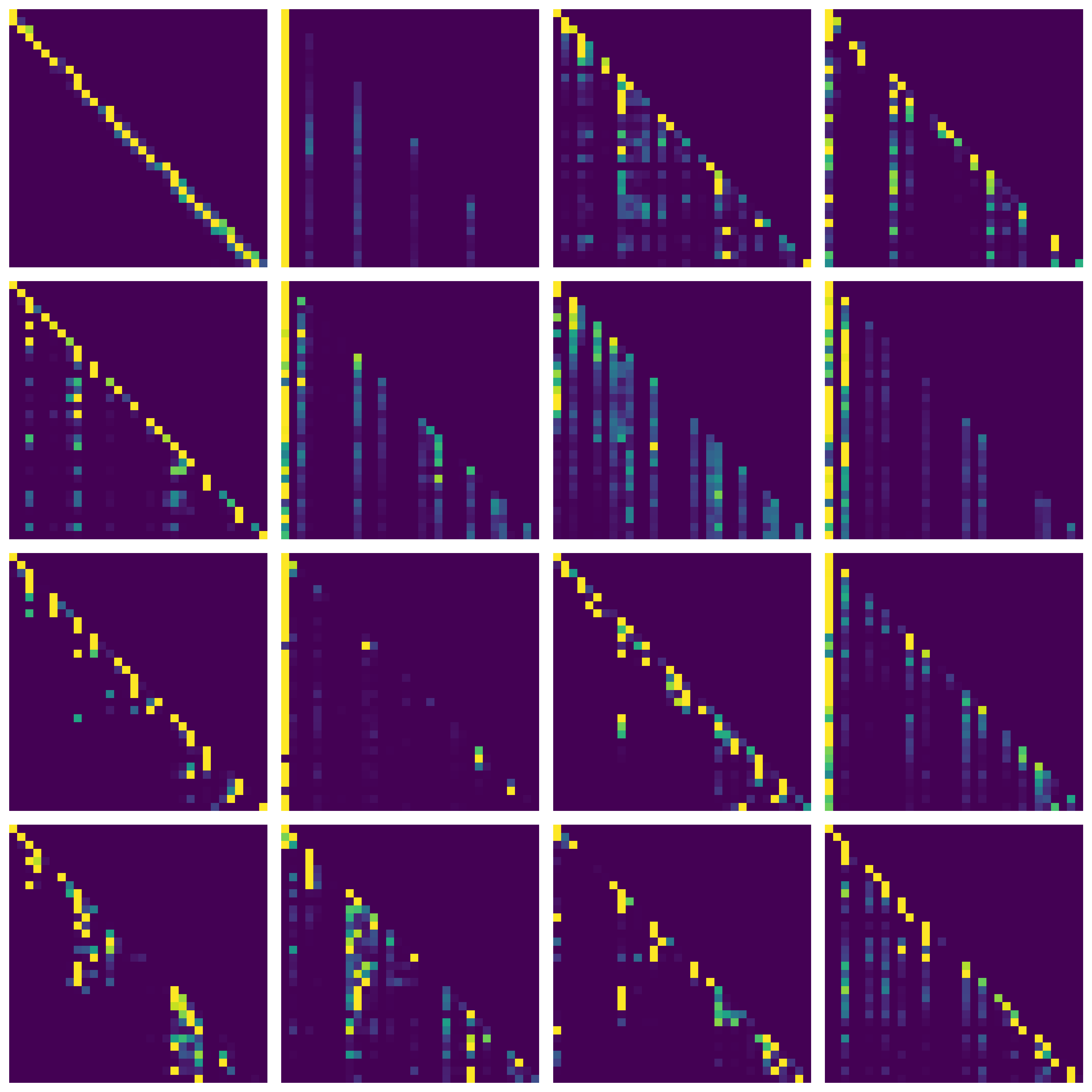}
        \caption{LSSAR($p=15$)}
    \end{subfigure}
    \caption{Comparison of attention maps from the last layer of GPT-2-355M,
        showing standard Softmax attention (above) versus LSSAR with $p=15$ (below).
        Each panel displays the 16 attention heads in a 4x4 grid. For visualisation
        purposes, attention scores are clamped to the range [0, 0.5].}
    \label{fig:attmap_d24}
\end{figure}

\subsection{Additional Evaluation on LAMBADA}
\label{app:lambada}

To further investigate the robustness of LSSAR against distribution shifts and
its capability to maintain long-range coherence, we evaluated our models on
the LAMBADA dataset (OpenAI split) \citep{paperno2016lambada}. This task
requires the model to predict the last word of a passage, necessitating a deep
understanding of the broader context rather than relying solely on local semantic
cues. The results, presented in \cref{tab:lambada_results}, reveal a
striking contrast between models trained on FineWeb-10B and FineWeb-Edu.

\begin{table}[h]
    \centering
    \caption{Zero-shot performance on LAMBADA for models trained on
        different datasets. Note that the 124M Softmax model trained on FineWeb-Edu
        exhibits severe collapse.}
    \label{tab:lambada_results}
    \begin{tabular}{llcc}
        \toprule Model Size (Dataset)                & Attention & Accuracy ($\uparrow$) & Perplexity ($\downarrow$) \\
        \midrule \multirow{2}{*}{45M (FineWeb-Edu)}  & Softmax   & 5.86                  & 1397.36                   \\
                                                     & LSSAR     & \textbf{18.57}        & \textbf{361.96}           \\
        \midrule \multirow{2}{*}{124M (FineWeb)}     & Softmax   & \textbf{31.51}        & \textbf{45.12}            \\
                                                     & LSSAR     & 30.52                 & 48.27                     \\
        \midrule \multirow{2}{*}{124M (FineWeb-Edu)} & Softmax   & 8.79                  & 10720.58                  \\
                                                     & LSSAR     & \textbf{30.37}        & \textbf{48.35}            \\
        \midrule \multirow{2}{*}{355M (FineWeb-Edu)} & Softmax   & 12.40                 & 263.69                    \\
                                                     & LSSAR     & \textbf{35.47}        & \textbf{33.27}            \\
        \bottomrule
    \end{tabular}
\end{table}

The results in \cref{tab:lambada_results} reveal a critical vulnerability in
standard Softmax attention. When comparing the 124M models, the version trained
on the standard FineWeb dataset achieves a respectable accuracy of 31.51\%. However,
the same architecture trained on the rigorously filtered FineWeb-Edu dataset
suffers a catastrophic collapse, with accuracy dropping to 8.79\% and
perplexity exploding to over 10,000. This indicates that while "textbook-quality"
data (FineWeb-Edu) may improve performance on scientific benchmarks, it induces
a severe bias in Softmax attention that destroys its ability to handle the narrative,
long-range dependencies required by LAMBADA. The standard Softmax mechanism appears
unable to generalise from the highly coherent training distribution to out-of-distribution
contexts.

In stark contrast, LSSAR demonstrates exceptional stability and invariance
to data distribution shifts. For the 124M model, LSSAR achieves similar
performance (around 30.4\% accuracy) regardless of whether it was trained on
FineWeb or FineWeb-Edu, completely avoiding the collapse observed with
Softmax. Furthermore, the 355M LSSAR model effectively leverages the
increased capacity provided by the FineWeb-Edu data, achieving the best overall
performance. This confirms that LSSAR's re-weighting mechanism successfully
counteracts the overfitting risks associated with highly filtered data, ensuring
that the model maintains precise, long-range attention capabilities
essential for robust reasoning and extrapolation.

\subsubsection{Implications for Reasoning Models}
\label{app:reasoning}

Recent advancements in reasoning models, such as DeepSeek-R1 \citep{deepseekai2025deepseekr1},
have demonstrated that reinforcing a ``Chain of Thought'' (CoT) process can
lead to emergent thinking capabilities. These models rely on generating extensive
intermediate reasoning steps to solve complex problems. The properties of
LSSAR discussed in this paper suggest it could be particularly advantageous
for such architectures.

The core mechanism of reasoning models involves extending the sequence
length to accommodate profound logical derivations. Standard Softmax attention
suffers from attention smoothing as sequence length increases, which dilutes
the model's focus and limits the effective depth of reasoning. LSSAR, with
its entropy-invariant scaling and sharpening stage, maintains distinct attention
peaks regardless of sequence length. This theoretically enables models to
sustain coherent thinking processes over indefinitely long sequences without
losing track of critical premises established early in the chain.

Furthermore, logical deduction is inherently precise. A conclusion often
hinges on a specific, discrete piece of information rather than a diffuse context.
The re-weighting mechanism in LSSAR forces the model to make decisive
attention allocations, effectively filtering out noise and ensuring that the
reasoning process attends to the exact tokens required for the next logical step.
As evidenced by the Passkey Retrieval and LAMBADA results on FineWeb-Edu, this
precision is maintained even when the model is trained on highly coherent data
that typically induces smoothing in Softmax models.

Finally, reasoning models, which often employ RoPE, are susceptible to the
attention sink phenomenon, where a significant portion of attention capacity
is wasted on the initial tokens \citep{xiaoefficient}. LSSAR mitigates this issue
(Appendix~\ref{app:attmap}), promoting a more balanced distribution where all
attention heads are utilised for semantic processing. For reasoning-heavy models,
this efficient utilisation of attention capacity is crucial for capturing
the complex, multi-faceted relationships inherent in deep logical tasks.

Building upon the foundations of R1, the recently released DeepSeek-V3.2
\citep{deepseekv32} introduces DeepSeek Sparse Attention (DSA) and advanced Group
Relative Policy Optimisation (GRPO) to further push the frontiers of reasoning.
LSSAR offers complementary advantages to these architectural innovations, particularly
in enhancing the efficacy of sparse attention mechanisms. While DSA improves
computational efficiency by selecting a top-$k$ subset of tokens via a
lightning indexer, the standard Softmax applied within this retrieved subset
remains liable to smooth attention scores, particularly as $k$ increases or
the context expands. Replacing the internal Softmax of DSA with LSSAR would provide
a crucial sharpening effect within the sparse window, ensuring that the
model attends to the most relevant tokens among the candidates, thereby
maximising the information density of the sparse selection.

Furthermore, LSSAR contributes to the stabilisation of large-scale
reinforcement learning. DeepSeek-V3.2 employs complex stabilisation
techniques for GRPO, such as unbiased KL estimation, to counteract the
instability often observed with exponential activations. LSSAR relies on the
Softplus activation, which exhibits linear asymptotic growth, as opposed to the
exponential growth of Softmax. This fundamental change in gradient dynamics
offers improved numerical stability, potentially mitigating the policy gradient
variance inherent in large-scale RL training and offering a more robust optimisation
landscape for methods like GRPO.

Crucially, LSSAR addresses the token efficiency gap observed between open
and proprietary models. As highlighted in \citet{deepseekv32}, the DeepSeek-V3.2-Speciale
model often requires substantially longer reasoning trajectories to match
the performance of Gemini-3.0-Pro, a phenomenon attributed to ``redundant self-verification''
and lower ``intelligence density''. We posit that this redundancy stems from
the uncertainty induced by attention smoothing in Softmax, which necessitates
verbose verification loops. By enforcing a more decisive attention
distribution through its re-weighting mechanism, LSSAR theoretically enables
the model to retrieve premises with higher confidence, thereby compacting
the reasoning chain. This improvement directly alleviates the context bottleneck
in agentic workflows; by reducing trajectory bloat and leveraging LSSAR's
proven length extrapolation capabilities (Appendix~\ref{app:scaling_experiments}), future
agents could retain full interaction histories without resorting to destructive
context management strategies like ``Discard-all''.

\subsection{Force Predictions for Other Planets}
\label{app:symbolic_regression_planets}
\begin{figure}[tb]
    \centering
    \begin{subfigure}[b]{0.48\textwidth}
        \centering
        \includegraphics[width=\linewidth]{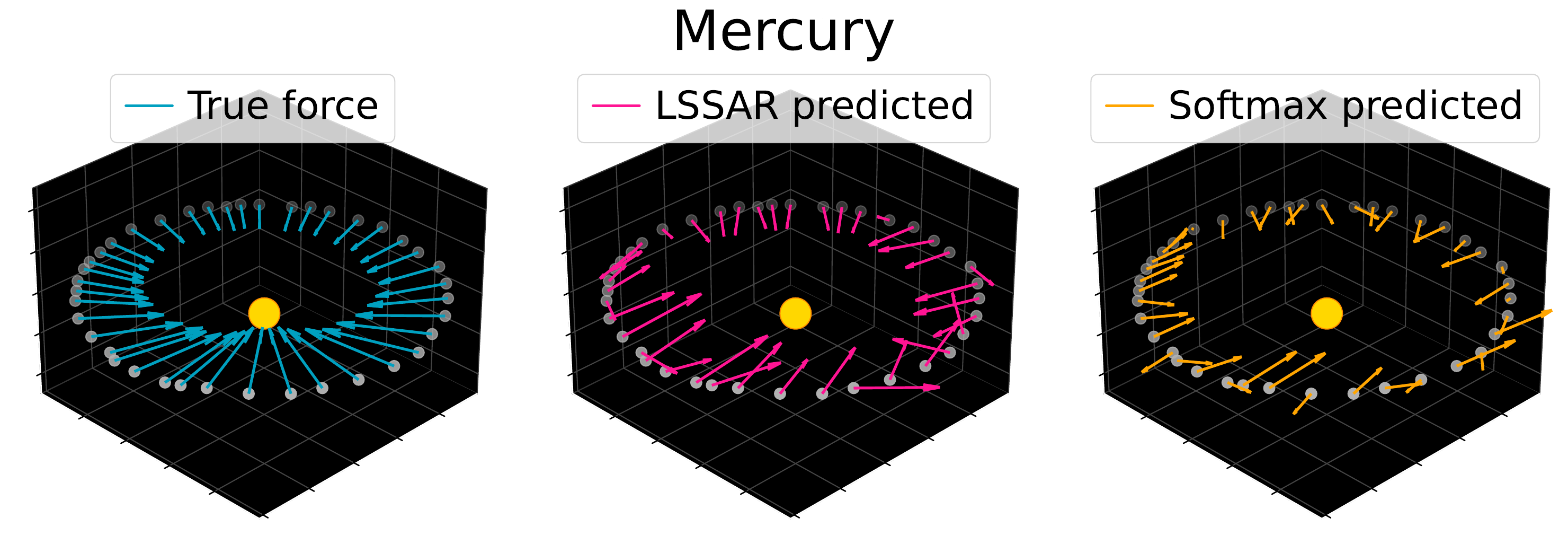}

        \label{fig:force_mercury}
    \end{subfigure}
    \hfill
    \begin{subfigure}[b]{0.48\textwidth}
        \centering
        \includegraphics[width=\linewidth]{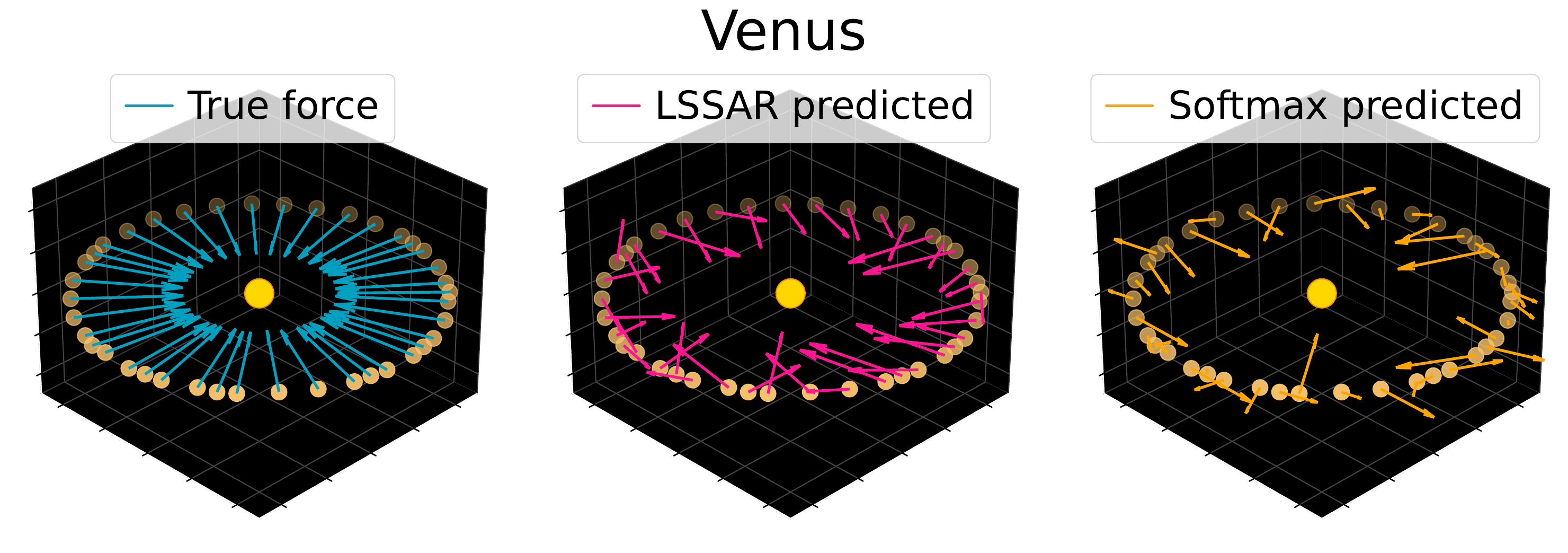}

        \label{fig:force_venus}
    \end{subfigure}

    \begin{subfigure}[b]{0.48\textwidth}
        \centering
        \includegraphics[width=\linewidth]{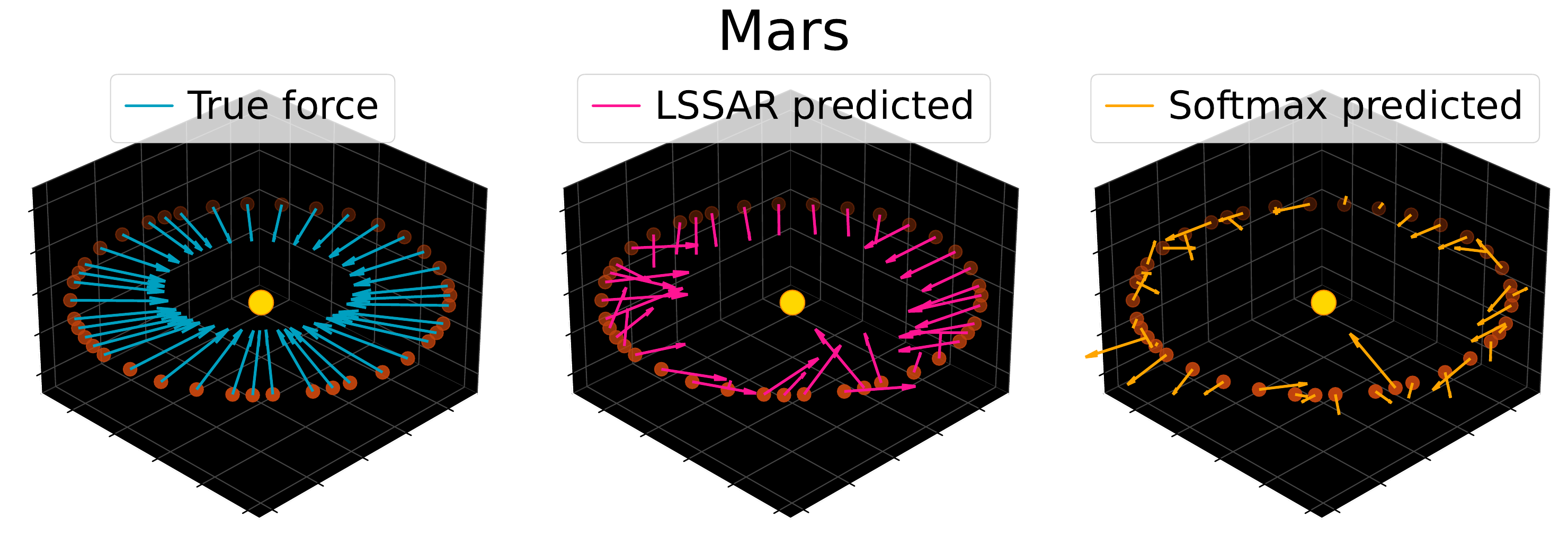}

        \label{fig:force_mars}
    \end{subfigure}
    \hfill
    \begin{subfigure}[b]{0.48\textwidth}
        \centering
        \includegraphics[width=\linewidth]{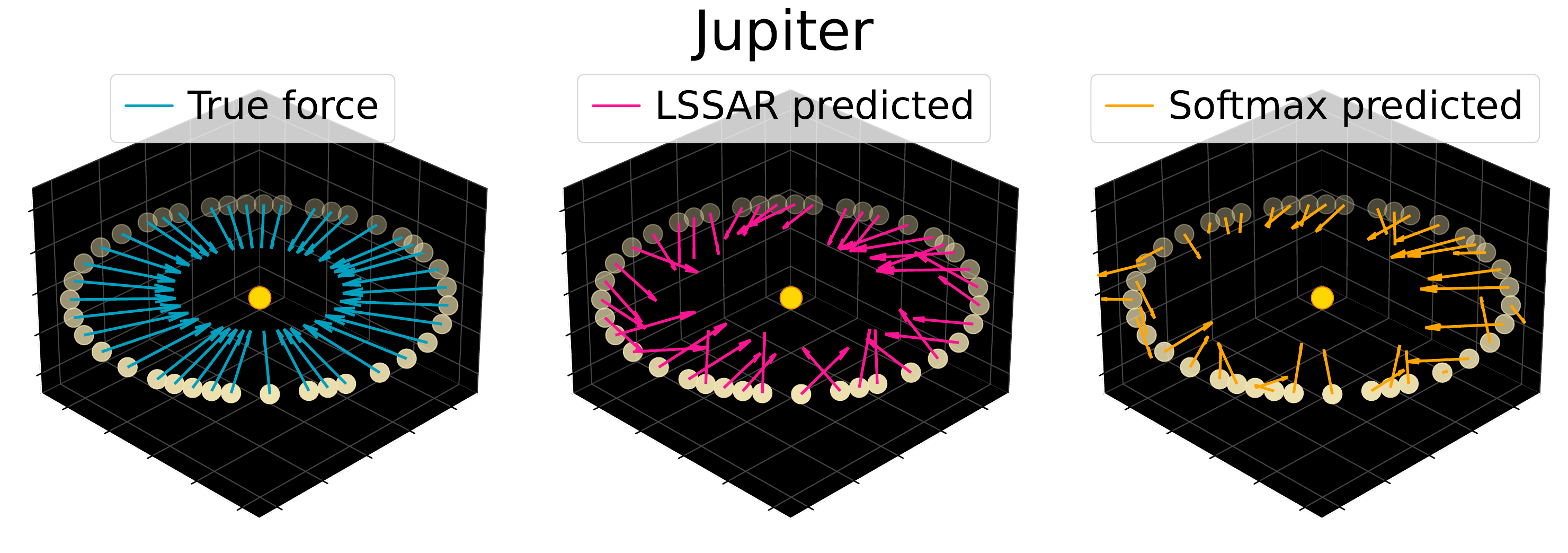}

        \label{fig:force_jupiter}
    \end{subfigure}

    \begin{subfigure}[b]{0.48\textwidth}
        \centering
        \includegraphics[width=\linewidth]{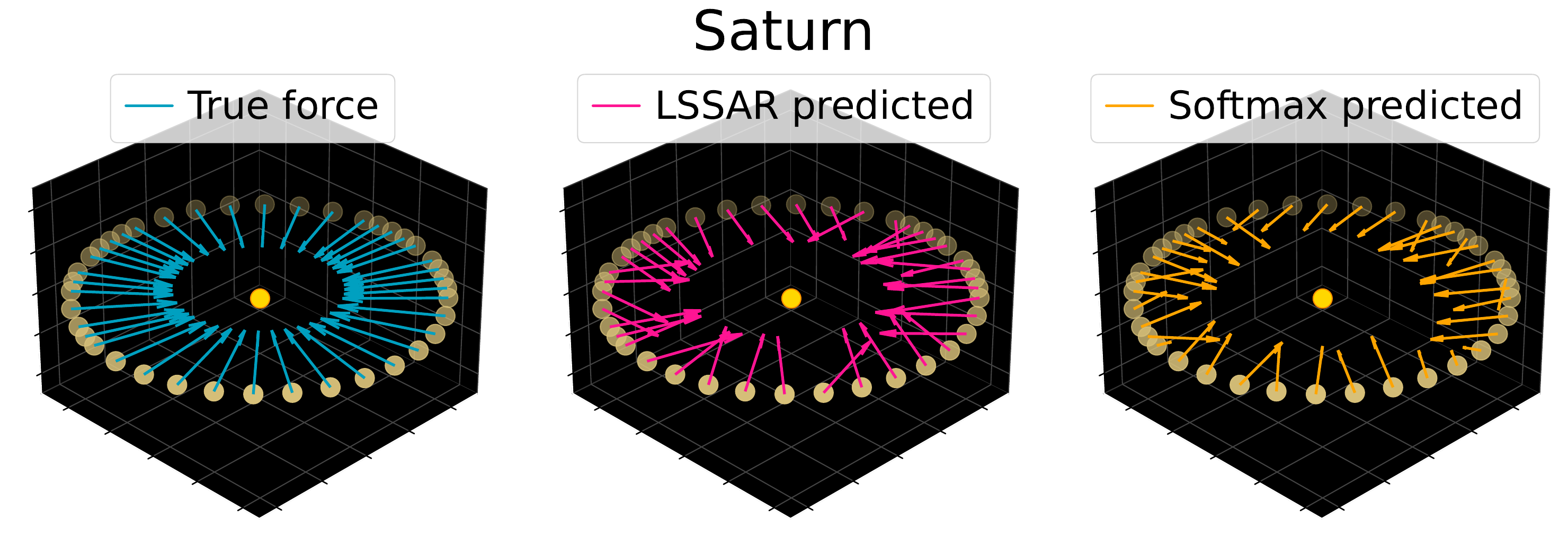}

        \label{fig:force_saturn}
    \end{subfigure}
    \hfill
    \begin{subfigure}[b]{0.48\textwidth}
        \centering
        \includegraphics[width=\linewidth]{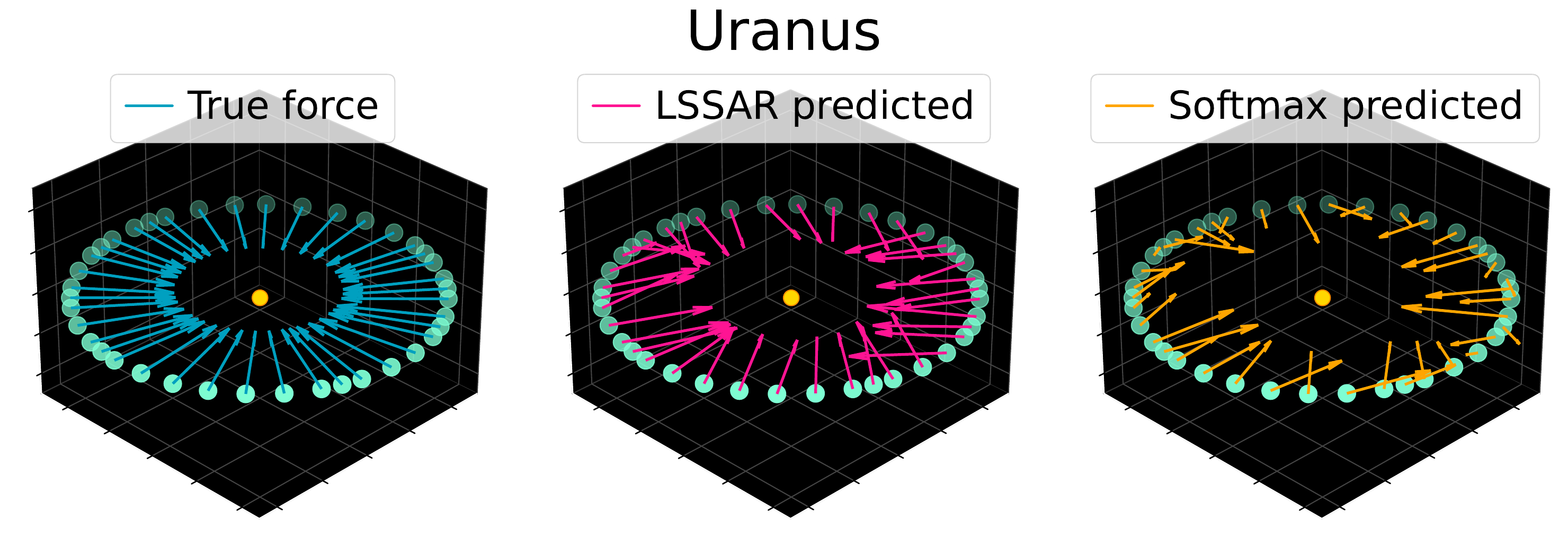}

        \label{fig:force_uranus}
    \end{subfigure}

    \begin{subfigure}[b]{0.48\textwidth}
        \centering
        \includegraphics[width=\linewidth]{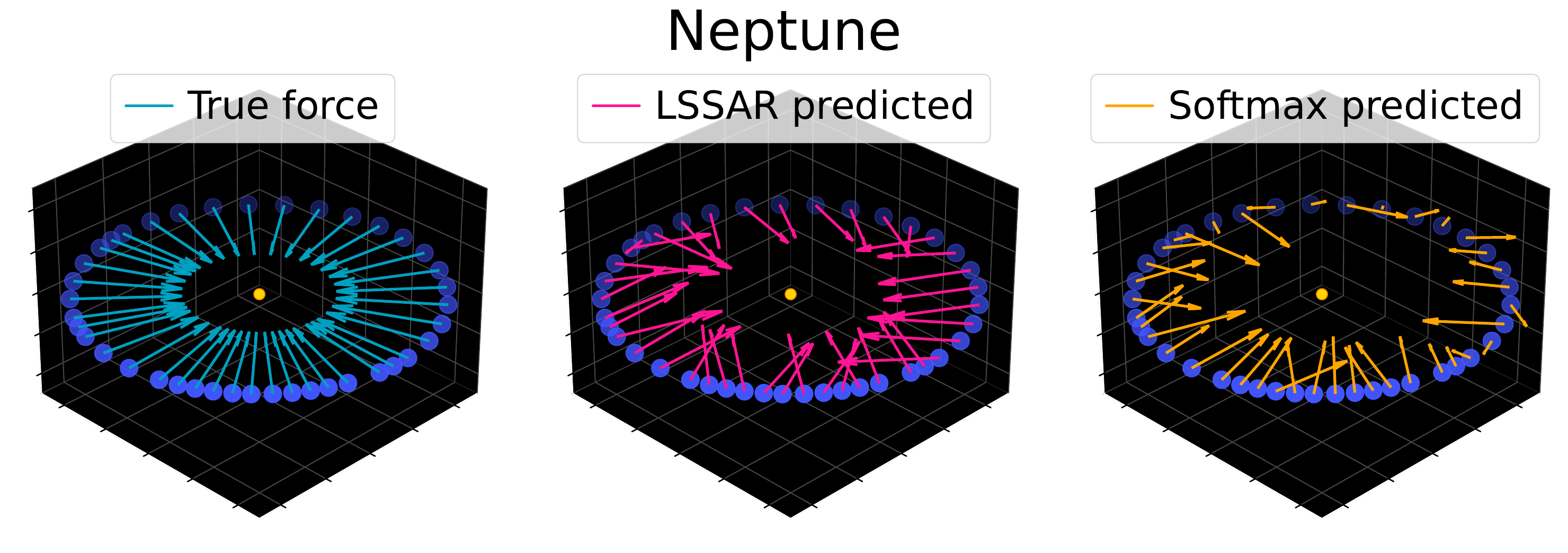}

        \label{fig:force_neptune}
    \end{subfigure}
    \caption{Gravitational force vector predictions for planetary orbits in the solar system (excluding Earth, shown in \cref{fig:force_earth}). For each planet: left panel shows true Newtonian forces, middle panel shows GPT (LSSAR) predictions, and right panel shows GPT (Softmax) predictions. LSSAR generally produces force vectors that point toward the Sun.}
    \label{fig:force_all_planets}
\end{figure}

This section provides visualisations of the gravitational force vector predictions for planetary orbits in the solar system, complementing the Earth orbit analysis presented in Section~\ref{sec:symbolic_regression}. These figures examine whether the qualitative force-vector structure observed for Earth is consistent across planets with varying orbital characteristics.

Across all seven planets shown in \cref{fig:force_all_planets}, LSSAR produces
force vectors that generally point toward the Sun (the gravitational centre). In contrast,
the standard GPT model with Softmax attention generates force predictions that
lack coherent radial structure. These visualisations provide qualitative support
for the symbolic regression results in \cref{tab:symbolic_regression}: the
inverse-square functional form recovered by LSSAR is consistent across multiple
planetary orbits rather than being an artefact of fitting to a single trajectory.

\end{document}